\documentclass[sigconf]{acmart}

\usepackage{algorithm}
\usepackage{graphicx}
\usepackage{amsmath}
\usepackage{algorithmic}
\usepackage{textcomp}
\usepackage{subcaption}
\usepackage{tabularx}
\usepackage{xcolor}
\usepackage{soul}
\usepackage{multirow}
\usepackage{xspace}
\usepackage{adjustbox}
\usepackage{enumitem}
\usepackage{svg}
\usepackage{wrapfig, lipsum, booktabs}
\usepackage{newtxtext}
\usepackage{pgfplots}
\usetikzlibrary{pgfplots.groupplots}
\pgfplotsset{compat=1.3}
\usepackage{tikz}
\usetikzlibrary{patterns}
\usepackage{pgf-pie}
\usepackage{colortbl}
\usepackage{tcolorbox}
\tcbuselibrary{breakable}
\usepackage{hyperref}
\usepackage{makecell}
\usepackage{fontawesome5}
\usepackage[bottom]{footmisc}

\usepackage{adjustbox}

\definecolor{tabhighlight}{HTML}{e5e5e5}
\definecolor{grey}{RGB}{128,138,135}

\definecolor{oorange}{RGB}{215,122,71}
\definecolor{yyellow}{RGB}{230,185,79}
\definecolor{ppurple}{RGB}{122,30,97}
\definecolor{ggreen}{RGB}{112,173,71}

\definecolor{battleshipgrey}{rgb}{0.3, 0.3, 0.3}
\definecolor{brilliantrose}{rgb}{1.0, 0.33, 0.64}
\definecolor{americanrose}{rgb}{1.0, 0.01, 0.24}
\definecolor{jweigreen}{rgb}{0,0.45,0.24}
\definecolor{bluegray}{rgb}{0.1, 0.1, 0.4}
\definecolor{ao(english)}{rgb}{0.0, 0.5, 0.0}
\definecolor{blanchedalmond}{rgb}{1.0, 0.92, 0.8}
\definecolor{atomictangerine}{rgb}{1.0, 0.6, 0.4}
\definecolor{chocolate(web)}{rgb}{0.82, 0.41, 0.12}
\definecolor{bananayellow}{rgb}{1.0, 0.88, 0.21}
\definecolor{goldenbrown}{rgb}{0.6, 0.4, 0.08}
\definecolor{aliceblue}{rgb}{0.94, 0.97, 1.0}
\definecolor{beige}{rgb}{0.96, 0.96, 0.86}
\definecolor{babyblue}{rgb}{0.54, 0.81, 0.94}
\definecolor{camel}{rgb}{0.76, 0.6, 0.42}
\definecolor{cinnamon}{rgb}{0.82, 0.41, 0.12}

\definecolor{redlinkcolor}{rgb}{0.79607843, 0.25098039, 0.25882353}
\definecolor{bluecitecolor}{rgb}{0,0.36,0.69}
\usepackage{hyperref}
\usepackage{booktabs}
\usepackage{makecell}
\usepackage{multirow}
\usepackage{colortbl}
\usepackage{adjustbox}
\usepackage{hyperref}
\usepackage{subcaption}  
\usepackage{colortbl}
\usepackage{adjustbox}
\usepackage{makecell}
\usepackage{graphicx}
\usepackage{booktabs}
\usepackage{xspace}

\usepackage{fontawesome5}  
\usepackage[bottom]{footmisc}  

\newcommand{\impro}[1]{{\hspace{0.05cm}{\textcolor{jweigreen}{\text{(+#1)}}}}}

\newcommand{\model}{TimeChat-Online\xspace}
\newcommand{\dataset}{TimeChat-Online-139K\xspace}
\newcommand{\module}{DTD\xspace}
\newcommand{\videollms}{VideoLLMs\xspace}
\newcommand{\videollm}{VideoLLM\xspace}

\AtBeginDocument{%
  }
    
\renewcommand\footnotetextcopyrightpermission[1]{} 
\setcopyright{none}

\acmSubmissionID{709}

\begin{document}


 

\title{TimeChat-Online: 80\% Visual Tokens are Naturally \\ Redundant in Streaming Videos}

\author[L. Yao, Y. Li, Y. Wei, L. Li, et al.]{
Linli Yao$^{\heartsuit,*}$
\quad Yicheng Li$^{\heartsuit,*}$ 
\quad Yuancheng Wei$^{\diamond,*}$ 
\quad Lei Li$^{\spadesuit}$ 
\quad Shuhuai Ren$^{\heartsuit}$ \\
\quad Yuanxin Liu$^{\heartsuit}$  
\quad Kun Ouyang$^{\heartsuit}$
\quad Lean Wang$^{\heartsuit}$ 
\quad Shicheng Li$^{\heartsuit}$ 
\quad Sida Li$^{\heartsuit}$ \\
\quad Lingpeng Kong$^{\spadesuit}$
\quad Qi Liu$^{\spadesuit}$ 
\quad Yuanxing Zhang$^{\ddagger}$
\quad Xu Sun$^{\heartsuit}$
}
\affiliation{
\institution{\textsuperscript{\rm $\heartsuit$}Peking University, \textsuperscript{\rm $\diamond$}South China University of Technology, \textsuperscript{\rm $\spadesuit$}The University of Hong Kong, \textsuperscript{\rm $\ddagger$}Kuaishou Technology} 
\institution{\textbf{\href{https://timechat-online.github.io}{\textcolor{bluecitecolor}{https://timechat-online.github.io}}}}
}


\thanks{\textsuperscript{\rm *}Equal contribution.}
\thanks{\faEnvelope\ linliyao@stu.pku.edu.cn}




\begin{abstract}

The rapid growth of online video platforms, particularly live streaming services, has created an urgent need for real-time video understanding systems. These systems must process continuous video streams and respond to user queries instantaneously, presenting unique challenges for current Video Large Language Models (\videollms). While existing \videollms excel at processing complete videos, they face significant limitations in streaming scenarios due to their inability to handle dense, redundant frames efficiently.
We introduce \textbf{\model}, a novel online VideoLLM that revolutionizes real-time video interaction. At its core lies our innovative Differential Token Drop (\text{\module}) module, which addresses the fundamental challenge of visual redundancy in streaming videos. 
Drawing inspiration from human visual perception's Change Blindness phenomenon, \module preserves meaningful temporal changes while filtering out static, redundant content between frames. Remarkably, our experiments demonstrate that \module achieves an 82.8\% reduction in video tokens while maintaining 98\% performance on StreamingBench, revealing that over 80\% of visual content in streaming videos is naturally redundant without requiring language guidance.
To enable seamless real-time interaction, we present \text{\dataset}, a comprehensive streaming video dataset featuring diverse interaction patterns including backward-tracing, current-perception, and future-responding scenarios. \model's unique  \textit{Proactive Response} capability, naturally achieved through continuous monitoring of video scene transitions via \module, sets it apart from conventional approaches. 
Our extensive evaluation demonstrates \model's superior performance on streaming benchmarks (StreamingBench and OvOBench) and maintaining competitive results on long-form video tasks such as Video-MME and MLVU. Notably, when integrated with Qwen2.5VL-7B, \module achieves a 5.7-point accuracy improvement on the challenging VideoMME subset containing videos of 30-60 minutes, while reducing video tokens by 84.6\%. Our work establishes a new paradigm for efficient streaming video understanding and reveals the potential of leveraging natural video redundancy for future \videollms development. 

\end{abstract}



\begin{teaserfigure}
    \centering
    \includegraphics[width=\textwidth]{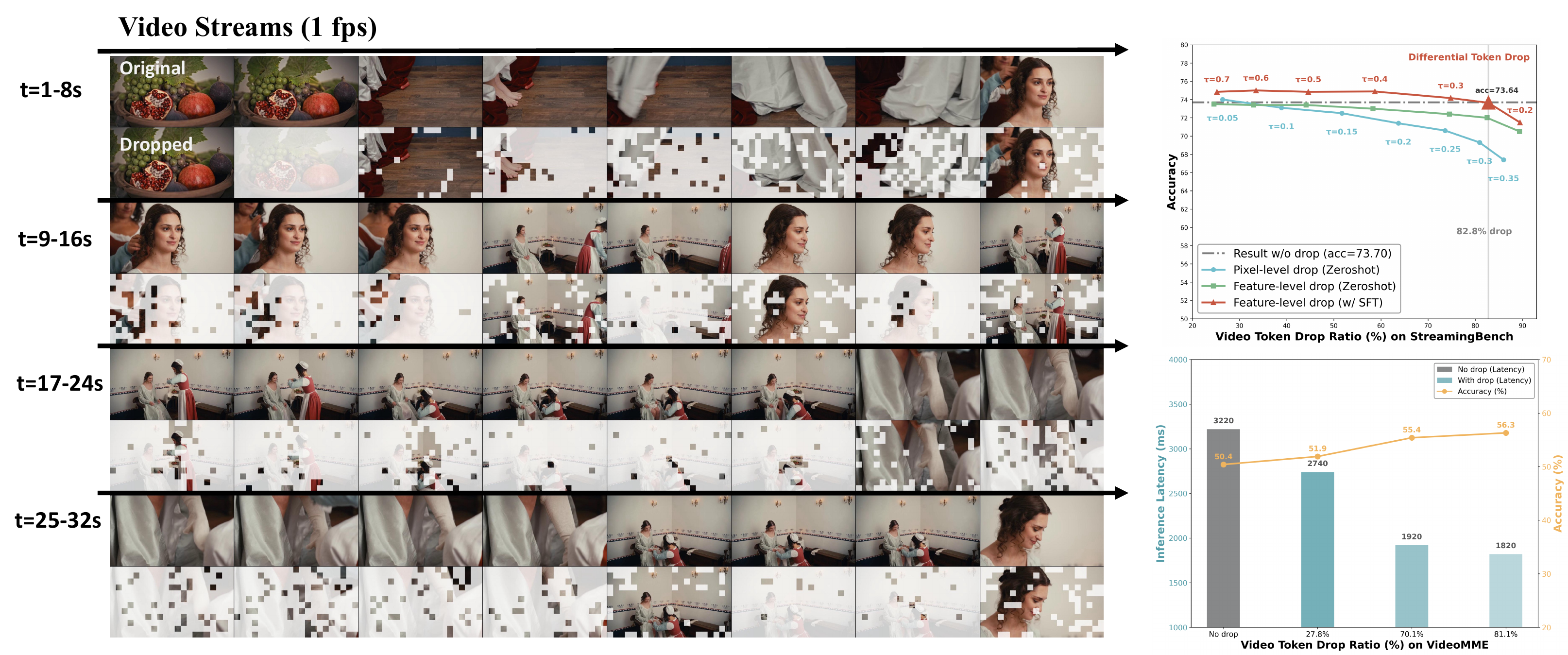}
    \vspace{-5pt}
    \caption{This paper presents \model for efficient Streaming Video Understanding~\cite{videollm-online}. Its core design is the Differential Token Dropping (\module) module that selectively preserves only significant temporal changes across video streams. The \module eliminates 82.8\% of redundant video tokens without any user-query guidance, while achieving a 1.76$\times$ speedup in response latency and maintaining over 98\% of original accuracy. Furthermore, it naturally monitors video scene transitions, facilitating online \textit{Proactive Responding}.}
    \label{fig:teaser}
\end{teaserfigure}

\settopmatter{printfolios=true}
\maketitle
\section{Introduction}


The proliferation of online video platforms (e.g., live broadcasts) and real-world applications (e.g., domestic robots and surveillance systems) has driven researchers to focus on \textit{online} video understanding with continuous streams, commonly referred to as Streaming VideoQA~\cite{videollm-online, flashvstream, videostreaming, ReKV, internlmxcomposer2d5-omni, qwen2.5-omni}.
In streaming video scenarios, video assistants must continuously process incoming frames while simultaneously enabling real-time user interaction, remaining responsive to user queries posed at any moment.


Streaming video tasks introduce two fundamental challenges:
Firstly, \textbf{Long-form High-redundant Video Context:} successive video frames are received at high frame rates (typically 1-10 FPS in real-world applications~\cite{videollm-online}), with neighboring frames exhibiting substantial similarity in backgrounds and static objects. Additionally, video streams are potentially infinite, creating extensive temporal context that must be maintained across the timeline.
Secondly, \textbf{Real-time Interaction with Proactive Responding}: streaming video tasks involve backward tracing, current-time perception, and forward responding. When presented with a user query at a specific moment, a \videollm must efficiently access \textit{past} and \textit{current} visual context to generate immediate responses with minimal latency. For questions requiring \textit{future} visual cues not yet available, the model must possess \textit{Proactive Responding} capabilities to automatically trigger responses at a prospective timestamp when relevant visual cues become available.


Despite recent advancements in Video Large Language Models (\videollms)~\cite{llava-next,timechat,llava-video-sft,qwen2.5vl,minigpt4video,2023videochat,zhang2025videollama3,videot3,internlmxcomposer2d5-omni}, these models struggle with \textit{online} video understanding. They are primarily designed for \textit{offline} video processing, where they receive and process entire videos at once. In online scenarios, they cannot implement proactive responses and face difficulties with long-form high-redundancy video streams: short-context \videollms that uniformly sample sparse video frames~\cite{vila, llava-video-sft,llava-next}, such as 32 or 64 frames, suffer from significant visual context loss; conversely, long-context \videollms that densely sample video frames at 1 FPS (frame-per-second), such as Qwen2.5VL-7B~\cite{qwen2.5vl}, incur substantial response delays when processing computationally intensive video tokens.
Recent approaches have proposed Average Pooling~\cite{llava-onevision, qwen2.5vl} or Resampler-based mechanisms~\cite{videollama, Qwen-VL, dai2023instructblip, timechat, llama-vid} to compress redundancy in long-form videos. However, these compression methods impose a \textit{fixed number of tokens per frame}, failing to adapt to the variable redundancy inherent in dynamic video streams. Concurrently, language-guided approaches~\cite{videochat-flash, longvu, dai2023instructblip} prove inefficient for streaming scenarios, as they necessitate reprocessing of all historical dense frames whenever a new user query is received.

To address these challenges, this paper presents \textbf{\model}, a novel online \videollm that efficiently enables real-time interaction with streaming video content.
To tackle the long-form and high-redundancy challenges of video streams, we propose a \text{\textbf{D}ifferential \textbf{T}oken \textbf{D}rop (\module)} mechanism. Inspired by human visual perception phenomena such as Change Blindness~\cite{change-blindness,changedetection}, our approach mimics how the human visual system processes continuous video streams: not by capturing every detail in each frame, but by selectively focusing on salient spatial-temporal changes while filtering out static, redundant content.
As Figure~\ref{fig:teaser} illustrates, our \module mechanism adaptively preserves only the changed visual tokens between successive frames from a joint spatial-temporal perspective. This approach significantly reduces video token count by 82.8\% in streaming videos purely at the visual level without requiring any textual information. Notably, it maintains VideoQA accuracy at comparable levels to full-token processing, indicating that more than 80\% of the visual context in video streams is naturally redundant.


While \module effectively addresses the challenge of processing long-form high-redundant video streams, we must also tackle the real-time interaction requirements of online video understanding. To this end, we introduce a new training dataset \dataset that addresses the scarcity of data specifically designed for streaming VideoQA. We collect long-form videos averaging 11.1 minutes in length and utilize GPT-4o~\cite{gpt4o} to annotate them with diverse streaming VideoQA pairs that encompass backward tracing, current-time understanding, and forward responding task types. To enable proactive responding for future-oriented questions, we construct \emph{negative samples} in which the questions cannot be answered using the currently available streaming video content.  
For future-responding question interactions, \model is designed to be triggered at video scene transition timestamps, generating new responses with the updated video context. 
As Figure~\ref{fig:teaser} illustrates, scene transitions are naturally indicated by frames with few tokens dropped, signifying significant visual differences from previous frames.


To summarize, our contributions are three-fold:
\textbf{I)} We propose \model with a novel Differential Token Drop module that significantly reduces spatial-temporal redundancy in streaming videos (eliminating 82.8\% of tokens, achieving a 1.76$\times$ speed-up while maintaining 98\% accuracy on StreamingBench).
The \module can also be seamlessly integrated with general VideoLLMs such as Qwen2.5VL to substantially enhance its efficiency in long video tasks (reducing 87.2\% of video tokens and yielding a 5.6 absolute accuracy gain on the VideoMME long set).
\textbf{II)} We develop a new instruction-tuning dataset, \dataset, specifically designed to facilitate more flexible streaming VideoQA interactions.
\textbf{III)} Experimental results demonstrate that \model achieves state-of-the-art performance on streaming video benchmarks, including StreamingBench (58.0) and OVO-Bench (47.6). Furthermore, it maintains competitive performance on general long-form video understanding tasks such as VideoMME (63.3), MLVU (65.4), and LongVideoBench (57.7).

\section{Related Work}
\noindent \textbf{Streaming Video Understanding.}
Streaming video understanding, which processes continuously updating video frames, was first introduced by VideoLLM-online~\cite{videollm-online}. Recent advancements follow two primary directions. 
The \textbf{first} focuses on efficient encoding of dense video streams. Memory-bank-based approaches like VideoStreaming~\cite{videostreaming}, Flash-VStream~\cite{flashvstream}, StreamChat~\cite{streamchat}, and VideoChat-online~\cite{videochat-online} utilize dynamic memory banks to retain informative video tokens. ReKV~\cite{ReKV} and Inf-MLLM~\cite{inf-MLLM} optimize KV Cache management for video context, while VideoLLM-MoD~\cite{videollm-mod} employs Mixture-of-Depth to reduce token count. Most of these methods require language guidance from user queries to select relevant video content. 
In contrast, our proposed \module reduces  video tokens by over 80\% purely-visually as an early step to lighten the computational burden of video tokens before language model processing.
The \textbf{second direction} enhances real-time interaction experiences in streaming scenarios. IXC2.5-Omni~\cite{internlm-xcomposer2.5-omnilive} and Qwen2.5-Omni~\cite{qwen2.5-omni} incorporate audio modality, MMDuet~\cite{mmduet} refines video-text duet interaction formats, and Dispider~\cite{dispider} introduces a disentangled Perception-Decision-Reaction paradigm. Our work differs by implementing a novel proactive response paradigm that  leverages our synthetic streaming training dataset \dataset and is intelligently triggered by scene transitions naturally revealed by \module's token drop ratio curves.

\noindent \textbf{Efficient Video Token Pruning.}
Recent advancements in long video processing have led to diverse approaches for video token pruning~\cite{deco, visionzip, yang2025topv, videochat-flash, pvc, liu2025hybrid, Ren2023TESTA}. A branch of methods~\cite{timechat, Qwen2VL, slowfast-llava, videochat-flash, pvc} compress frames or clips to a fixed number, disregarding the dynamic visual redundancy inherent in different videos. For instance, LLama-vid~\cite{li2024llamavid} represents each frame with fixed two tokens.
To address the limitations, several methods~\cite{visionzip, dycoke, chat-univi, Ren2023TESTA} design adaptive token merging in either spatial or temporal dimensions. While they offer improved flexibility, they blur the vanilla spatial-temporal positions which will hurt the fine-grained video perception such as spatial localization, action ordering, or temporal counting task.
Furthermore, selection-based methods without merging~\cite{wen2025stop} primarily consider spatial redundancy while neglecting temporal redundancy. In contrast, our \module adaptively drops video tokens via dynamic temporal redundancy while preserving the related spatial-temporal positions.
Another category of approaches~\cite{longvu, fastv, h2o} leverages language guidance through user queries or vision-language cross-attention mechanisms for token pruning. However, these language-guided methods are inefficient for streaming scenarios, as they require reprocessing all historical frames for each new user query. This significantly increases computational burden and introduces response delays, making them impractical for real-world applications. In contrast, our \module efficiently processes video streams by calculating redundancy only for newly-arriving frames with faster speed.

\begin{figure*}[t!]
    \centering
    \includegraphics[width=0.98\textwidth]{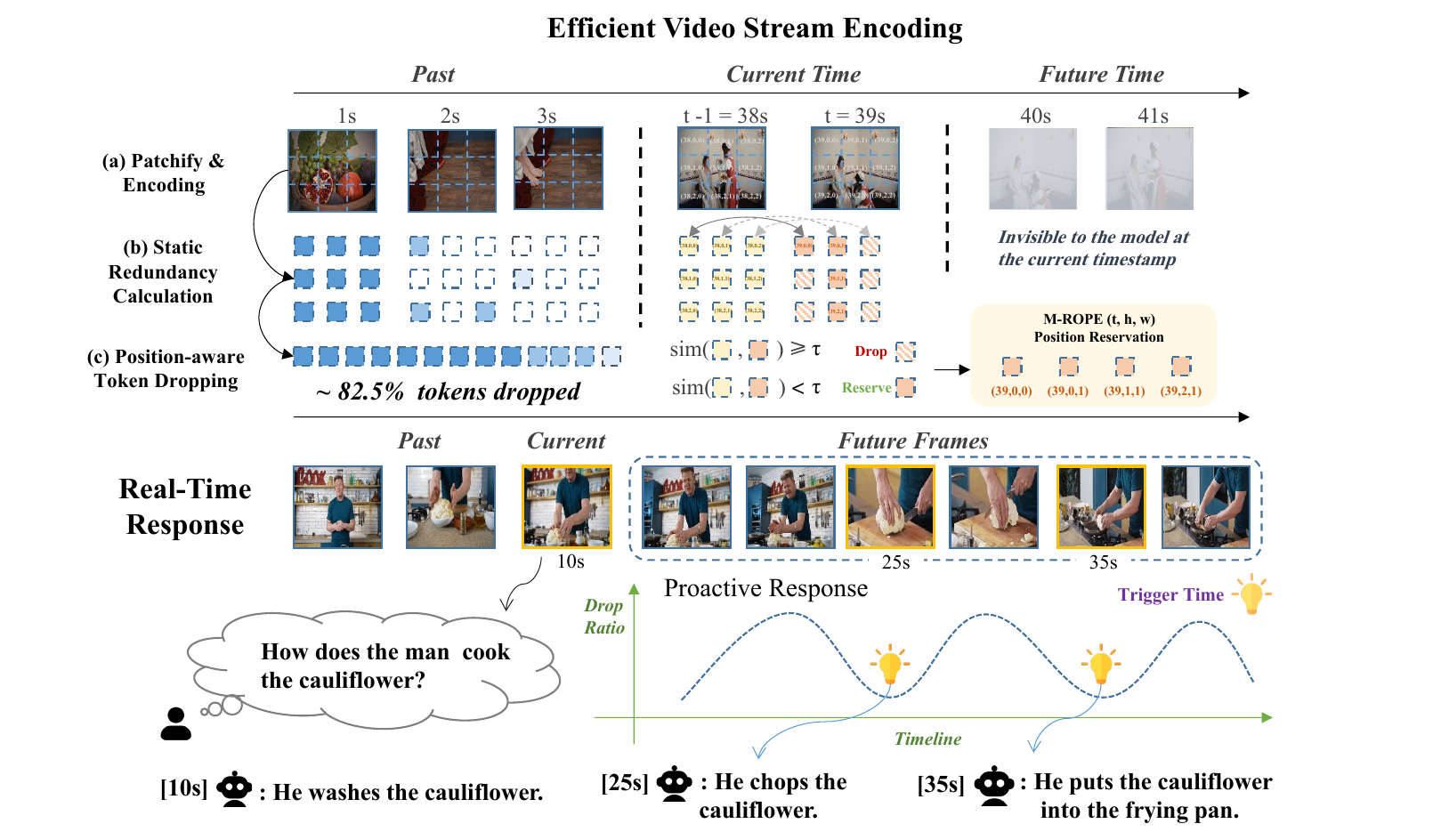}
    \caption{The core of \model lies in the Differential Token Dropping (\module) design for efficiently encoding video streams. \module captures significant temporal changes through three steps: (a) patchifying and encoding dense video frames, (b) calculating static redundancy between temporally-consecutive and spatially-identical video tokens, (c) dropping temporally-redundant video tokens while preserving the (temporal, height, width) positions of remaining tokens. \module dynamically eliminates visual redundancy in the temporal dimension, yielding an adaptive \textit{drop ratio} for each frame. During Real-Time Interaction, frames with low \textit{drop ratios} in the timeline indicate video scene transitions, triggering \model to achieve Proactive Responding at these scene-oriented timestamps.}
    \label{fig:model}
\end{figure*}


\section{\model Framework}
This paper addresses the streaming video question-answering (Streaming VideoQA) task~\cite{streamchat,streamingbench} and introduces an online video assistant, \model, capable of providing real-time responses to user queries posed at specific timestamps within video streams.
In Section~\ref{sec:task_formula}, we first formulate the Streaming VideoQA task and distinguish it from offline video understanding tasks.
Then, in Section~\ref{sec:token_drop}, we present the design of the Differential Token Drop, which eliminates over 80\% of video tokens by retaining significant temporal changes while removing static redundancy.
Next, we introduce the curation process of \dataset for training streaming video tasks in Section~\ref{sec:data_collection} and describe the training strategy and inference procedure in Section~\ref{sec:training}.

\subsection{Streaming VideoQA Formulation}
\label{sec:task_formula}
Given a video consisting of $T$ frames in total, the streaming task will consistently update the incoming video frames $\mathcal{F} = [f_1, f_2, ..., f_t]$ where $1 \leq t \leq T$ along the timeline rather than given all video frames at once. When a user question $Q_t$ is proposed at a \textit{Current} timestamp $t$, the goal of Streaming-VideoQA task is to leverage the \textit{Current} frame content $f_t$ and the \textit{Past} video streams $\mathcal{F}_{1:t-1}$ to answer the $Q_t$.

\noindent\textbf{Proactive Responding.} When the historical video content $\mathcal{F}_{1:t}$ is insufficient to answer the user question $Q_t$, the model should be able to proactively respond at a \textit{Future} timestamp $t'$ ($t \leq t' \leq T$) when the updated video content $\mathcal{F}_{t+1:t'}$ is sufficient to answer $Q_t$. Moreover, for questions that can be partially answered with current content, a satisfactory video assistant should also be able to provide new responses at \textit{Future} timestamps as more visual information becomes available. For example, as illustrated in Figure~\ref{fig:model}, when the user asks "How does the man cook the cauliflower?" at $t=10s$, the video assistant can proactively respond at future timestamps $t'=25s$ (``He chops the cauliflower'') and $t'=35s$ (``He puts the cauliflower into the frying pan'') with newly visual content about ``the cooking man'' as the video progresses over time.

\noindent\textbf{Online \videollms vs. Offline \videollms.} For streaming video content $\mathcal{F}_{1:t}$ and a specific user question $Q_t$, it can be converted to an offline VideoQA task by inputting the historical frames $\mathcal{F}_{1:t}$ as the whole video content along with the user question $Q_t$ to existing offline \videollms. However, these offline \videollms have two main limitations: \textbf{Firstly}, existing offline \videollms cannot efficiently handle the dense frames of high-FPS video streams (e.g., 1-10 FPS). 
Vanilla dense-sampling \videollms such as Qwen2VL-7B~\cite{qwen2.5vl} that can take 1 FPS frames as input incur unsatisfactory response delays when processing long-form video streams.
Conversely, sparse sampling \videollms such as LLava-Video-7B~\cite{llava-video-sft} and VILA~\cite{vila}, which uniformly sample a fixed number of frames (64 and 14 frames) as input, suffer from significant visual information loss.
\textbf{Secondly}, existing offline \videollms are not able to provide proactive responses at future timestamps for questions that require upcoming visual context, which significantly limits the user interaction experience.
Therefore, this paper aims to propose an online video assistant that can both efficiently handle high-FPS sampled video streams and provide proactive interaction for future-responding questions.

\subsection{Differential Temporal Token Drop}
\label{sec:token_drop}
Inspired by the \textit{Change Blindness} phenomenon in human visual perception~\cite{change-blindness}, we propose a  Differential Token Drop (\module) module to efficiently reduce video redundancy by only preserving the significant temporal changes in  video streams.
As Figure~\ref{fig:model} shows, \module consists of three main steps: (a) Patchify and Encoding, (b) Static Redundancy Calculation, and (c) Position-aware Token Dropping. 

\subsubsection{Patchify and Encoding.} We use a ViT~\cite{vit} to split each video frame $f_t$ into a sequence of visual patches $\mathcal{P}_t=[p_1, p_2, ..., p_{H\times W}]$ and encode the related spatial tokens as  $\mathcal{V}_t=[v_1, v_2, ..., v_{H\times W}]$, where $H$ and $W$ denotes the maximum position index in height and width respectively. For video streams $\mathcal{F}_{1:t}$ received at timestamp $t$, we can get the temporal sequence of patches $\mathcal{P}_{1:t}$ and the related visual tokens $\mathcal{V}_{1:t}$.

\subsubsection{Static Redundancy Calculation.} 
\label{sec:redunt_calculate}
Intuitively, if temporally consecutive frames $(f_{t-1}, f_{t})$ are visually similar, we determine the latter frame $f_{t}$ as a redundant frame, because it contains the same static visual content as the former frame $f_{t-1}$.
From a more fine-grained perspective, we can formulate static redundancy by comparing temporally-consecutive patches $\mathcal{P}$ or visual tokens $\mathcal{V}$ at pixel-level and feature-level, respectively.

\noindent\textbf{Pixel-level Redundancy.}
Inspired by the RLT~\cite{rlt} method, we select two temporally consecutive and spatially identical \textit{patches} $(p^{t-1}_{hw}, p^{t}_{hw})$, where $h$ and $w$ are the height and width indices in the spatial dimension. Next, we calculate the pixel similarity between them using the L1 distance~\cite{l1} as:

\begin{equation}
    \label{eq:patch_similarity}
    Sim(p^{t-1}_{hw}, p^{t}_{hw}) = \left\| p^{t-1}_{hw} - p^{t}_{hw} \right\|_1 < \tau_{pixel}
\end{equation}

If the selected two patches are visually similar, they will have close pixel values, leading to a small L1 distance. We define a hyperparameter $\tau_{pixel}$ to determine the threshold of pixel redundancy.

\noindent\textbf{Feature-level Redundancy.}
Besides pixel-level redundancy, we can also calculate the feature-level visual redundancy. The cosine similarity~\cite{steck2024cosine} between temporally-consecutive spatially-aligned \textit{visual tokens} ${v}^{t-1}_{hw}$ and ${v}^{t}_{hw}$ is defined as:

\begin{equation}
    \label{eq:token_similarity}
    Sim(v^{t-1}_{hw}, v^{t}_{hw}) = \frac{v^{t-1}_{hw} \cdot v^{t}_{hw}}{\left\| v^{t-1}_{hw} \right\|_2 \left\| v^{t}_{hw} \right\|_2} > \tau_{feat}
\end{equation}

\begin{table*}[t]
    \small
    \centering
    \caption{Performance comparison on StreamingBench focusing on \textit{Real-Time Visual Understanding} tasks. Real-Time Visual Understanding encompasses Object Perception (OP), Causal Reasoning (CR), Clips Summarization (CS), Attribute Perception (ATP), Event Understanding (EU), Text-Rich Understanding (TR), Prospective Reasoning (PR), Spatial Understanding (SU), Action Perception (ACP), and Counting (CT). ``\text{VTokens(\%)}'' represents the percentage of video tokens remaining after dropping, where $100\%$ indicates no dropping, and $\downarrow82.6\%$ signifies an 82.6\% reduction in video tokens. The \textbf{bold} values indicate the best performance and \underline{underlined} values indicate the second best.}
    \label{tab:streamingbench}
    \small
    \begin{adjustbox}{max width=1.9\columnwidth}
    \begin{tabular}{l@{\hspace{3pt}}c@{\hspace{3pt}}c | c c c c c c c c c c | c}
    \toprule
    \multirow{1}{*}{\textbf{Model}} & \multirow{1}{*}{\textbf{\#Frames}} & \textbf{VTokens(\%)} & \textbf{OP} & \textbf{CR} & \textbf{CS} & \textbf{ATP} & \textbf{EU} & \textbf{TR} & \textbf{PR} & \textbf{SU} & \textbf{ACP} & \textbf{CT} & \textbf{All} \\
    \midrule
    \midrule
    Human & - & - 
        & 89.47 & 92.00 & 93.60 & 91.47 & 95.65 & 92.52 & 88.00 & 88.75 & 89.74 & 91.30 & 91.46 \\
    \midrule
    \multicolumn{14}{c}{\textbf{Proprietary MLLMs}} \\
    \midrule
    Gemini 1.5 pro~\cite{gemini} & 1 fps & - 
        & 79.02 & 80.47 & 83.54 & 79.67 & 80.00 & 84.74 & 77.78 & 64.23 & 71.95 & 48.70 & 75.69 \\
    GPT-4o~\cite{gpt4o} & 64 & - 
        & 77.11 & 80.47 & 83.91 & 76.47 & 70.19 & 83.80 & 66.67 & 62.19 & 69.12 & 49.22 & 73.28 \\
    Claude 3.5 Sonnet~\cite{claude3_5} & 20 & - 
        & 73.33 & 80.47 & 84.09 & 82.02 & 75.39 & 79.53 & 61.11 & 61.79 & 69.32 & 43.09 & 72.44 \\
    \midrule
    \multicolumn{14}{c}{\textbf{Open-source Offline \videollms}} \\
    \midrule
    \textcolor{grey}{Video-LLaMA2-7B~\cite{videollama2}} & \textcolor{grey}{32} & \textcolor{grey}{-} 
        & \textcolor{grey}{55.86} & \textcolor{grey}{55.47} & \textcolor{grey}{57.41} & \textcolor{grey}{58.17} & \textcolor{grey}{52.80} & \textcolor{grey}{43.61} & \textcolor{grey}{39.81} & \textcolor{grey}{42.68} & \textcolor{grey}{45.61} & \textcolor{grey}{35.23} & \textcolor{grey}{49.52} \\
    \textcolor{grey}{VILA-1.5-8B~\cite{vila}} & \textcolor{grey}{14} & \textcolor{grey}{-} 
        & \textcolor{grey}{53.68} & \textcolor{grey}{49.22} & \textcolor{grey}{70.98} & \textcolor{grey}{56.86} & \textcolor{grey}{53.42} & \textcolor{grey}{53.89} & \textcolor{grey}{54.63} & \textcolor{grey}{48.78} & \textcolor{grey}{50.14} & \textcolor{grey}{17.62} & \textcolor{grey}{52.32} \\
    \textcolor{grey}{Video-CCAM-14B~\cite{videoccam}} & \textcolor{grey}{96} & \textcolor{grey}{-} 
        & \textcolor{grey}{56.40} & \textcolor{grey}{57.81} & \textcolor{grey}{65.30} & \textcolor{grey}{62.75} & \textcolor{grey}{64.60} & \textcolor{grey}{51.40} & \textcolor{grey}{42.59} & \textcolor{grey}{47.97} & \textcolor{grey}{49.58} & \textcolor{grey}{31.61} & \textcolor{grey}{53.96} \\
    \textcolor{grey}{LongVA-7B~\cite{longva}} & \textcolor{grey}{128} & \textcolor{grey}{-} 
        & \textcolor{grey}{70.03} & \textcolor{grey}{63.28} & \textcolor{grey}{61.20} & \textcolor{grey}{70.92} & \textcolor{grey}{62.73} & \textcolor{grey}{59.50} & \textcolor{grey}{61.11} & \textcolor{grey}{53.66} & \textcolor{grey}{54.67} & \textcolor{grey}{34.72} & \textcolor{grey}{59.96} \\
    \textcolor{grey}{InternVL-V2-8B~\cite{internvl1.5}} & \textcolor{grey}{16} & \textcolor{grey}{-} 
        & \textcolor{grey}{68.12} & \textcolor{grey}{60.94} & \textcolor{grey}{69.40} & \textcolor{grey}{77.12} & \textcolor{grey}{67.70} & \textcolor{grey}{62.93} & \textcolor{grey}{59.26} & \textcolor{grey}{53.25} & \textcolor{grey}{54.96} & \textcolor{grey}{56.48} & \textcolor{grey}{63.72} \\
    \textcolor{grey}{Kangaroo-7B~\cite{kangaroo}} & \textcolor{grey}{64} & \textcolor{grey}{-} 
        & \textcolor{grey}{71.12} & \textcolor{grey}{84.38} & \textcolor{grey}{70.66} & \textcolor{grey}{73.20} & \textcolor{grey}{67.08} & \textcolor{grey}{61.68} & \textcolor{grey}{56.48} & \textcolor{grey}{55.69} & \textcolor{grey}{62.04} & \textcolor{grey}{38.86} & \textcolor{grey}{64.60} \\
    \textcolor{grey}{LLaVA-NeXT-Video-32B~\cite{llava-next}} & \textcolor{grey}{64} & \textcolor{grey}{-} 
        & \textcolor{grey}{78.20} & \textcolor{grey}{70.31} & \textcolor{grey}{73.82} & \textcolor{grey}{76.80} & \textcolor{grey}{63.35} & \textcolor{grey}{69.78} & \textcolor{grey}{57.41} & \textcolor{grey}{56.10} & \textcolor{grey}{64.31} & \textcolor{grey}{38.86} & \textcolor{grey}{66.96} \\
    \textcolor{grey}{MiniCPM-V-2.6-8B~\cite{minicpm}} & \textcolor{grey}{32} & \textcolor{grey}{-} 
        & \textcolor{grey}{71.93} & \textcolor{grey}{71.09} & \textcolor{grey}{77.92} & \textcolor{grey}{75.82} & \textcolor{grey}{64.60} & \textcolor{grey}{65.73} & \textcolor{grey}{70.37} & \textcolor{grey}{56.10} & \textcolor{grey}{62.32} & \textcolor{grey}{53.37} & \textcolor{grey}{67.44} \\
    \textcolor{grey}{LLaVA-OneVision-7B~\cite{llava-onevision}} & \textcolor{grey}{32} & \textcolor{grey}{-} 
        & \textcolor{grey}{80.38} & \textcolor{grey}{74.22} & \textcolor{grey}{76.03} & \textcolor{grey}{80.72} & \textcolor{grey}{72.67} & \textcolor{grey}{71.65} & \textcolor{grey}{67.59} & \textcolor{grey}{65.45} & \textcolor{grey}{65.72} & \textcolor{grey}{45.08} & \textcolor{grey}{71.12} \\
    \textcolor{grey}{Qwen2.5-VL-7B~\cite{qwen2.5vl}} & \textcolor{grey}{1 fps} & \textcolor{grey}{}
        & \textcolor{grey}{78.32} & \textcolor{grey}{80.47} & \textcolor{grey}{78.86} & \textcolor{grey}{80.45} & \textcolor{grey}{76.73} & \textcolor{grey}{78.50} & \textcolor{grey}{79.63} & \textcolor{grey}{63.41} & \textcolor{grey}{66.19} & \textcolor{grey}{53.19} & \textcolor{grey}{73.68} \\
    \addlinespace[1pt]
    \midrule
    \multicolumn{14}{c}{\textbf{Open-source  Online \videollms}} \\
    \midrule
    Flash-VStream-7B~\cite{flashvstream} & - & -
        & 25.89 & 43.57 & 24.91 & 23.87 & 27.33 & 13.08 & 18.52 & 25.20 & 23.87 & 48.70 & 23.23 \\
    VideoLLM-online-8B~\cite{videollm-online} & 2 fps & -
        & 39.07 & 40.06 & 34.49 & 31.05 & 45.96 & 32.40 & 31.48 & 34.16 & 42.49 & 27.89 & 35.99 \\
    Dispider-7B~\cite{dispider}~\textsubscript{\textcolor{grey}{[CVPR 2025]}} & 1 fps & -
        & 74.92 & 75.53 & 74.10 & 73.08 & 74.44 & 59.92 & 76.14 & 62.91 & 62.16 & 45.80 & 67.63 \\
    \rowcolor{gray!20} \textbf{\model-7B} & 1 fps & $\downarrow$44.2\% &80.76 &79.69 &80.76 &83.33 &74.84 &78.82 &78.70 &64.23 &68.75 &57.98 &$\textbf{75.28}$  \\
    \rowcolor{gray!40} \textbf{\model-7B} & 1 fps & $\downarrow$82.6\%
            &79.13 &81.25 &78.86 &80.77 &70.44 &77.26 &77.78 &67.07 &66.19 &53.72 &$\underline{73.64}$ \\
    \midrule
    \textbf{\model-7B} &1 fps&100\%
        & 80.22 & 82.03 & 79.50 & 83.33 & 76.10 & 78.50 & 78.70 & 64.63 & 69.60 & 57.98 & 75.36 \\
    \bottomrule
    \end{tabular}
\end{adjustbox}
\end{table*}

If the selected two visual tokens are visually similar, they will have similar feature embeddings, leading to a large cosine similarity. We define a hyperparameter $\tau_{feat}$ to determine the threshold of feature-level redundancy.
The values of $\tau_{pixel}$ and $\tau_{feat}$ can control the overall \textit{drop ratio} of video tokens. They measure how different two temporally consecutive patches/tokens are, depending only on the inherent nature of videos. Empirically, we verify that the $\tau$ values are consistent across different datasets. For example, $\tau_{feat}=0.25$ corresponds to around 85\% of video tokens being dropped across three datasets, while $\tau_{feat}=0.5$ corresponds to around 45\%.
Notably, unlike feature-level token pruning approaches~\cite{fastv} that operate inside LLMs, our pixel-level and feature-level calculations are purely vision-based strategies that operate \textit{before} the Large Language Model (LLM)~\cite{llama3, qwen2} module, which achieves greater efficiency.
\subsubsection{Position-aware Token Dropping.} 
Based on the calculated similarity between token (or patch) pairs, we identify redundant tokens (or patches) in the \textit{Current} frame $f_t$ and generate a binary mask $M^{t}_\text{drop}$ that indicates whether each token (or patch) should be dropped or preserved. For feature-level dropping, we directly identify which visual tokens in $\mathcal{V}_t$ to drop. For pixel-level dropping, we map the dropped patch indices to their corresponding visual tokens, leveraging the one-to-one correspondence maintained by the ViT encoder between input patches and output visual tokens. We then eliminate redundant static tokens from the current frame $f_t$ by applying the binary mask $M^{t}_\text{drop}$ to the visual tokens $\mathcal{V}_t$ and related position embeddings $\mathcal{P}_t$:

\begin{equation}
    \label{eq:token_dropping}
    \widetilde{\mathcal{V}}_{t} = \mathcal{V}_t \circ M^{t}_\text{drop}, \quad \widetilde{\mathcal{P}}_{t} = \mathcal{P}_t \circ M^{t}_\text{drop}
\end{equation}

\noindent\textbf{Spatial-Temporal Position Reserving.}
After dropping the marked redundant tokens, the vanilla spatial and temporal position relations of the remaining tokens will be disrupted. This disruption would harm fine-grained spatial/temporal perception tasks such as Spatial Localization~\cite{llava-st} or Action Recognition~\cite{zhang2024vinoground}. To preserve the relative spatial-temporal positions, we leverage Multi-modal Rotary Position Embedding (M-ROPE) derived from QwenVL-series models~\cite{Qwen2VL,qwen2.5vl}. M-ROPE indexes the 3D \{temporal, height, width\} position $p=\{t,h,w\}$ of each video token. We calculate the 3D M-ROPE positions \textit{before dropping} to record the original spatial-temporal structure of video tokens. When dropping visual tokens, we simultaneously drop their M-ROPE position embeddings as shown in the equation~\ref{eq:token_dropping}.

As depicted in Figure~\ref{fig:model}, the remaining visual tokens $\widetilde{\mathcal{V}}_{t}$ are aligned with their vanilla 3D position embeddings calculated before dropping.
Finally, we can parallelize the position-aware token dropping operation for all temporal-consecutive frame pairs in $\mathcal{F}_{1:t}$ and get the final dropped video token sequence $\widetilde{\mathcal{V}}_{1:t}$ across the timeline. 

\begin{table*}[t]
    \small
    \centering
    \caption{Evaluation results on OVO-Bench~\cite{ovobench} comprising three categories: i) \textit{Real-Time Visual Perception} (OCR: Optical Character Recognition, ACR: Action Recognition, ATR: Attribute Recognition, STU: Spatial Understanding, FPD: Future Prediction, OJR: Object Recognition), ii) \textit{Backward Tracing} (EPM: Episodic Memory, ASI: Action Sequence Identification, HLD: Hallucination Detection), and iii) \textit{Forward Active Responding} (REC: Repetition Event Count, SSR: Sequential Steps Recognition, CRR: Clues Reveal Responding).}
    \label{tab:ovo-bench}

    \begin{adjustbox}{max width=1.9\columnwidth}
    \begin{tabular}{l@{}c@{\hspace{3pt}}|ccccccc|cccc|cccc|c}
    \toprule
    \multirow{2}{*}{\textbf{Model}} & \multirow{2}{*}{\textbf{\#Frames}} & 
    \multicolumn{7}{c|}{\textbf{Real-Time Visual Perception}} & 
    \multicolumn{4}{c|}{\textbf{Backward Tracing}} & 
    \multicolumn{4}{c|}{\textbf{Forward Active Responding}} & \multirow{2}{*}{\textbf{Overall}} \\
    \addlinespace[2pt]
    \cmidrule[0.5pt](lr){3-9} \cmidrule[0.5pt](lr){10-13} \cmidrule[0.5pt](lr){14-17}
    \addlinespace[2pt]
    & & OCR & ACR & ATR & STU & FPD & OJR & Avg. & 
    EPM & ASI & HLD & Avg. & 
    REC & SSR & CRR & Avg. & Avg. \\
    \midrule
    \midrule
    Human Agents & - & 94.0 & 92.6 & 94.8 & 92.7 & 91.1 & 94.0 & 93.2 & 92.6 & 93.0 & 91.4 & 92.3 & 95.5 & 89.7 & 93.6 & 92.9 & 92.8 \\
    \midrule
    \multicolumn{18}{c}{\textbf{Proprietary Multimodal Models}} \\
    \midrule
    Gemini 1.5 Pro & 1fps & 87.3 & 67.0 & 80.2 & 54.5 & 68.3 & 67.4 & 70.8 & 68.6 & 75.7 & 52.7 & 62.3 & 35.5 & 74.2 & 61.7 & 57.2 & 65.3 \\
    GPT-4o & 64 & 69.1 & 65.1 & 65.5 & 50.0 & 68.3 & 63.7 & 63.6 & 49.8 & 71.0 & 55.4 & 58.7 & 27.6 & 73.2 & 59.4 & 53.4 & 58.6 \\
    \midrule
    \multicolumn{18}{c}{\textbf{Open-source Offline \videollms}} \\
    \midrule
    \textcolor{grey}{LLaVA-NeXT-Video-7B} & \textcolor{grey}{64} & \textcolor{grey}{69.8} & \textcolor{grey}{59.6} & \textcolor{grey}{66.4} & \textcolor{grey}{50.6} & \textcolor{grey}{72.3} & \textcolor{grey}{61.4} & \textcolor{grey}{63.3} & \textcolor{grey}{51.2} & \textcolor{grey}{64.2} & \textcolor{grey}{9.7} & \textcolor{grey}{41.7} & \textcolor{grey}{34.1} & \textcolor{grey}{67.6} & \textcolor{grey}{60.8} & \textcolor{grey}{54.2} & \textcolor{grey}{53.1} \\
    \textcolor{grey}{LLaVA-OneVision-7B} & \textcolor{grey}{64} & \textcolor{grey}{67.1} & \textcolor{grey}{58.7} & \textcolor{grey}{69.8} & \textcolor{grey}{49.4} & \textcolor{grey}{71.3} & \textcolor{grey}{60.3} & \textcolor{grey}{62.8} & \textcolor{grey}{52.5} & \textcolor{grey}{58.8} & \textcolor{grey}{23.7} & \textcolor{grey}{45.0} & \textcolor{grey}{24.8} & \textcolor{grey}{66.9} & \textcolor{grey}{60.8} & \textcolor{grey}{50.9} & \textcolor{grey}{52.9} \\
    \textcolor{grey}{Qwen2-VL-7B} & \textcolor{grey}{64} & \textcolor{grey}{69.1} & \textcolor{grey}{53.2} & \textcolor{grey}{63.8} & \textcolor{grey}{50.6} & \textcolor{grey}{66.3} & \textcolor{grey}{60.9} & \textcolor{grey}{60.7} & \textcolor{grey}{44.4} & \textcolor{grey}{66.9} & \textcolor{grey}{34.4} & \textcolor{grey}{48.6} & \textcolor{grey}{30.1} & \textcolor{grey}{65.7} & \textcolor{grey}{50.8} & \textcolor{grey}{48.9} & \textcolor{grey}{52.7} \\
    \textcolor{grey}{InternVL-V2-8B} & \textcolor{grey}{64} & \textcolor{grey}{68.5} & \textcolor{grey}{58.7} & \textcolor{grey}{69.0} & \textcolor{grey}{44.9} & \textcolor{grey}{67.3} & \textcolor{grey}{56.0} & \textcolor{grey}{60.7} & \textcolor{grey}{43.1} & \textcolor{grey}{61.5} & \textcolor{grey}{27.4} & \textcolor{grey}{44.0} & \textcolor{grey}{25.8} & \textcolor{grey}{57.6} & \textcolor{grey}{52.9} & \textcolor{grey}{45.4} & \textcolor{grey}{50.1} \\
    \textcolor{grey}{LongVU-7B} & \textcolor{grey}{1fps} & \textcolor{grey}{55.7} & \textcolor{grey}{49.5} & \textcolor{grey}{59.5} & \textcolor{grey}{48.3} & \textcolor{grey}{68.3} & \textcolor{grey}{63.0} & \textcolor{grey}{57.4} & \textcolor{grey}{43.1} & \textcolor{grey}{66.2} & \textcolor{grey}{9.1} & \textcolor{grey}{39.5} & \textcolor{grey}{16.6} & \textcolor{grey}{69.0} & \textcolor{grey}{60.0} & \textcolor{grey}{48.5} & \textcolor{grey}{48.5} \\
    \midrule
    \multicolumn{18}{c}{\textbf{Open-source Online Video-LLMs}} \\
    \midrule
    Flash-VStream-7B & 1fps & 25.5 & 32.1 & 29.3 & 33.7 & 29.7 & 28.8 & 29.9 & 36.4 & 33.8 & 5.9 & 25.4 & 5.4 & 67.3 & 60.0 & 44.2 & 33.2 \\
    VideoLLM-online-8B & 2fps & 8.1 & 23.9 & 12.1 & 14.0 & 45.5 & 21.2 & 20.8 & 22.2 & 18.8 & 12.2 & 17.7 & - & - & - & - & - \\
    \rowcolor{gray!20} \textbf{\model-7B} & 1fps ($\downarrow$44.6\%) &74.5  &48.6  &68.1  &48.3  &69.3  &59.8  &61.4  &56.9  &64.9  &11.8  &44.5  &31.8  &38.5  &40.0  &36.8  & \textbf{47.6\impro{14.4}} \\
    \rowcolor{gray!40} \textbf{\model-7B} & 1fps ($\downarrow$84.8\%) &69.8  &48.6  &64.7  &44.9  &68.3  &55.4  &58.6  &53.9  &62.8  &9.1  &42.0  &32.5  &36.5  &40.0  &36.4  & \underline{45.6}\impro{12.4} \\
    \midrule
    \textbf{\model-7B} & 1fps (100\%) & 75.2 & 46.8 & 70.7 & 47.8 & 69.3 & 61.4 & 61.9 & 55.9 & 59.5 & 9.7 & 41.7 & 31.6 & 38.5 & 40.0 & 36.7 & 46.7 \\
    \bottomrule
    \end{tabular}
    \end{adjustbox}
\end{table*}

\noindent\textbf{Discussion: Advantages of Differential Token Dropping.}
\module is a purely vision-based approach to reduce video tokens by preserving significant temporal changes across video frames.
Compared with existing token pruning methods~\cite{slowfast-llava,longvu,dycoke,inf-MLLM,streamchat}, \module offers 
three key advantages: (1) \textit{video-aware dynamic pruning} that adaptively reduces video tokens from a holistic video perspective, well-suited for both high-speed and slow-motion videos; (2) \textit{positional reservation} that maintains the fine-grained spatial-temporal positions of retained tokens; and (3) \textit{streaming-friendly} operation that calculates visual redundancy only for newly incoming frames without re-processing historical video content. Moreover, \module is orthogonal to memory-based~\cite{flashvstream} or KV-Cache Retrieval streaming methods~\cite{ReKV,videostreaming}, serving as an initial efficient step to reduce the computational video burden and being complementary to these approaches.

\subsection{\dataset Collection}
\label{sec:data_collection}
To better apply the \module design, we introduce \dataset, a comprehensive synthetic streaming VideoQA dataset specifically designed for training online-\videollms.
Existing works~\cite{videochat-online,videollm-online, timechat} convert dense video narrations~\cite{videollm-online} or timestamp-related tasks~\cite{etbench} into streaming dialogue datasets. However, these transformed data samples are limited in question-answer diversity and fail to mimic real-world interactions.
Instead, our dataset encompasses diverse online tasks across backward tracing, 
real-time perception, and forward active responding to facilitate more flexible \textit{Real-time Interaction}.
Specifically, we collect long-range and visually informative videos, annotate scene-oriented dense captions, and 
produce diverse streaming question-answer samples using GPT-4o~\cite{gpt4o}.

\noindent\textbf{Step 1: Visually Informative Video Collection.}
The quality of video content is crucial for the Streaming VideoQA task. Videos with monotonous or static scenes provide limited information for multiple-turn real-time interactions. To address this, we collect visually informative videos characterized by diverse scene changes. First, we source long-form videos~\cite{timechat} and segment them into successive scenes using PySceneDetect\footnote{https://www.scenedetect.com/}. We then extract key frames from each scene segment and eliminate redundant frames based on visual similarity measured by DINO-v2~\cite{dinov2}. 
To ensure sufficient visual diversity and information richness, we select videos containing more than 5 distinct scenes.

\noindent\textbf{Step 2: Scene-oriented Detailed Caption Generation.}
Following the collection of videos and extraction of key frames, we employ GPT-4o~\cite{gpt4o} to generate scene-oriented dense captions for each frame. These comprehensive descriptions encompass both static and dynamic visual elements, including shot types, object appearances and actions, environmental and background variations, and camera movements. To enhance the model's ability to discern scene transitions, we provide preceding frame caption as contextual information.

\noindent\textbf{Step 3: Streaming VideoQA Generation.}
We generate streaming VideoQA samples by providing the dense captions of the current scene and the previous scenes as context. We utilize GPT-4o to generate diverse question-answer pairs, including \textit{Temporal-enhanced}, \textit{Backward Tracing}, \textit{Real-Time Visual Perception}, and \textit{Forward Active Responding} QAs following previous works\cite{ovobench,streamingbench}.

\noindent\textbf{Step 4: Negative Samples for Future-Response Training.}
To facilitate the training of proactive responding capabilities, we construct negative samples that intentionally cannot be answered using the currently available streaming video content. We carefully select irrelevant video frames before the question timestamp and generate corresponding answers labeled as ``unanswerable''.

\noindent\textbf{Data Statistics.} We collect 11,043 visually informative videos ranging from 5 minutes to several hours in length, with an average duration of 11.1 minutes per video. From each video, we extract an average of 87.8 key frames with approximately 7.14 seconds between consecutive frames. Each key frame is annotated with a detailed description averaging 176 words. Based on the dense descriptions, we generate 139K question-answer pairs.  
Details regarding data statistics, task types, and GPT-4o prompts are provided in the Appendix.

\begin{table*}[t]
    \small
    \centering
    \begin{minipage}[t]{0.59\textwidth}
        \centering
        \caption{\small Results on offline long video benchmarks. We report the accuracy on the MLVU~\cite{MLVU}, LongVideoBench~\cite{longvideobench} and VideoMME(w/o subtitles)~\cite{videomme}. $\dagger$ indicates the reproduced results.}
        \label{tab:long_video_benchmarks}
        \begin{adjustbox}{max width=\textwidth}
        \begin{tabular}{lc|cccc}
        \toprule
        \multirow{2}{*}{\textbf{Model}} & \multirow{2}{*}{\textbf{\#Frames}} & \multirow{2}{*}{\textbf{MLVU}} & \multirow{2}{*}{\textbf{LongVideoBench}} & \multicolumn{2}{c}{\textbf{VideoMME}} \\
        \cmidrule(lr){5-6}
        \addlinespace[1pt]
        & & & & overall & long \\
        \midrule
        \textbf{Video Length} & - & 3$\sim$120 min & 8 sec$\sim$60 min & 1$\sim$60 min & 30$\sim$60 min \\
        \midrule
        \multicolumn{6}{c}{\textbf{Open-Source Offline \videollms}} \\
        \midrule
        LLaMA-VID-7B & 1fps & 33.2 & - & - & - \\
        MovieChat-7B & 2048 & 25.8 & - & 38.2 & 33.4 \\
        LLaVA-Next-Video-7B & 32 & - & 43.5 & 46.6 & - \\
        VideoChat2-7B & 16 & 47.9 & 39.3 & 39.5 & 33.2 \\
        LongVA-7B & 128 & 56.3 & - & 52.6 & 46.2 \\
        Kangaroo-7B & 64 & 61.0 & 54.2 & 56.0 & 46.6 \\
        Video-CCAM-14B & 96 & 63.1 & - & 53.2 & 46.7 \\
        VideoXL-7B & 128 & 64.9 & - & 55.5 & 49.2 \\
        Qwen2.5-VL-7B$\dagger$ & 1fps (100\%) & 66.9 & 61.5 & 63.2 & 50.4 \\
        \rowcolor{gray!20} Qwen2.5-VL-7B~$_{\text{w/ \module}}$ & 1fps ($\downarrow$46.2\%) & 68.6 & 61.6 & 63.4 & 51.9 \\
        \rowcolor{gray!40} Qwen2.5-VL-7B~$_{\text{w/ \module}}$ & 1fps ($\downarrow$84.6\%) & 68.8 & 59.3 & 64.9 & 56.1 \\
        \midrule
        \multicolumn{6}{c}{\textbf{Open-source Online  \videollms}} \\
        \midrule
        Dispider-7B~\textsubscript{\textcolor{grey}{[CVPR 2025]}} & 1fps & 61.7 & - & 57.2 & - \\
        VideoChat-Online-8B~\textsubscript{\textcolor{grey}{[CVPR 2025]}} & 2fps & - & - & 52.8 & 44.9 \\ \midrule
        \textbf{\model-7B} & 1fps (100\%) & 62.6 & 55.4 & 62.4 & 48.4  \\ 
        \rowcolor{gray!20} \textbf{\model-7B} & 1fps ($\downarrow$46.3\%) & 62.9 & \text{57.1} & \textbf{63.3} & \textbf{52.4} \\
        \rowcolor{gray!40} \textbf{\model-7B} & 1fps ($\downarrow$85.0\%) & \textbf{65.4} & \textbf{57.7} & 62.5 & 49.2 \\
        \bottomrule
        \end{tabular}
        \end{adjustbox}
    \end{minipage}\hfill
    \begin{minipage}[t]{0.39\textwidth}
        \centering
        \caption{Ablation study of different token dropping strategies on StreamingBench \textit{(Real-Time Visual Understanding)}. For fair comparison, all methods are evaluated with Qwen2.5VL-7B (Vanilla) without supervised fine-tuning (SFT). $\Delta$ represents accuracy retention rate (\%) relative to the full token setting.}
        \label{tab:abla_drop}
        \begin{adjustbox}{max width=\textwidth}
        \begin{tabular}{l@{\hspace{4pt}}c@{\hspace{4pt}}c@{\hspace{4pt}}}
        \toprule
        \multirow{2}{*}{\textbf{Method w/o SFT}} & \textbf{StreamingBench} & \\
        & \textbf{(3s$\sim$24min)} & $\Delta$ \\
        \midrule
        \multicolumn{3}{l}{\textbf{Vanilla, Avg \underline{22K} Video Tokens per Video (1 fps)}} \\
        \midrule
        Qwen2.5VL-7B  & 73.7 & 100\% \\
        \midrule
        \multicolumn{3}{l}{\textbf{Drop 44.1\%, Avg \underline{12K} Video Tokens per Video}} \\
        \midrule
        \quad + VisionZip~\textsubscript{\textcolor{grey}{[CVPR 2025]}} & 72.3 & 98.1\% \\
        \quad + Pixel-level drop & 72.8 & 98.8\% \\
        \quad + Feature-level drop (frame-aware) & 73.0 & 99.1\% \\
        \rowcolor{gray!20} \quad + Feature-level drop (video-aware) & \textbf{73.4} & \textbf{99.6\%} \\
        \midrule
        \multicolumn{3}{l}{\textbf{Drop 82.5\%, Avg \underline{4K} Tokens per Video}} \\
        \midrule
        \quad + 
        VisionZip~\textsubscript{\textcolor{grey}{[CVPR 2025]}} & 68.8 & 93.4\% \\
        \quad + Pixel-level drop & 68.8 & 93.4\% \\
        \quad + Feature-level drop (frame-aware) & \textbf{72.0} & \textbf{97.7\%} \\
        \rowcolor{gray!20} \quad + Feature-level drop (video-aware) & \textbf{72.0} & \textbf{97.7\%} \\
        \bottomrule
        \end{tabular}
        \end{adjustbox}
    \end{minipage}
\end{table*}

\subsection{Training and Real-Time Inference}
\label{sec:training}
We implement \model based on the long-context Qwen2.5-VL \cite{qwen2.5vl} architecture to support high-FPS dense frame processing. We incorporate \module to efficiently reduce visual redundancy in video streams. Our model is trained on a combination of our streaming dataset \dataset and offline video understanding datasets (LLaVA-Video-178K~\cite{llava-video-sft}, Tarsier2~\cite{tarsier2} and VideoChat-Flash~\cite{videochat-falsh}) to ensure robust performance. During training, we densely sample frames at 1 FPS to simulate streaming video input with a maximum sequence length of N frames. For \module, we apply token dropping with 50\% probability to each training batch.

During inference, \model processes video frames at 1 FPS and uses \module to drop approximately 85\% of video tokens. A first-in-first-out memory bank stores the slimmed historical video tokens. It is worth noting that this memory design is not the emphasis of this paper, and can be replaced with alternatives~\cite{flashvstream,ReKV}. 

\noindent\textbf{Proactive Response with Trigger Time.}
For \textit{Future-responding} questions, the model must decide the optimal timestamps to generate a proactive response. We define these decision time points as \textit{Trigger times}, which always occur when the video transitions to a new scene~\cite{dispider}.
These \textit{Trigger times} can be effectively monitored through the token drop ratio curve along the timeline, as illustrated in Figure~\ref{fig:model} (bottom). The valleys in this curve indicate moments when the current frame content changes significantly from previous frames, revealing a scene transition.
We implement proactive response by generating responses at each trigger time using the most recent video content. 
Through training with the negative samples in \dataset (Section~\ref{sec:data_collection}), the model learns to generate an ``unanswerable'' response when the available video content is insufficient at a particular trigger time, effectively instructing the system to wait for the next trigger time to respond again.


\vspace{-10pt}

\section{Experiments}

\subsection{Implementation Details}

We implement our model using the Qwen2.5VL 7B architecture. \textbf{For training}, we sample frames at 1 FPS with a maximum sequence length of 64 frames for streaming VideoQA samples. We configure the model with a maximum input resolution of 448$\times$448 pixels, threshold parameters $\tau_{feat}=0.7$, batch size 128, and learning rate 1e-5.
Our training procedure combines both offline and online datasets, including subsets of LLaVA-Video-178K (100K samples), Tarsier2 (100K samples), VideoChat-Flash (3K samples) and \dataset, among which negative samples with the ``unanswerable'' label are 20K. We keep the vision encoder frozen while fine-tuning the full-parameter projector and language model for 1 epoch. All experiments are performed on 8$\times$A800 80G GPUs. The hard drop ratio to determine \textit{Trigger Time} is 60\%.
\textbf{During inference}, we maintain the 1 FPS frame processing rate while extending the maximum frame length to 1016. We use feature-level dropping for all experiments by default since it performs consistently better than pixel-level dropping as shown in Figure~\ref{fig:teaser}. We set $\tau_{feat}=0.25 / 0.5$ corresponding to around 85\% / 45\% token dropping rate by default. For real-time interaction, the most recent 6K slimmed video tokens after dropping are preserved in a first-in-first-out memory bank, introducing a maximum latency of 2 seconds for responding.


\subsection{Results on Streaming Video Benchmarks}
We first evaluate our model's performance on two streaming VideoQA benchmarks: StreamingBench~\cite{streamingbench} and OVO-Bench~\cite{ovobench}. For these evaluations, \videollms processes the historical video content received before the \textit{Current} timestamp, which represents the moment when a user question is posed.

\noindent\textbf{StreamingBench.} Table~\ref{tab:streamingbench} shows that \model achieves state-of-the-art performance with a score of 75.28 on the Real-time Visual Understanding subtask of StreamingBench, representing a 7.65 improvement over the recent online model Dispider-7B~\cite{dispider} scored 67.63. This demonstrates that \model effectively combines the superior VideoQA capabilities of offline \videollms with the real-time streaming inference capabilities of online \videollms.
Moreover, this 75.28 score surpasses both the best offline \videollm Qwen2.5VL-7B (73.68) and proprietary models including GPT-4o (73.28) and Claude-3.5-Sonnet (72.44).

When compared to Qwen2.5VL-7B with 1fps full token inputting, \model achieves superior performance (75.28 vs. 73.68) while requiring 44.2\% fewer video tokens. Even at an extreme token dropping ratio of 82.8\%, \model maintains comparable results to the full token setting of Qwen2.5VL-7B (73.64 vs. 73.68). These findings highlight the substantial redundancy in 1 fps video streams and the effectiveness of our \module approach.

\noindent\textbf{OVO-Bench.} Table~\ref{tab:ovo-bench} presents the results on OVO-Bench, which comprehensively evaluates \textit{Backward Tracing} and \textit{Forward Active Responding} capabilities across 12 diverse subtasks. \model substantially outperforms existing online \videollms, achieving a final score of 47.6, which represents a significant 14.4-point absolute improvement over Flash-VStream~\cite{flashvstream} and VideoLLM-online~\cite{videollm-online}. Notably, with 84.8\% of video tokens dropped, \model maintains robust performance with a score of 45.6.

\begin{table}[tbp]
    \small
    \centering
    \caption{Impact of training datasets on StreamingBench \textit{(Real-Time Visual Understanding)} performance across different token drop ratios. \textit{Ours} refers to our proposed dataset \dataset.}
    \label{tab:abla_dataset}
    \begin{adjustbox}{max width=\columnwidth}
    \begin{tabular}{l c c}
    \toprule
    \multirow{2}{*}{\textbf{Dataset}} & \multicolumn{2}{c}{\textbf{Token Drop Ratio}} \\
    \addlinespace[2pt]
    \cmidrule(lr){2-3}
    \addlinespace[2pt]
    & \textbf{$\downarrow$44\%} & \textbf{$\downarrow$82.6\%} \\
    \hline
    \addlinespace[2pt]
    Qwen2.5-VL-7B (Vanilla) & 73.4 & 72.0 \\
    \quad+ LLaVA-Video\textsubscript{\tiny{100K}} & 73.4 & 72.3 \\
    \quad+ LLaVA-Video\textsubscript{\tiny{100K}} + Tarsier2\textsubscript{\tiny{129K}} & 73.5 & 72.5 \\
    \quad+ LLaVA-Video\textsubscript{\tiny{100K}} + Tarsier2\textsubscript{\tiny{129K}} + VideoChat-Flash\textsubscript{\tiny{3K}} & 74.0 & 72.8 \\
    \rowcolor{gray!20} \quad + Above Offline Datasets + Ours & \textbf{75.3} & \textbf{73.6} \\
    \bottomrule
    \end{tabular}
    \end{adjustbox}
\end{table}

\noindent\textbf{Case Study.} Figure~\ref{fig:streamingbench_case} visualizes a proactive responding case of \model. When a user proposes a question ``What specifically did the woman in red do?'' that can also be answered by future moments, \model proactively generates responses at future trigger times (i.e., the video scene transition timestamps), which are indicated by frames with low token drop ratios.



\subsection{Results on Offline Long Video Tasks}

We also evaluate \model on three offline long-form video understanding benchmarks: VideoMME~\cite{videomme}, MLVU~\cite{MLVU}, and LongVideoBench~\cite{longvideobench}. In the offline setting, the entire video is provided as input to the \videollms.
Table~\ref{tab:long_video_benchmarks} demonstrates that \model exhibits superior offline video understanding capabilities compared to recent state-of-the-art online \videollms, including VideoChat-Online~\cite{videochat-online} and Dispider-7B. Leveraging the efficiency of \module, \model performs particularly well on extremely long-form videos, such as the long subset of VideoMME. Compared to VideoChat-Online, \model achieves a 7.5-point improvement (from 44.9 to 52.4) on the long subset of VideoMME which contains videos ranging from 30 to 60 minutes in length. 
We also report Qwen2.5-VL-7B~$_{\text{w/ \module}}$ zeroshot results, as \module can be integrated directly without additional training. Surprisingly, increasing token drop ratio from 46.2\% to 84.6\% consistently improves Qwen2.5-VL-7B's performance on MLVU and VideoMME, making it superior to the 100\% full token setting. For VideoMME's long subset (30-60min), accuracy rises from 50.4 to 56.1 using only 15.4\% of retained video tokens. This indicates substantial vision redundancy in long videos, and reducing this redundancy can simplify \videollm's vision perception with shorter context, thereby enhancing overall performance.

\begin{figure}[t]
    \centering
    \includegraphics[width=0.95\linewidth]{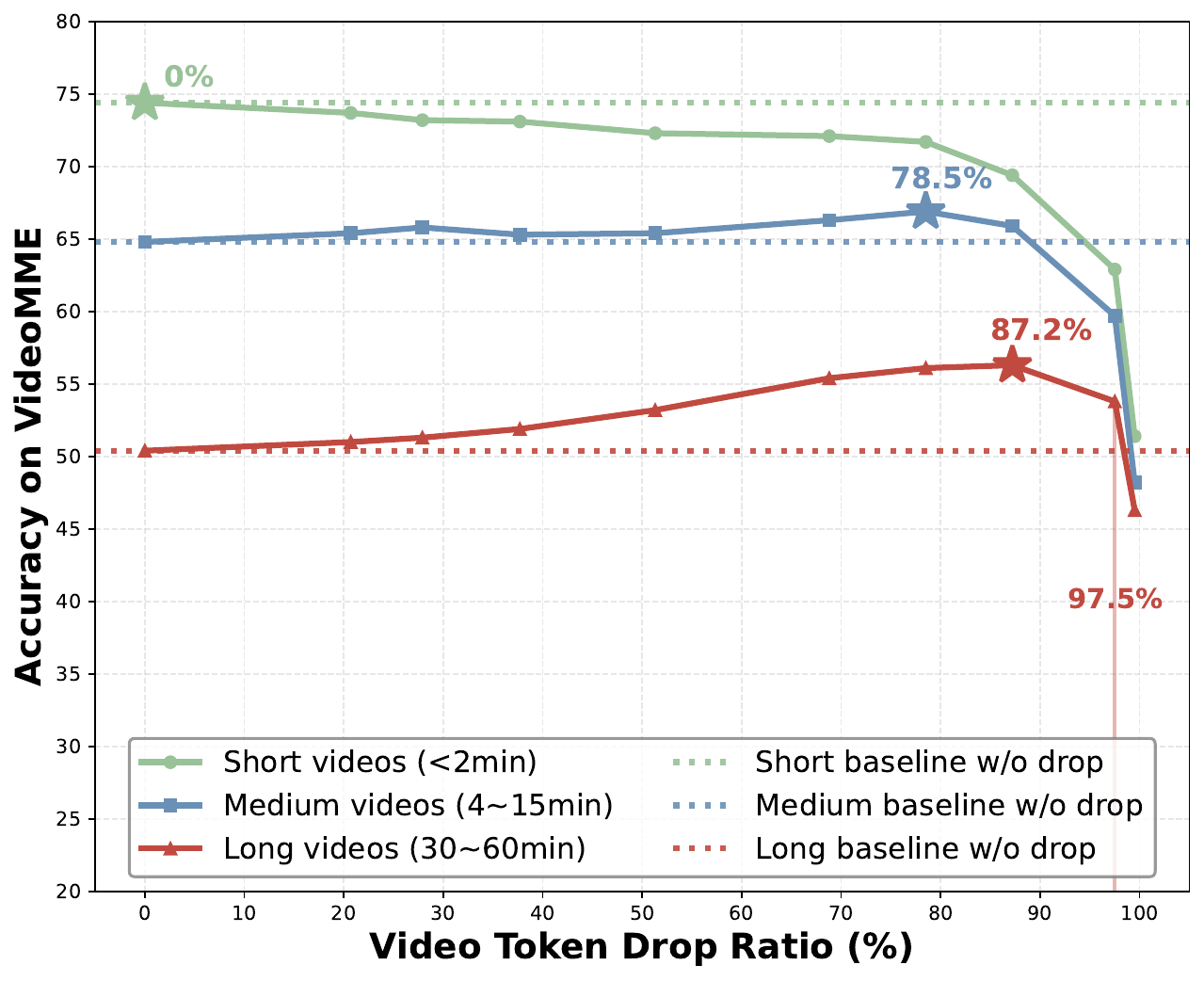}
    \caption{\small Video redundancy of different video length on VideoMME~\cite{videomme}.}
    \label{fig:combined_analysis}
\end{figure}

\begin{figure*}[t]
    \centering
    \includegraphics[width=\linewidth]{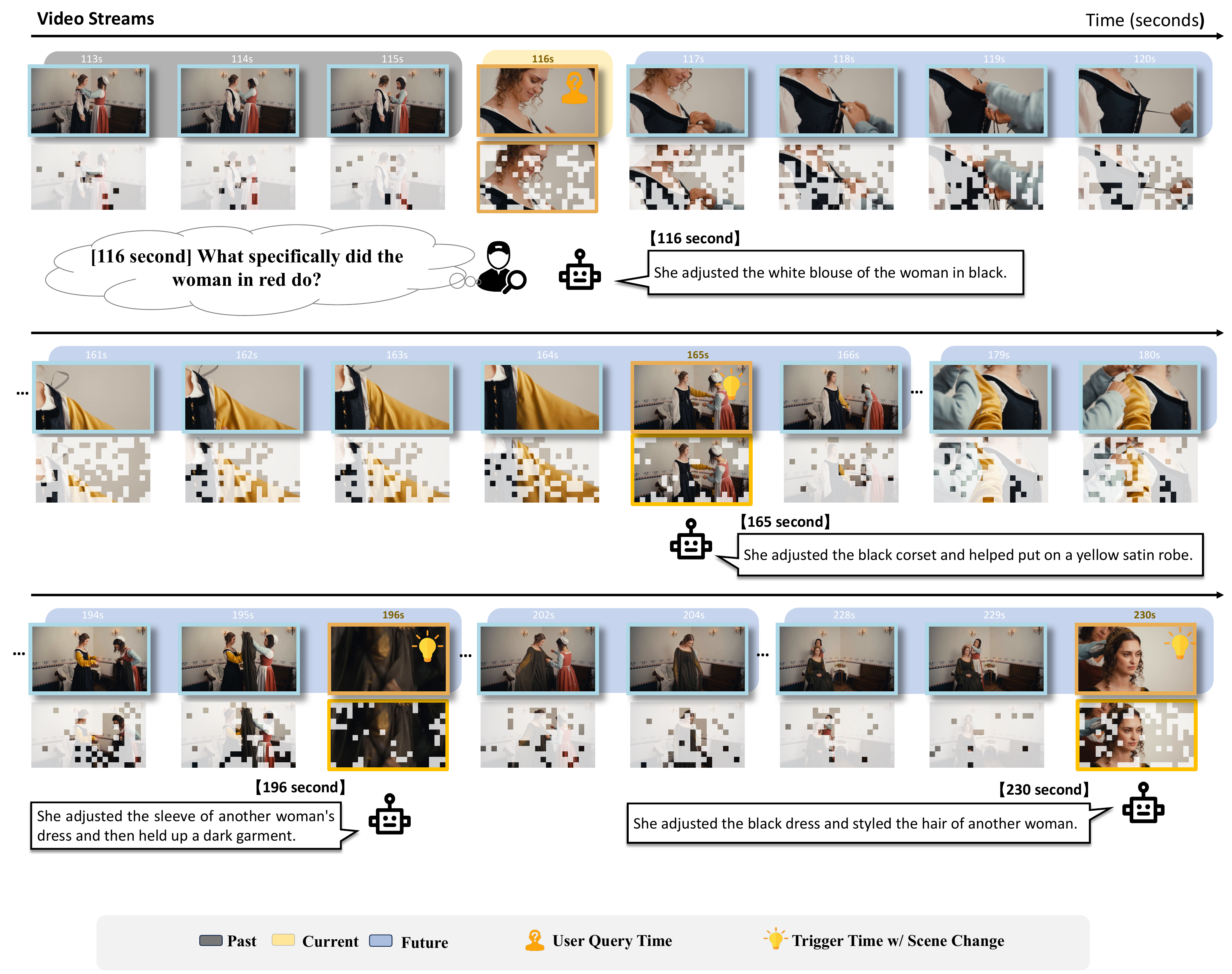}
    \caption{Case study of \model on StreamingBench. When a user proposes a question ``What specifically did the woman in red do?'' that can also be answered by the future moments, \model will proactively generate responses at the future trigger time (i.e., the video scene transition timestamps), which are indicated by the frames with low token drop ratios.}
    \label{fig:streamingbench_case}
\end{figure*}

\subsection{Ablation Study}

\noindent\textbf{Effectiveness of \module Design.}
We compare different token dropping methods in Table~\ref{tab:abla_drop}. VisionZip~\cite{visionzip} represents a similar purely-vision spatial token selection method, while pixel-level and feature-level dropping approaches are introduced in this paper (Section~\ref{sec:redunt_calculate}). Zero-shot results with Qwen2.5-VL-7B demonstrate that \textit{Feature-level (video-aware) Dropping}, i.e., the final design of \module, achieves the best performance under the same dropping ratio. The \textit{frame-aware} dropping approach applies a fixed dropping ratio to each frame, in contrast to the \textit{video-aware} approach that dynamically selects tokens across the entire video.
These results reveal that joint spatio-temporal dynamic token pruning proves most effective.

\noindent\textbf{Effectiveness of \dataset dataset.} 
Table~\ref{tab:abla_dataset} demonstrates the impact of our training strategy that combines online and offline datasets. The results clearly show that integrating \dataset with existing offline VideoQA datasets significantly enhances streaming performance.

\vspace{-5pt}
\section{Analysis of \module}

\noindent \textbf{Performance-Drop Ratio Tradeoff.}
As shown in Figure~\ref{fig:teaser}(right top), feature-level dropping consistently outperforms pixel-level dropping. Meanwhile, training with \module can make VideoLLMs better adapt to dropped token distribution and perform better on extreme dropping ratio such as 80\%+ token dropping. Overall, the \module approach maintains robust performance even with substantial token reduction. At an extreme 82.8\% drop ratio, the model achieves comparable results to the full token setting (73.64 vs. 73.70). 

\noindent\textbf{Redundancy Across Different Video Lengths.}
Figure~\ref{fig:combined_analysis} demonstrates that longer videos contain higher redundancy, permitting more aggressive token dropping. While short videos show obvious performance drops at high dropping rates, long videos (30-60 minutes) maintain performance integrity even with extreme dropping rates up to 97.5\%. 

\noindent\textbf{Efficiency Improvements.}
Figure~\ref{fig:teaser}(right down) quantifies the computational benefits of token dropping. At 81.1\% drop ratio, inference latency decreases from 3220ms to 1820ms (1.76$\times$ faster) while accuracy improves from 50.4\% to 56.3\%. This suggests that focusing on fewer, more informative tokens not only accelerates inference but can enhance model performance by reducing visual noise.

\section{Conclusion}
This paper introduces \model, a novel online \videollm that incorporates Differential Token Dropping, a simple yet efficient approach addressing long-form high-redundancy challenges in streaming video understanding. Comprehensive experiments demonstrate that \model achieves state-of-the-art performance on streaming video benchmarks while eliminating up to 82.8\% of visual tokens. Additionally, our analysis reveals that reducing temporal redundancy is particularly critical for hours-long videos, which can eliminate over 95\% of tokens via the \module strategy without performance degradation. 
For real-time interaction in streaming scenarios, we also propose a synthetic \dataset to endow \model with backward-tracing, real-time perception, and proactive future-responding  capabilities.
This work not only highlights the substantial redundancy in video streams but also establishes a promising direction for computationally efficient video encoding in streaming VideoQA tasks.

\bibliographystyle{ACM-Reference-Format}
\bibliography{sample-base}


\begin{thebibliography}{83}


\ifx \showCODEN    \undefined \def \showCODEN     #1{\unskip}     \fi
\ifx \showISBNx    \undefined \def \showISBNx     #1{\unskip}     \fi
\ifx \showISBNxiii \undefined \def \showISBNxiii  #1{\unskip}     \fi
\ifx \showISSN     \undefined \def \showISSN      #1{\unskip}     \fi
\ifx \showLCCN     \undefined \def \showLCCN      #1{\unskip}     \fi
\ifx \shownote     \undefined \def \shownote      #1{#1}          \fi
\ifx \showarticletitle \undefined \def \showarticletitle #1{#1}   \fi
\ifx \showURL      \undefined \def \showURL       {\relax}        \fi
\providecommand\bibfield[2]{#2}
\providecommand\bibinfo[2]{#2}
\providecommand\natexlab[1]{#1}
\providecommand\showeprint[2][]{arXiv:#2}

\bibitem[Aggarwal et~al\mbox{.}(2001)]%
        {l1}
\bibfield{author}{\bibinfo{person}{Charu~C Aggarwal}, \bibinfo{person}{Alexander Hinneburg}, {and} \bibinfo{person}{Daniel~A Keim}.} \bibinfo{year}{2001}\natexlab{}.
\newblock \showarticletitle{On the surprising behavior of distance metrics in high dimensional space}. In \bibinfo{booktitle}{\emph{International conference on database theory}}. Springer, \bibinfo{pages}{420--434}.
\newblock


\bibitem[{Anthropic}(2024)]%
        {claude3_5}
\bibfield{author}{\bibinfo{person}{{Anthropic}}.} \bibinfo{year}{2024}\natexlab{}.
\newblock \bibinfo{title}{Claude 3.5 Sonnet}.
\newblock
\urldef\tempurl%
\url{https://www.anthropic.com/news/claude-3-5-sonnet}
\showURL{%
\tempurl}


\bibitem[Ataallah et~al\mbox{.}(2024)]%
        {minigpt4video}
\bibfield{author}{\bibinfo{person}{Kirolos Ataallah}, \bibinfo{person}{Xiaoqian Shen}, \bibinfo{person}{Eslam Abdelrahman}, \bibinfo{person}{Essam Sleiman}, \bibinfo{person}{Deyao Zhu}, \bibinfo{person}{Jian Ding}, {and} \bibinfo{person}{Mohamed Elhoseiny}.} \bibinfo{year}{2024}\natexlab{}.
\newblock \showarticletitle{MiniGPT4-Video: Advancing Multimodal LLMs for Video Understanding with Interleaved Visual-Textual Tokens}.
\newblock \bibinfo{journal}{\emph{ArXiv preprint}}  \bibinfo{volume}{abs/2404.03413} (\bibinfo{year}{2024}).
\newblock


\bibitem[Bai et~al\mbox{.}(2023)]%
        {Qwen-VL}
\bibfield{author}{\bibinfo{person}{Jinze Bai}, \bibinfo{person}{Shuai Bai}, \bibinfo{person}{Shusheng Yang}, \bibinfo{person}{Shijie Wang}, \bibinfo{person}{Sinan Tan}, \bibinfo{person}{Peng Wang}, \bibinfo{person}{Junyang Lin}, \bibinfo{person}{Chang Zhou}, {and} \bibinfo{person}{Jingren Zhou}.} \bibinfo{year}{2023}\natexlab{}.
\newblock \showarticletitle{Qwen-VL: A Frontier Large Vision-Language Model with Versatile Abilities}.
\newblock \bibinfo{journal}{\emph{ArXiv preprint}}  \bibinfo{volume}{abs/2308.12966} (\bibinfo{year}{2023}).
\newblock


\bibitem[Bai et~al\mbox{.}(2025)]%
        {qwen2.5vl}
\bibfield{author}{\bibinfo{person}{Shuai Bai}, \bibinfo{person}{Keqin Chen}, \bibinfo{person}{Xuejing Liu}, \bibinfo{person}{Jialin Wang}, \bibinfo{person}{Wenbin Ge}, \bibinfo{person}{Sibo Song}, \bibinfo{person}{Kai Dang}, \bibinfo{person}{Peng Wang}, \bibinfo{person}{Shijie Wang}, \bibinfo{person}{Jun Tang}, \bibinfo{person}{Humen Zhong}, \bibinfo{person}{Yuanzhi Zhu}, \bibinfo{person}{Mingkun Yang}, \bibinfo{person}{Zhaohai Li}, \bibinfo{person}{Jianqiang Wan}, \bibinfo{person}{Pengfei Wang}, \bibinfo{person}{Wei Ding}, \bibinfo{person}{Zheren Fu}, \bibinfo{person}{Yiheng Xu}, \bibinfo{person}{Jiabo Ye}, \bibinfo{person}{Xi Zhang}, \bibinfo{person}{Tianbao Xie}, \bibinfo{person}{Zesen Cheng}, \bibinfo{person}{Hang Zhang}, \bibinfo{person}{Zhibo Yang}, \bibinfo{person}{Haiyang Xu}, {and} \bibinfo{person}{Junyang Lin}.} \bibinfo{year}{2025}\natexlab{}.
\newblock \showarticletitle{Qwen2.5-VL Technical Report}.
\newblock \bibinfo{journal}{\emph{arXiv preprint arXiv:2502.13923}} (\bibinfo{year}{2025}).
\newblock


\bibitem[Chen et~al\mbox{.}(2024a)]%
        {videollm-online}
\bibfield{author}{\bibinfo{person}{Joya Chen}, \bibinfo{person}{Zhaoyang Lv}, \bibinfo{person}{Shiwei Wu}, \bibinfo{person}{Kevin~Qinghong Lin}, \bibinfo{person}{Chenan Song}, \bibinfo{person}{Difei Gao}, \bibinfo{person}{Jia-Wei Liu}, \bibinfo{person}{Ziteng Gao}, \bibinfo{person}{Dongxing Mao}, {and} \bibinfo{person}{Mike~Zheng Shou}.} \bibinfo{year}{2024}\natexlab{a}.
\newblock \showarticletitle{Videollm-online: Online video large language model for streaming video}. In \bibinfo{booktitle}{\emph{Proceedings of the IEEE/CVF Conference on Computer Vision and Pattern Recognition}}. \bibinfo{pages}{18407--18418}.
\newblock


\bibitem[Chen et~al\mbox{.}(2024c)]%
        {fastv}
\bibfield{author}{\bibinfo{person}{Liang Chen}, \bibinfo{person}{Haozhe Zhao}, \bibinfo{person}{Tianyu Liu}, \bibinfo{person}{Shuai Bai}, \bibinfo{person}{Junyang Lin}, \bibinfo{person}{Chang Zhou}, {and} \bibinfo{person}{Baobao Chang}.} \bibinfo{year}{2024}\natexlab{c}.
\newblock \bibinfo{title}{An Image is Worth 1/2 Tokens After Layer 2: Plug-and-Play Inference Acceleration for Large Vision-Language Models}.
\newblock
\showeprint[arxiv]{2403.06764}~[cs.CV]


\bibitem[Chen et~al\mbox{.}(2024b)]%
        {internvl1.5}
\bibfield{author}{\bibinfo{person}{Zhe Chen}, \bibinfo{person}{Weiyun Wang}, \bibinfo{person}{Hao Tian}, \bibinfo{person}{Shenglong Ye}, \bibinfo{person}{Zhangwei Gao}, \bibinfo{person}{Erfei Cui}, \bibinfo{person}{Wenwen Tong}, \bibinfo{person}{Kongzhi Hu}, \bibinfo{person}{Jiapeng Luo}, \bibinfo{person}{Zheng Ma}, \bibinfo{person}{Ji Ma}, \bibinfo{person}{Jiaqi Wang}, \bibinfo{person}{Xiaoyi Dong}, \bibinfo{person}{Hang Yan}, \bibinfo{person}{Hewei Guo}, \bibinfo{person}{Conghui He}, \bibinfo{person}{Botian Shi}, \bibinfo{person}{Zhenjiang Jin}, \bibinfo{person}{Chao Xu}, \bibinfo{person}{Bin Wang}, \bibinfo{person}{Xingjian Wei}, \bibinfo{person}{Wei Li}, \bibinfo{person}{Wenjian Zhang}, \bibinfo{person}{Bo Zhang}, \bibinfo{person}{Pinlong Cai}, \bibinfo{person}{Licheng Wen}, \bibinfo{person}{Xiangchao Yan}, \bibinfo{person}{Min Dou}, \bibinfo{person}{Lewei Lu}, \bibinfo{person}{Xizhou Zhu}, \bibinfo{person}{Tong Lu}, \bibinfo{person}{Dahua Lin}, \bibinfo{person}{Yu Qiao}, \bibinfo{person}{Jifeng Dai}, {and} \bibinfo{person}{Wenhai Wang}.} \bibinfo{year}{2024}\natexlab{b}.
\newblock \bibinfo{title}{How Far Are We to GPT-4V? Closing the Gap to Commercial Multimodal Models with Open-Source Suites}.
\newblock
\showeprint[arxiv]{2404.16821}~[cs.CV]
\urldef\tempurl%
\url{https://arxiv.org/abs/2404.16821}
\showURL{%
\tempurl}


\bibitem[Cheng et~al\mbox{.}(2024)]%
        {videollama2}
\bibfield{author}{\bibinfo{person}{Zesen Cheng}, \bibinfo{person}{Sicong Leng}, \bibinfo{person}{Hang Zhang}, \bibinfo{person}{Yifei Xin}, \bibinfo{person}{Xin Li}, \bibinfo{person}{Guanzheng Chen}, \bibinfo{person}{Yongxin Zhu}, \bibinfo{person}{Wenqi Zhang}, \bibinfo{person}{Ziyang Luo}, \bibinfo{person}{Deli Zhao}, {et~al\mbox{.}}} \bibinfo{year}{2024}\natexlab{}.
\newblock \showarticletitle{Videollama 2: Advancing spatial-temporal modeling and audio understanding in video-llms}.
\newblock \bibinfo{journal}{\emph{arXiv preprint arXiv:2406.07476}} (\bibinfo{year}{2024}).
\newblock


\bibitem[Choudhury et~al\mbox{.}(2024)]%
        {rlt}
\bibfield{author}{\bibinfo{person}{Rohan Choudhury}, \bibinfo{person}{Guanglei Zhu}, \bibinfo{person}{Sihan Liu}, \bibinfo{person}{Koichiro Niinuma}, \bibinfo{person}{Kris Kitani}, {and} \bibinfo{person}{L{\'a}szl{\'o} Jeni}.} \bibinfo{year}{2024}\natexlab{}.
\newblock \showarticletitle{Don't Look Twice: Faster Video Transformers with Run-Length Tokenization}.
\newblock \bibinfo{journal}{\emph{Advances in Neural Information Processing Systems}}  \bibinfo{volume}{37} (\bibinfo{year}{2024}), \bibinfo{pages}{28127--28149}.
\newblock


\bibitem[Dai et~al\mbox{.}(2023)]%
        {dai2023instructblip}
\bibfield{author}{\bibinfo{person}{Wenliang Dai}, \bibinfo{person}{Junnan Li}, \bibinfo{person}{Dongxu Li}, \bibinfo{person}{Anthony Meng~Huat Tiong}, \bibinfo{person}{Junqi Zhao}, \bibinfo{person}{Weisheng Wang}, \bibinfo{person}{Boyang Li}, \bibinfo{person}{Pascale Fung}, {and} \bibinfo{person}{Steven C.~H. Hoi}.} \bibinfo{year}{2023}\natexlab{}.
\newblock \showarticletitle{InstructBLIP: Towards General-purpose Vision-Language Models with Instruction Tuning}. In \bibinfo{booktitle}{\emph{Advances in Neural Information Processing Systems 36: Annual Conference on Neural Information Processing Systems 2023, NeurIPS 2023, New Orleans, LA, USA, December 10 - 16, 2023}}, \bibfield{editor}{\bibinfo{person}{Alice Oh}, \bibinfo{person}{Tristan Naumann}, \bibinfo{person}{Amir Globerson}, \bibinfo{person}{Kate Saenko}, \bibinfo{person}{Moritz Hardt}, {and} \bibinfo{person}{Sergey Levine}} (Eds.).
\newblock


\bibitem[Di et~al\mbox{.}(2025)]%
        {ReKV}
\bibfield{author}{\bibinfo{person}{Shangzhe Di}, \bibinfo{person}{Zhelun Yu}, \bibinfo{person}{Guanghao Zhang}, \bibinfo{person}{Haoyuan Li}, \bibinfo{person}{Tao Zhong}, \bibinfo{person}{Hao Cheng}, \bibinfo{person}{Bolin Li}, \bibinfo{person}{Wanggui He}, \bibinfo{person}{Fangxun Shu}, {and} \bibinfo{person}{Hao Jiang}.} \bibinfo{year}{2025}\natexlab{}.
\newblock \showarticletitle{Streaming video question-answering with in-context video kv-cache retrieval}.
\newblock \bibinfo{journal}{\emph{arXiv preprint arXiv:2503.00540}} (\bibinfo{year}{2025}).
\newblock


\bibitem[Dosovitskiy et~al\mbox{.}(2020)]%
        {vit}
\bibfield{author}{\bibinfo{person}{Alexey Dosovitskiy}, \bibinfo{person}{Lucas Beyer}, \bibinfo{person}{Alexander Kolesnikov}, \bibinfo{person}{Dirk Weissenborn}, \bibinfo{person}{Xiaohua Zhai}, \bibinfo{person}{Thomas Unterthiner}, \bibinfo{person}{Mostafa Dehghani}, \bibinfo{person}{Matthias Minderer}, \bibinfo{person}{G Heigold}, \bibinfo{person}{S Gelly}, {et~al\mbox{.}}} \bibinfo{year}{2020}\natexlab{}.
\newblock \showarticletitle{An Image is Worth 16x16 Words: Transformers for Image Recognition at Scale}. In \bibinfo{booktitle}{\emph{International Conference on Learning Representations}}.
\newblock


\bibitem[Dubey et~al\mbox{.}(2024)]%
        {llama3}
\bibfield{author}{\bibinfo{person}{Abhimanyu Dubey}, \bibinfo{person}{Abhinav Jauhri}, \bibinfo{person}{Abhinav Pandey}, \bibinfo{person}{Abhishek Kadian}, \bibinfo{person}{Ahmad Al{-}Dahle}, \bibinfo{person}{Aiesha Letman}, \bibinfo{person}{Akhil Mathur}, \bibinfo{person}{Alan Schelten}, \bibinfo{person}{Amy Yang}, \bibinfo{person}{Angela Fan}, \bibinfo{person}{Anirudh Goyal}, \bibinfo{person}{Anthony Hartshorn}, \bibinfo{person}{Aobo Yang}, \bibinfo{person}{Archi Mitra}, \bibinfo{person}{Archie Sravankumar}, \bibinfo{person}{Artem Korenev}, \bibinfo{person}{Arthur Hinsvark}, \bibinfo{person}{Arun Rao}, \bibinfo{person}{Aston Zhang}, \bibinfo{person}{Aur{\'{e}}lien Rodriguez}, \bibinfo{person}{Austen Gregerson}, \bibinfo{person}{Ava Spataru}, \bibinfo{person}{Baptiste Rozi{\`{e}}re}, \bibinfo{person}{Bethany Biron}, \bibinfo{person}{Binh Tang}, \bibinfo{person}{Bobbie Chern}, \bibinfo{person}{Charlotte Caucheteux}, \bibinfo{person}{Chaya Nayak}, \bibinfo{person}{Chloe Bi}, \bibinfo{person}{Chris Marra}, \bibinfo{person}{Chris McConnell}, \bibinfo{person}{Christian Keller}, \bibinfo{person}{Christophe Touret}, \bibinfo{person}{Chunyang Wu}, \bibinfo{person}{Corinne Wong}, \bibinfo{person}{Cristian~Canton Ferrer}, \bibinfo{person}{Cyrus Nikolaidis}, \bibinfo{person}{Damien Allonsius}, \bibinfo{person}{Daniel Song}, \bibinfo{person}{Danielle Pintz}, \bibinfo{person}{Danny Livshits}, \bibinfo{person}{David Esiobu}, \bibinfo{person}{Dhruv Choudhary}, \bibinfo{person}{Dhruv Mahajan}, \bibinfo{person}{Diego Garcia{-}Olano}, \bibinfo{person}{Diego Perino}, \bibinfo{person}{Dieuwke Hupkes}, \bibinfo{person}{Egor Lakomkin}, \bibinfo{person}{Ehab AlBadawy}, \bibinfo{person}{Elina Lobanova}, \bibinfo{person}{Emily Dinan}, \bibinfo{person}{Eric~Michael Smith}, \bibinfo{person}{Filip Radenovic}, \bibinfo{person}{Frank Zhang}, \bibinfo{person}{Gabriel Synnaeve}, \bibinfo{person}{Gabrielle Lee}, \bibinfo{person}{Georgia~Lewis Anderson}, \bibinfo{person}{Graeme Nail}, \bibinfo{person}{Gr{\'{e}}goire Mialon}, \bibinfo{person}{Guan Pang}, \bibinfo{person}{Guillem Cucurell}, \bibinfo{person}{Hailey Nguyen}, \bibinfo{person}{Hannah Korevaar}, \bibinfo{person}{Hu Xu}, \bibinfo{person}{Hugo Touvron}, \bibinfo{person}{Iliyan Zarov}, \bibinfo{person}{Imanol~Arrieta Ibarra}, \bibinfo{person}{Isabel~M. Kloumann}, \bibinfo{person}{Ishan Misra}, \bibinfo{person}{Ivan Evtimov}, \bibinfo{person}{Jade Copet}, \bibinfo{person}{Jaewon Lee}, \bibinfo{person}{Jan Geffert}, \bibinfo{person}{Jana Vranes}, \bibinfo{person}{Jason Park}, \bibinfo{person}{Jay Mahadeokar}, \bibinfo{person}{Jeet Shah}, \bibinfo{person}{Jelmer van~der Linde}, \bibinfo{person}{Jennifer Billock}, \bibinfo{person}{Jenny Hong}, \bibinfo{person}{Jenya Lee}, \bibinfo{person}{Jeremy Fu}, \bibinfo{person}{Jianfeng Chi}, \bibinfo{person}{Jianyu Huang}, \bibinfo{person}{Jiawen Liu}, \bibinfo{person}{Jie Wang}, \bibinfo{person}{Jiecao Yu}, \bibinfo{person}{Joanna Bitton}, \bibinfo{person}{Joe Spisak}, \bibinfo{person}{Jongsoo Park}, \bibinfo{person}{Joseph Rocca}, \bibinfo{person}{Joshua Johnstun}, \bibinfo{person}{Joshua Saxe}, \bibinfo{person}{Junteng Jia}, \bibinfo{person}{Kalyan~Vasuden Alwala}, \bibinfo{person}{Kartikeya Upasani}, \bibinfo{person}{Kate Plawiak}, \bibinfo{person}{Ke Li}, \bibinfo{person}{Kenneth Heafield}, \bibinfo{person}{Kevin Stone}, {and} \bibinfo{person}{et al.}} \bibinfo{year}{2024}\natexlab{}.
\newblock \showarticletitle{The Llama 3 Herd of Models}.
\newblock \bibinfo{journal}{\emph{ArXiv preprint}}  \bibinfo{volume}{abs/2407.21783} (\bibinfo{year}{2024}).
\newblock


\bibitem[Fabian Caba~Heilbron and Niebles(2015)]%
        {activitynet}
\bibfield{author}{\bibinfo{person}{Bernard~Ghanem Fabian Caba~Heilbron, Victor~Escorcia} {and} \bibinfo{person}{Juan~Carlos Niebles}.} \bibinfo{year}{2015}\natexlab{}.
\newblock \showarticletitle{ActivityNet: A Large-Scale Video Benchmark for Human Activity Understanding}. In \bibinfo{booktitle}{\emph{Proceedings of the IEEE Conference on Computer Vision and Pattern Recognition}}. \bibinfo{pages}{961--970}.
\newblock


\bibitem[Fei et~al\mbox{.}(2024)]%
        {videoccam}
\bibfield{author}{\bibinfo{person}{Jiajun Fei}, \bibinfo{person}{Dian Li}, \bibinfo{person}{Zhidong Deng}, \bibinfo{person}{Zekun Wang}, \bibinfo{person}{Gang Liu}, {and} \bibinfo{person}{Hui Wang}.} \bibinfo{year}{2024}\natexlab{}.
\newblock \showarticletitle{Video-ccam: Enhancing video-language understanding with causal cross-attention masks for short and long videos}.
\newblock \bibinfo{journal}{\emph{arXiv preprint arXiv:2408.14023}} (\bibinfo{year}{2024}).
\newblock


\bibitem[Fu et~al\mbox{.}(2024)]%
        {videomme}
\bibfield{author}{\bibinfo{person}{Chaoyou Fu}, \bibinfo{person}{Yuhan Dai}, \bibinfo{person}{Yondong Luo}, \bibinfo{person}{Lei Li}, \bibinfo{person}{Shuhuai Ren}, \bibinfo{person}{Renrui Zhang}, \bibinfo{person}{Zihan Wang}, \bibinfo{person}{Chenyu Zhou}, \bibinfo{person}{Yunhang Shen}, \bibinfo{person}{Mengdan Zhang}, {et~al\mbox{.}}} \bibinfo{year}{2024}\natexlab{}.
\newblock \showarticletitle{Video-MME: The First-Ever Comprehensive Evaluation Benchmark of Multi-modal LLMs in Video Analysis}.
\newblock \bibinfo{journal}{\emph{ArXiv preprint}}  \bibinfo{volume}{abs/2405.21075} (\bibinfo{year}{2024}).
\newblock


\bibitem[{Gemini Team}(2024)]%
        {gemini}
\bibfield{author}{\bibinfo{person}{{Gemini Team}}.} \bibinfo{year}{2024}\natexlab{}.
\newblock \showarticletitle{Gemini 1.5: Unlocking multimodal understanding across millions of tokens of context}.
\newblock \bibinfo{journal}{\emph{ArXiv preprint}}  \bibinfo{volume}{abs/2403.05530} (\bibinfo{year}{2024}).
\newblock


\bibitem[Hu et~al\mbox{.}(2024)]%
        {minicpm}
\bibfield{author}{\bibinfo{person}{Shengding Hu}, \bibinfo{person}{Yuge Tu}, \bibinfo{person}{Xu Han}, \bibinfo{person}{Chaoqun He}, \bibinfo{person}{Ganqu Cui}, \bibinfo{person}{Xiang Long}, \bibinfo{person}{Zhi Zheng}, \bibinfo{person}{Yewei Fang}, \bibinfo{person}{Yuxiang Huang}, \bibinfo{person}{Weilin Zhao}, {et~al\mbox{.}}} \bibinfo{year}{2024}\natexlab{}.
\newblock \showarticletitle{Minicpm: Unveiling the potential of small language models with scalable training strategies}.
\newblock \bibinfo{journal}{\emph{arXiv preprint arXiv:2404.06395}} (\bibinfo{year}{2024}).
\newblock


\bibitem[Huang et~al\mbox{.}(2020)]%
        {vitt}
\bibfield{author}{\bibinfo{person}{Gabriel Huang}, \bibinfo{person}{Bo Pang}, \bibinfo{person}{Zhenhai Zhu}, \bibinfo{person}{Clara Rivera}, {and} \bibinfo{person}{Radu Soricut}.} \bibinfo{year}{2020}\natexlab{}.
\newblock \bibinfo{title}{Multimodal Pretraining for Dense Video Captioning}.
\newblock
\showeprint[arxiv]{2011.11760}~[cs.CV]
\urldef\tempurl%
\url{https://arxiv.org/abs/2011.11760}
\showURL{%
\tempurl}


\bibitem[Huang et~al\mbox{.}(2024)]%
        {videochat-online}
\bibfield{author}{\bibinfo{person}{Zhenpeng Huang}, \bibinfo{person}{Xinhao Li}, \bibinfo{person}{Jiaqi Li}, \bibinfo{person}{Jing Wang}, \bibinfo{person}{Xiangyu Zeng}, \bibinfo{person}{Cheng Liang}, \bibinfo{person}{Tao Wu}, \bibinfo{person}{Xi Chen}, \bibinfo{person}{Liang Li}, {and} \bibinfo{person}{Limin Wang}.} \bibinfo{year}{2024}\natexlab{}.
\newblock \showarticletitle{Online video understanding: A comprehensive benchmark and memory-augmented method}.
\newblock \bibinfo{journal}{\emph{arXiv preprint arXiv:2501.00584}} (\bibinfo{year}{2024}).
\newblock


\bibitem[Jin et~al\mbox{.}(2024)]%
        {chat-univi}
\bibfield{author}{\bibinfo{person}{Peng Jin}, \bibinfo{person}{Ryuichi Takanobu}, \bibinfo{person}{Wancai Zhang}, \bibinfo{person}{Xiaochun Cao}, {and} \bibinfo{person}{Li Yuan}.} \bibinfo{year}{2024}\natexlab{}.
\newblock \showarticletitle{Chat-univi: Unified visual representation empowers large language models with image and video understanding}. In \bibinfo{booktitle}{\emph{Proceedings of the IEEE/CVF Conference on Computer Vision and Pattern Recognition}}. \bibinfo{pages}{13700--13710}.
\newblock


\bibitem[Lei et~al\mbox{.}(2021)]%
        {qvhighlights}
\bibfield{author}{\bibinfo{person}{Jie Lei}, \bibinfo{person}{Tamara~L. Berg}, {and} \bibinfo{person}{Mohit Bansal}.} \bibinfo{year}{2021}\natexlab{}.
\newblock \bibinfo{title}{QVHighlights: Detecting Moments and Highlights in Videos via Natural Language Queries}.
\newblock
\showeprint[arxiv]{2107.09609}~[cs.CV]
\urldef\tempurl%
\url{https://arxiv.org/abs/2107.09609}
\showURL{%
\tempurl}


\bibitem[Li et~al\mbox{.}(2024e)]%
        {llava-onevision}
\bibfield{author}{\bibinfo{person}{Bo Li}, \bibinfo{person}{Yuanhan Zhang}, \bibinfo{person}{Dong Guo}, \bibinfo{person}{Renrui Zhang}, \bibinfo{person}{Feng Li}, \bibinfo{person}{Hao Zhang}, \bibinfo{person}{Kaichen Zhang}, \bibinfo{person}{Yanwei Li}, \bibinfo{person}{Ziwei Liu}, {and} \bibinfo{person}{Chunyuan Li}.} \bibinfo{year}{2024}\natexlab{e}.
\newblock \showarticletitle{LLaVA-OneVision: Easy Visual Task Transfer}.
\newblock \bibinfo{journal}{\emph{ArXiv preprint}}  \bibinfo{volume}{abs/2408.03326} (\bibinfo{year}{2024}).
\newblock


\bibitem[Li et~al\mbox{.}(2025a)]%
        {llava-st}
\bibfield{author}{\bibinfo{person}{Hongyu Li}, \bibinfo{person}{Jinyu Chen}, \bibinfo{person}{Ziyu Wei}, \bibinfo{person}{Shaofei Huang}, \bibinfo{person}{Tianrui Hui}, \bibinfo{person}{Jialin Gao}, \bibinfo{person}{Xiaoming Wei}, {and} \bibinfo{person}{Si Liu}.} \bibinfo{year}{2025}\natexlab{a}.
\newblock \showarticletitle{LLaVA-ST: A Multimodal Large Language Model for Fine-Grained Spatial-Temporal Understanding}.
\newblock \bibinfo{journal}{\emph{arXiv preprint arXiv:2501.08282}} (\bibinfo{year}{2025}).
\newblock


\bibitem[Li et~al\mbox{.}(2023)]%
        {2023videochat}
\bibfield{author}{\bibinfo{person}{Kunchang Li}, \bibinfo{person}{Yinan He}, \bibinfo{person}{Yi Wang}, \bibinfo{person}{Yizhuo Li}, \bibinfo{person}{Wenhai Wang}, \bibinfo{person}{Ping Luo}, \bibinfo{person}{Yali Wang}, \bibinfo{person}{Limin Wang}, {and} \bibinfo{person}{Yu Qiao}.} \bibinfo{year}{2023}\natexlab{}.
\newblock \showarticletitle{VideoChat: Chat-Centric Video Understanding}.
\newblock \bibinfo{journal}{\emph{ArXiv preprint}}  \bibinfo{volume}{abs/2305.06355} (\bibinfo{year}{2023}).
\newblock


\bibitem[Li et~al\mbox{.}(2025b)]%
        {videot3}
\bibfield{author}{\bibinfo{person}{Lei Li}, \bibinfo{person}{Yuanxin Liu}, \bibinfo{person}{Linli Yao}, \bibinfo{person}{Peiyuan Zhang}, \bibinfo{person}{Chenxin An}, \bibinfo{person}{Lean Wang}, \bibinfo{person}{Xu Sun}, \bibinfo{person}{Lingpeng Kong}, {and} \bibinfo{person}{Qi Liu}.} \bibinfo{year}{2025}\natexlab{b}.
\newblock \showarticletitle{Temporal Reasoning Transfer from Text to Video}. In \bibinfo{booktitle}{\emph{ICLR 2025}}. \bibinfo{publisher}{OpenReview.net}.
\newblock
\urldef\tempurl%
\url{https://openreview.net/forum?id=sHAvMp5J4R}
\showURL{%
\tempurl}


\bibitem[Li et~al\mbox{.}(2024c)]%
        {videochat-flash}
\bibfield{author}{\bibinfo{person}{Xinhao Li}, \bibinfo{person}{Yi Wang}, \bibinfo{person}{Jiashuo Yu}, \bibinfo{person}{Xiangyu Zeng}, \bibinfo{person}{Yuhan Zhu}, \bibinfo{person}{Haian Huang}, \bibinfo{person}{Jianfei Gao}, \bibinfo{person}{Kunchang Li}, \bibinfo{person}{Yinan He}, \bibinfo{person}{Chenting Wang}, {et~al\mbox{.}}} \bibinfo{year}{2024}\natexlab{c}.
\newblock \showarticletitle{Videochat-flash: Hierarchical compression for long-context video modeling}.
\newblock \bibinfo{journal}{\emph{arXiv preprint arXiv:2501.00574}} (\bibinfo{year}{2024}).
\newblock


\bibitem[Li et~al\mbox{.}(2024d)]%
        {videochat-falsh}
\bibfield{author}{\bibinfo{person}{Xinhao Li}, \bibinfo{person}{Yi Wang}, \bibinfo{person}{Jiashuo Yu}, \bibinfo{person}{Xiangyu Zeng}, \bibinfo{person}{Yuhan Zhu}, \bibinfo{person}{Haian Huang}, \bibinfo{person}{Jianfei Gao}, \bibinfo{person}{Kunchang Li}, \bibinfo{person}{Yinan He}, \bibinfo{person}{Chenting Wang}, {et~al\mbox{.}}} \bibinfo{year}{2024}\natexlab{d}.
\newblock \showarticletitle{Videochat-flash: Hierarchical compression for long-context video modeling}.
\newblock \bibinfo{journal}{\emph{arXiv preprint arXiv:2501.00574}} (\bibinfo{year}{2024}).
\newblock


\bibitem[Li et~al\mbox{.}(2025c)]%
        {ovobench}
\bibfield{author}{\bibinfo{person}{Yifei Li}, \bibinfo{person}{Junbo Niu}, \bibinfo{person}{Ziyang Miao}, \bibinfo{person}{Chunjiang Ge}, \bibinfo{person}{Yuanhang Zhou}, \bibinfo{person}{Qihao He}, \bibinfo{person}{Xiaoyi Dong}, \bibinfo{person}{Haodong Duan}, \bibinfo{person}{Shuangrui Ding}, \bibinfo{person}{Rui Qian}, \bibinfo{person}{Pan Zhang}, \bibinfo{person}{Yuhang Zang}, \bibinfo{person}{Yuhang Cao}, \bibinfo{person}{Conghui He}, {and} \bibinfo{person}{Jiaqi Wang}.} \bibinfo{year}{2025}\natexlab{c}.
\newblock \bibinfo{title}{OVO-Bench: How Far is Your Video-LLMs from Real-World Online Video Understanding?}
\newblock
\showeprint[arxiv]{2501.05510}~[cs.CV]
\urldef\tempurl%
\url{https://arxiv.org/abs/2501.05510}
\showURL{%
\tempurl}


\bibitem[Li et~al\mbox{.}(2024a)]%
        {llama-vid}
\bibfield{author}{\bibinfo{person}{Yanwei Li}, \bibinfo{person}{Chengyao Wang}, {and} \bibinfo{person}{Jiaya Jia}.} \bibinfo{year}{2024}\natexlab{a}.
\newblock \showarticletitle{Llama-vid: An image is worth 2 tokens in large language models}. In \bibinfo{booktitle}{\emph{European Conference on Computer Vision}}. Springer, \bibinfo{pages}{323--340}.
\newblock


\bibitem[Li et~al\mbox{.}(2024b)]%
        {li2024llamavid}
\bibfield{author}{\bibinfo{person}{Yanwei Li}, \bibinfo{person}{Chengyao Wang}, {and} \bibinfo{person}{Jiaya Jia}.} \bibinfo{year}{2024}\natexlab{b}.
\newblock \showarticletitle{Llama-vid: An image is worth 2 tokens in large language models}. In \bibinfo{booktitle}{\emph{European Conference on Computer Vision}}. Springer, \bibinfo{pages}{323--340}.
\newblock


\bibitem[Lin et~al\mbox{.}(2024a)]%
        {streamingbench}
\bibfield{author}{\bibinfo{person}{Junming Lin}, \bibinfo{person}{Zheng Fang}, \bibinfo{person}{Chi Chen}, \bibinfo{person}{Zihao Wan}, \bibinfo{person}{Fuwen Luo}, \bibinfo{person}{Peng Li}, \bibinfo{person}{Yang Liu}, {and} \bibinfo{person}{Maosong Sun}.} \bibinfo{year}{2024}\natexlab{a}.
\newblock \showarticletitle{StreamingBench: Assessing the Gap for MLLMs to Achieve Streaming Video Understanding}.
\newblock \bibinfo{journal}{\emph{arXiv preprint arXiv:2411.03628}} (\bibinfo{year}{2024}).
\newblock


\bibitem[Lin et~al\mbox{.}(2024b)]%
        {vila}
\bibfield{author}{\bibinfo{person}{Ji Lin}, \bibinfo{person}{Hongxu Yin}, \bibinfo{person}{Wei Ping}, \bibinfo{person}{Pavlo Molchanov}, \bibinfo{person}{Mohammad Shoeybi}, {and} \bibinfo{person}{Song Han}.} \bibinfo{year}{2024}\natexlab{b}.
\newblock \showarticletitle{Vila: On pre-training for visual language models}. In \bibinfo{booktitle}{\emph{Proceedings of the IEEE/CVF conference on computer vision and pattern recognition}}. \bibinfo{pages}{26689--26699}.
\newblock


\bibitem[Liu et~al\mbox{.}(2024a)]%
        {llava-next}
\bibfield{author}{\bibinfo{person}{Haotian Liu}, \bibinfo{person}{Chunyuan Li}, \bibinfo{person}{Yuheng Li}, \bibinfo{person}{Bo Li}, \bibinfo{person}{Yuanhan Zhang}, \bibinfo{person}{Sheng Shen}, {and} \bibinfo{person}{Yong~Jae Lee}.} \bibinfo{year}{2024}\natexlab{a}.
\newblock \bibinfo{title}{LLaVA-NeXT: Improved reasoning, OCR, and world knowledge}.
\newblock


\bibitem[Liu et~al\mbox{.}(2024c)]%
        {kangaroo}
\bibfield{author}{\bibinfo{person}{Jiajun Liu}, \bibinfo{person}{Yibing Wang}, \bibinfo{person}{Hanghang Ma}, \bibinfo{person}{Xiaoping Wu}, \bibinfo{person}{Xiaoqi Ma}, \bibinfo{person}{xiaoming Wei}, \bibinfo{person}{Jianbin Jiao}, \bibinfo{person}{Enhua Wu}, {and} \bibinfo{person}{Jie Hu}.} \bibinfo{year}{2024}\natexlab{c}.
\newblock \showarticletitle{Kangaroo: A Powerful Video-Language Model Supporting Long-context Video Input}.
\newblock \bibinfo{journal}{\emph{arXiv preprint arXiv:2408.15542}} (\bibinfo{year}{2024}).
\newblock


\bibitem[Liu et~al\mbox{.}(2024b)]%
        {etbench}
\bibfield{author}{\bibinfo{person}{Ye Liu}, \bibinfo{person}{Zongyang Ma}, \bibinfo{person}{Zhongang Qi}, \bibinfo{person}{Yang Wu}, \bibinfo{person}{Ying Shan}, {and} \bibinfo{person}{Chang~W Chen}.} \bibinfo{year}{2024}\natexlab{b}.
\newblock \showarticletitle{Et bench: Towards open-ended event-level video-language understanding}.
\newblock \bibinfo{journal}{\emph{Advances in Neural Information Processing Systems}}  \bibinfo{volume}{37} (\bibinfo{year}{2024}), \bibinfo{pages}{32076--32110}.
\newblock


\bibitem[Liu et~al\mbox{.}(2025)]%
        {liu2025hybrid}
\bibfield{author}{\bibinfo{person}{Zhihang Liu}, \bibinfo{person}{Chen-Wei Xie}, \bibinfo{person}{Pandeng Li}, \bibinfo{person}{Liming Zhao}, \bibinfo{person}{Longxiang Tang}, \bibinfo{person}{Yun Zheng}, \bibinfo{person}{Chuanbin Liu}, {and} \bibinfo{person}{Hongtao Xie}.} \bibinfo{year}{2025}\natexlab{}.
\newblock \showarticletitle{Hybrid-Level Instruction Injection for Video Token Compression in Multi-modal Large Language Models}.
\newblock \bibinfo{journal}{\emph{arXiv preprint arXiv:2503.16036}} (\bibinfo{year}{2025}).
\newblock


\bibitem[Ning et~al\mbox{.}(2024)]%
        {inf-MLLM}
\bibfield{author}{\bibinfo{person}{Zhenyu Ning}, \bibinfo{person}{Jieru Zhao}, \bibinfo{person}{Qihao Jin}, \bibinfo{person}{Wenchao Ding}, {and} \bibinfo{person}{Minyi Guo}.} \bibinfo{year}{2024}\natexlab{}.
\newblock \showarticletitle{Inf-MLLM: Efficient streaming inference of multimodal large language models on a single GPU}.
\newblock \bibinfo{journal}{\emph{arXiv preprint arXiv:2409.09086}} (\bibinfo{year}{2024}).
\newblock


\bibitem[Oncescu et~al\mbox{.}(2021)]%
        {queryd}
\bibfield{author}{\bibinfo{person}{Andreea-Maria Oncescu}, \bibinfo{person}{Joao~F Henriques}, \bibinfo{person}{Yang Liu}, \bibinfo{person}{Andrew Zisserman}, {and} \bibinfo{person}{Samuel Albanie}.} \bibinfo{year}{2021}\natexlab{}.
\newblock \showarticletitle{Queryd: A video dataset with high-quality text and audio narrations}. In \bibinfo{booktitle}{\emph{ICASSP 2021-2021 IEEE International Conference on Acoustics, Speech and Signal Processing (ICASSP)}}. IEEE, \bibinfo{pages}{2265--2269}.
\newblock


\bibitem[OpenAI(2024)]%
        {gpt4o}
\bibfield{author}{\bibinfo{person}{OpenAI}.} \bibinfo{year}{2024}\natexlab{}.
\newblock \bibinfo{title}{GPT-4o System Card}.
\newblock


\bibitem[Oquab et~al\mbox{.}(2023)]%
        {dinov2}
\bibfield{author}{\bibinfo{person}{Maxime Oquab}, \bibinfo{person}{Timoth{\'e}e Darcet}, \bibinfo{person}{Th{\'e}o Moutakanni}, \bibinfo{person}{Huy Vo}, \bibinfo{person}{Marc Szafraniec}, \bibinfo{person}{Vasil Khalidov}, \bibinfo{person}{Pierre Fernandez}, \bibinfo{person}{Daniel Haziza}, \bibinfo{person}{Francisco Massa}, \bibinfo{person}{Alaaeldin El-Nouby}, {et~al\mbox{.}}} \bibinfo{year}{2023}\natexlab{}.
\newblock \showarticletitle{Dinov2: Learning robust visual features without supervision}.
\newblock \bibinfo{journal}{\emph{arXiv preprint arXiv:2304.07193}} (\bibinfo{year}{2023}).
\newblock


\bibitem[Qian et~al\mbox{.}(2025)]%
        {dispider}
\bibfield{author}{\bibinfo{person}{Rui Qian}, \bibinfo{person}{Shuangrui Ding}, \bibinfo{person}{Xiaoyi Dong}, \bibinfo{person}{Pan Zhang}, \bibinfo{person}{Yuhang Zang}, \bibinfo{person}{Yuhang Cao}, \bibinfo{person}{Dahua Lin}, {and} \bibinfo{person}{Jiaqi Wang}.} \bibinfo{year}{2025}\natexlab{}.
\newblock \showarticletitle{Dispider: Enabling Video LLMs with Active Real-Time Interaction via Disentangled Perception, Decision, and Reaction}.
\newblock \bibinfo{journal}{\emph{arXiv preprint arXiv:2501.03218}} (\bibinfo{year}{2025}).
\newblock


\bibitem[Qian et~al\mbox{.}(2024)]%
        {videostreaming}
\bibfield{author}{\bibinfo{person}{Rui Qian}, \bibinfo{person}{Xiaoyi Dong}, \bibinfo{person}{Pan Zhang}, \bibinfo{person}{Yuhang Zang}, \bibinfo{person}{Shuangrui Ding}, \bibinfo{person}{Dahua Lin}, {and} \bibinfo{person}{Jiaqi Wang}.} \bibinfo{year}{2024}\natexlab{}.
\newblock \showarticletitle{Streaming long video understanding with large language models}.
\newblock \bibinfo{journal}{\emph{Advances in Neural Information Processing Systems}}  \bibinfo{volume}{37} (\bibinfo{year}{2024}), \bibinfo{pages}{119336--119360}.
\newblock


\bibitem[Ren et~al\mbox{.}(2023)]%
        {Ren2023TESTA}
\bibfield{author}{\bibinfo{person}{Shuhuai Ren}, \bibinfo{person}{Sishuo Chen}, \bibinfo{person}{Shicheng Li}, \bibinfo{person}{Xu Sun}, {and} \bibinfo{person}{Lu Hou}.} \bibinfo{year}{2023}\natexlab{}.
\newblock \showarticletitle{TESTA: Temporal-Spatial Token Aggregation for Long-form Video-Language Understanding}.
\newblock \bibinfo{journal}{\emph{ArXiv}}  \bibinfo{volume}{abs/2310.19060} (\bibinfo{year}{2023}).
\newblock


\bibitem[Ren et~al\mbox{.}(2024)]%
        {timechat}
\bibfield{author}{\bibinfo{person}{Shuhuai Ren}, \bibinfo{person}{Linli Yao}, \bibinfo{person}{Shicheng Li}, \bibinfo{person}{Xu Sun}, {and} \bibinfo{person}{Lu Hou}.} \bibinfo{year}{2024}\natexlab{}.
\newblock \showarticletitle{Timechat: A time-sensitive multimodal large language model for long video understanding}. In \bibinfo{booktitle}{\emph{Proceedings of the IEEE/CVF Conference on Computer Vision and Pattern Recognition}}. \bibinfo{pages}{14313--14323}.
\newblock


\bibitem[Rensink(2002)]%
        {changedetection}
\bibfield{author}{\bibinfo{person}{Ronald~A. Rensink}.} \bibinfo{year}{2002}\natexlab{}.
\newblock \showarticletitle{Change detection}.
\newblock \bibinfo{journal}{\emph{Annual Review of Psychology}}  \bibinfo{volume}{53} (\bibinfo{year}{2002}), \bibinfo{pages}{245--277}.
\newblock


\bibitem[Shen et~al\mbox{.}(2024)]%
        {longvu}
\bibfield{author}{\bibinfo{person}{Xiaoqian Shen}, \bibinfo{person}{Yunyang Xiong}, \bibinfo{person}{Changsheng Zhao}, \bibinfo{person}{Lemeng Wu}, \bibinfo{person}{Jun Chen}, \bibinfo{person}{Chenchen Zhu}, \bibinfo{person}{Zechun Liu}, \bibinfo{person}{Fanyi Xiao}, \bibinfo{person}{Balakrishnan Varadarajan}, \bibinfo{person}{Florian Bordes}, {et~al\mbox{.}}} \bibinfo{year}{2024}\natexlab{}.
\newblock \showarticletitle{Longvu: Spatiotemporal adaptive compression for long video-language understanding}.
\newblock \bibinfo{journal}{\emph{arXiv preprint arXiv:2410.17434}} (\bibinfo{year}{2024}).
\newblock


\bibitem[Simons and Rensink(2005)]%
        {change-blindness}
\bibfield{author}{\bibinfo{person}{Daniel~J Simons} {and} \bibinfo{person}{Ronald~A Rensink}.} \bibinfo{year}{2005}\natexlab{}.
\newblock \showarticletitle{Change blindness: Past, present, and future}.
\newblock \bibinfo{journal}{\emph{Trends in cognitive sciences}} \bibinfo{volume}{9}, \bibinfo{number}{1} (\bibinfo{year}{2005}), \bibinfo{pages}{16--20}.
\newblock


\bibitem[Song et~al\mbox{.}(2015)]%
        {tvsum}
\bibfield{author}{\bibinfo{person}{Yale Song}, \bibinfo{person}{Jordi Vallmitjana}, \bibinfo{person}{Amanda Stent}, {and} \bibinfo{person}{Alejandro Jaimes}.} \bibinfo{year}{2015}\natexlab{}.
\newblock \showarticletitle{TVSum: Summarizing web videos using titles}. In \bibinfo{booktitle}{\emph{2015 IEEE Conference on Computer Vision and Pattern Recognition (CVPR)}}. \bibinfo{pages}{5179--5187}.
\newblock
\href{https://doi.org/10.1109/CVPR.2015.7299154}{doi:\nolinkurl{10.1109/CVPR.2015.7299154}}


\bibitem[Steck et~al\mbox{.}(2024)]%
        {steck2024cosine}
\bibfield{author}{\bibinfo{person}{Harald Steck}, \bibinfo{person}{Chaitanya Ekanadham}, {and} \bibinfo{person}{Nathan Kallus}.} \bibinfo{year}{2024}\natexlab{}.
\newblock \showarticletitle{Is cosine-similarity of embeddings really about similarity?}. In \bibinfo{booktitle}{\emph{Companion Proceedings of the ACM Web Conference 2024}}. \bibinfo{pages}{887--890}.
\newblock


\bibitem[Tang et~al\mbox{.}(2019)]%
        {coin}
\bibfield{author}{\bibinfo{person}{Yansong Tang}, \bibinfo{person}{Dajun Ding}, \bibinfo{person}{Yongming Rao}, \bibinfo{person}{Yu Zheng}, \bibinfo{person}{Danyang Zhang}, \bibinfo{person}{Lili Zhao}, \bibinfo{person}{Jiwen Lu}, {and} \bibinfo{person}{Jie Zhou}.} \bibinfo{year}{2019}\natexlab{}.
\newblock \bibinfo{title}{COIN: A Large-scale Dataset for Comprehensive Instructional Video Analysis}.
\newblock
\showeprint[arxiv]{1903.02874}~[cs.CV]
\urldef\tempurl%
\url{https://arxiv.org/abs/1903.02874}
\showURL{%
\tempurl}


\bibitem[Tao et~al\mbox{.}(2024)]%
        {dycoke}
\bibfield{author}{\bibinfo{person}{Keda Tao}, \bibinfo{person}{Can Qin}, \bibinfo{person}{Haoxuan You}, \bibinfo{person}{Yang Sui}, {and} \bibinfo{person}{Huan Wang}.} \bibinfo{year}{2024}\natexlab{}.
\newblock \showarticletitle{DyCoke: Dynamic Compression of Tokens for Fast Video Large Language Models}.
\newblock \bibinfo{journal}{\emph{arXiv preprint arXiv:2411.15024}} (\bibinfo{year}{2024}).
\newblock


\bibitem[Wang et~al\mbox{.}(2024a)]%
        {Qwen2VL}
\bibfield{author}{\bibinfo{person}{Peng Wang}, \bibinfo{person}{Shuai Bai}, \bibinfo{person}{Sinan Tan}, \bibinfo{person}{Shijie Wang}, \bibinfo{person}{Zhihao Fan}, \bibinfo{person}{Jinze Bai}, \bibinfo{person}{Keqin Chen}, \bibinfo{person}{Xuejing Liu}, \bibinfo{person}{Jialin Wang}, \bibinfo{person}{Wenbin Ge}, \bibinfo{person}{Yang Fan}, \bibinfo{person}{Kai Dang}, \bibinfo{person}{Mengfei Du}, \bibinfo{person}{Xuancheng Ren}, \bibinfo{person}{Rui Men}, \bibinfo{person}{Dayiheng Liu}, \bibinfo{person}{Chang Zhou}, \bibinfo{person}{Jingren Zhou}, {and} \bibinfo{person}{Junyang Lin}.} \bibinfo{year}{2024}\natexlab{a}.
\newblock \showarticletitle{Qwen2-VL: Enhancing Vision-Language Model's Perception of the World at Any Resolution}.
\newblock \bibinfo{journal}{\emph{arXiv preprint arXiv:2409.12191}} (\bibinfo{year}{2024}).
\newblock


\bibitem[Wang et~al\mbox{.}(2020)]%
        {videoic}
\bibfield{author}{\bibinfo{person}{Weiying Wang}, \bibinfo{person}{Jieting Chen}, {and} \bibinfo{person}{Qin Jin}.} \bibinfo{year}{2020}\natexlab{}.
\newblock \showarticletitle{VideoIC: A Video Interactive Comments Dataset and Multimodal Multitask Learning for Comments Generation}. In \bibinfo{booktitle}{\emph{Proceedings of the 28th ACM International Conference on Multimedia}} (Seattle, WA, USA) \emph{(\bibinfo{series}{MM '20})}. \bibinfo{publisher}{Association for Computing Machinery}, \bibinfo{address}{New York, NY, USA}, \bibinfo{pages}{2599–2607}.
\newblock
\showISBNx{9781450379885}
\href{https://doi.org/10.1145/3394171.3413890}{doi:\nolinkurl{10.1145/3394171.3413890}}


\bibitem[Wang et~al\mbox{.}(2019)]%
        {youmakeup}
\bibfield{author}{\bibinfo{person}{Weiying Wang}, \bibinfo{person}{Yongcheng Wang}, \bibinfo{person}{Shizhe Chen}, {and} \bibinfo{person}{Qin Jin}.} \bibinfo{year}{2019}\natexlab{}.
\newblock \showarticletitle{YouMakeup: A Large-Scale Domain-Specific Multimodal Dataset for Fine-Grained Semantic Comprehension}. In \bibinfo{booktitle}{\emph{Proceedings of the 2019 Conference on Empirical Methods in Natural Language Processing and the 9th International Joint Conference on Natural Language Processing (EMNLP-IJCNLP)}}. \bibinfo{pages}{5136--5146}.
\newblock


\bibitem[Wang et~al\mbox{.}(2024b)]%
        {mmduet}
\bibfield{author}{\bibinfo{person}{Yueqian Wang}, \bibinfo{person}{Xiaojun Meng}, \bibinfo{person}{Yuxuan Wang}, \bibinfo{person}{Jianxin Liang}, \bibinfo{person}{Jiansheng Wei}, \bibinfo{person}{Huishuai Zhang}, {and} \bibinfo{person}{Dongyan Zhao}.} \bibinfo{year}{2024}\natexlab{b}.
\newblock \bibinfo{title}{VideoLLM Knows When to Speak: Enhancing Time-Sensitive Video Comprehension with Video-Text Duet Interaction Format}.
\newblock
\showeprint[arxiv]{2411.17991}~[cs.CV]
\urldef\tempurl%
\url{https://arxiv.org/abs/2411.17991}
\showURL{%
\tempurl}


\bibitem[Wen et~al\mbox{.}(2025)]%
        {wen2025stop}
\bibfield{author}{\bibinfo{person}{Zichen Wen}, \bibinfo{person}{Yifeng Gao}, \bibinfo{person}{Shaobo Wang}, \bibinfo{person}{Junyuan Zhang}, \bibinfo{person}{Qintong Zhang}, \bibinfo{person}{Weijia Li}, \bibinfo{person}{Conghui He}, {and} \bibinfo{person}{Linfeng Zhang}.} \bibinfo{year}{2025}\natexlab{}.
\newblock \showarticletitle{Stop Looking for Important Tokens in Multimodal Language Models: Duplication Matters More}.
\newblock \bibinfo{journal}{\emph{arXiv preprint arXiv:2502.11494}} (\bibinfo{year}{2025}).
\newblock


\bibitem[Wu et~al\mbox{.}(2024b)]%
        {longvideobench}
\bibfield{author}{\bibinfo{person}{Haoning Wu}, \bibinfo{person}{Dongxu Li}, \bibinfo{person}{Bei Chen}, {and} \bibinfo{person}{Junnan Li}.} \bibinfo{year}{2024}\natexlab{b}.
\newblock \showarticletitle{LongVideoBench: {A} Benchmark for Long-context Interleaved Video-Language Understanding}.
\newblock \bibinfo{journal}{\emph{ArXiv preprint}}  \bibinfo{volume}{abs/2407.15754} (\bibinfo{year}{2024}).
\newblock


\bibitem[Wu et~al\mbox{.}(2024a)]%
        {videollm-mod}
\bibfield{author}{\bibinfo{person}{Shiwei Wu}, \bibinfo{person}{Joya Chen}, \bibinfo{person}{Kevin~Qinghong Lin}, \bibinfo{person}{Qimeng Wang}, \bibinfo{person}{Yan Gao}, \bibinfo{person}{Qianli Xu}, \bibinfo{person}{Tong Xu}, \bibinfo{person}{Yao Hu}, \bibinfo{person}{Enhong Chen}, {and} \bibinfo{person}{Mike~Zheng Shou}.} \bibinfo{year}{2024}\natexlab{a}.
\newblock \showarticletitle{Videollm-mod: Efficient video-language streaming with mixture-of-depths vision computation}.
\newblock \bibinfo{journal}{\emph{Advances in Neural Information Processing Systems}}  \bibinfo{volume}{37} (\bibinfo{year}{2024}), \bibinfo{pages}{109922--109947}.
\newblock


\bibitem[Xiong et~al\mbox{.}(2025)]%
        {streamchat}
\bibfield{author}{\bibinfo{person}{Haomiao Xiong}, \bibinfo{person}{Zongxin Yang}, \bibinfo{person}{Jiazuo Yu}, \bibinfo{person}{Yunzhi Zhuge}, \bibinfo{person}{Lu Zhang}, \bibinfo{person}{Jiawen Zhu}, {and} \bibinfo{person}{Huchuan Lu}.} \bibinfo{year}{2025}\natexlab{}.
\newblock \showarticletitle{Streaming Video Understanding and Multi-round Interaction with Memory-enhanced Knowledge}.
\newblock \bibinfo{journal}{\emph{arXiv preprint arXiv:2501.13468}} (\bibinfo{year}{2025}).
\newblock


\bibitem[Xu et~al\mbox{.}(2025)]%
        {qwen2.5-omni}
\bibfield{author}{\bibinfo{person}{Jin Xu}, \bibinfo{person}{Zhifang Guo}, \bibinfo{person}{Jinzheng He}, \bibinfo{person}{Hangrui Hu}, \bibinfo{person}{Ting He}, \bibinfo{person}{Shuai Bai}, \bibinfo{person}{Keqin Chen}, \bibinfo{person}{Jialin Wang}, \bibinfo{person}{Yang Fan}, \bibinfo{person}{Kai Dang}, {et~al\mbox{.}}} \bibinfo{year}{2025}\natexlab{}.
\newblock \showarticletitle{Qwen2. 5-Omni Technical Report}.
\newblock \bibinfo{journal}{\emph{arXiv preprint arXiv:2503.20215}} (\bibinfo{year}{2025}).
\newblock


\bibitem[Xu et~al\mbox{.}(2024)]%
        {slowfast-llava}
\bibfield{author}{\bibinfo{person}{Mingze Xu}, \bibinfo{person}{Mingfei Gao}, \bibinfo{person}{Zhe Gan}, \bibinfo{person}{Hong-You Chen}, \bibinfo{person}{Zhengfeng Lai}, \bibinfo{person}{Haiming Gang}, \bibinfo{person}{Kai Kang}, {and} \bibinfo{person}{Afshin Dehghan}.} \bibinfo{year}{2024}\natexlab{}.
\newblock \showarticletitle{Slowfast-llava: A strong training-free baseline for video large language models}.
\newblock \bibinfo{journal}{\emph{arXiv preprint arXiv:2407.15841}} (\bibinfo{year}{2024}).
\newblock


\bibitem[Xue et~al\mbox{.}(2022)]%
        {hd-vila}
\bibfield{author}{\bibinfo{person}{Hongwei Xue}, \bibinfo{person}{Tiankai Hang}, \bibinfo{person}{Yanhong Zeng}, \bibinfo{person}{Yuchong Sun}, \bibinfo{person}{Bei Liu}, \bibinfo{person}{Huan Yang}, \bibinfo{person}{Jianlong Fu}, {and} \bibinfo{person}{Baining Guo}.} \bibinfo{year}{2022}\natexlab{}.
\newblock \bibinfo{title}{Advancing High-Resolution Video-Language Representation with Large-Scale Video Transcriptions}.
\newblock
\showeprint[arxiv]{2111.10337}~[cs.CV]
\urldef\tempurl%
\url{https://arxiv.org/abs/2111.10337}
\showURL{%
\tempurl}


\bibitem[Yang et~al\mbox{.}(2024c)]%
        {qwen2}
\bibfield{author}{\bibinfo{person}{An Yang}, \bibinfo{person}{Baosong Yang}, \bibinfo{person}{Binyuan Hui}, \bibinfo{person}{Bo Zheng}, \bibinfo{person}{Bowen Yu}, \bibinfo{person}{Chang Zhou}, \bibinfo{person}{Chengpeng Li}, \bibinfo{person}{Chengyuan Li}, \bibinfo{person}{Dayiheng Liu}, \bibinfo{person}{Fei Huang}, \bibinfo{person}{Guanting Dong}, \bibinfo{person}{Haoran Wei}, \bibinfo{person}{Huan Lin}, \bibinfo{person}{Jialong Tang}, \bibinfo{person}{Jialin Wang}, \bibinfo{person}{Jian Yang}, \bibinfo{person}{Jianhong Tu}, \bibinfo{person}{Jianwei Zhang}, \bibinfo{person}{Jianxin Ma}, \bibinfo{person}{Jin Xu}, \bibinfo{person}{Jingren Zhou}, \bibinfo{person}{Jinze Bai}, \bibinfo{person}{Jinzheng He}, \bibinfo{person}{Junyang Lin}, \bibinfo{person}{Kai Dang}, \bibinfo{person}{Keming Lu}, \bibinfo{person}{Keqin Chen}, \bibinfo{person}{Kexin Yang}, \bibinfo{person}{Mei Li}, \bibinfo{person}{Mingfeng Xue}, \bibinfo{person}{Na Ni}, \bibinfo{person}{Pei Zhang}, \bibinfo{person}{Peng Wang}, \bibinfo{person}{Ru Peng}, \bibinfo{person}{Rui Men}, \bibinfo{person}{Ruize Gao}, \bibinfo{person}{Runji Lin}, \bibinfo{person}{Shijie Wang}, \bibinfo{person}{Shuai Bai}, \bibinfo{person}{Sinan Tan}, \bibinfo{person}{Tianhang Zhu}, \bibinfo{person}{Tianhao Li}, \bibinfo{person}{Tianyu Liu}, \bibinfo{person}{Wenbin Ge}, \bibinfo{person}{Xiaodong Deng}, \bibinfo{person}{Xiaohuan Zhou}, \bibinfo{person}{Xingzhang Ren}, \bibinfo{person}{Xinyu Zhang}, \bibinfo{person}{Xipin Wei}, \bibinfo{person}{Xuancheng Ren}, \bibinfo{person}{Yang Fan}, \bibinfo{person}{Yang Yao}, \bibinfo{person}{Yichang Zhang}, \bibinfo{person}{Yu Wan}, \bibinfo{person}{Yunfei Chu}, \bibinfo{person}{Yuqiong Liu}, \bibinfo{person}{Zeyu Cui}, \bibinfo{person}{Zhenru Zhang}, {and} \bibinfo{person}{Zhihao Fan}.} \bibinfo{year}{2024}\natexlab{c}.
\newblock \showarticletitle{Qwen2 Technical Report}.
\newblock \bibinfo{journal}{\emph{ArXiv preprint}}  \bibinfo{volume}{abs/2407.10671} (\bibinfo{year}{2024}).
\newblock


\bibitem[Yang et~al\mbox{.}(2024b)]%
        {pvc}
\bibfield{author}{\bibinfo{person}{Chenyu Yang}, \bibinfo{person}{Xuan Dong}, \bibinfo{person}{Xizhou Zhu}, \bibinfo{person}{Weijie Su}, \bibinfo{person}{Jiahao Wang}, \bibinfo{person}{Hao Tian}, \bibinfo{person}{Zhe Chen}, \bibinfo{person}{Wenhai Wang}, \bibinfo{person}{Lewei Lu}, {and} \bibinfo{person}{Jifeng Dai}.} \bibinfo{year}{2024}\natexlab{b}.
\newblock \showarticletitle{PVC: Progressive Visual Token Compression for Unified Image and Video Processing in Large Vision-Language Models}.
\newblock \bibinfo{journal}{\emph{arXiv preprint arXiv:2412.09613}} (\bibinfo{year}{2024}).
\newblock


\bibitem[Yang et~al\mbox{.}(2025)]%
        {yang2025topv}
\bibfield{author}{\bibinfo{person}{Cheng Yang}, \bibinfo{person}{Yang Sui}, \bibinfo{person}{Jinqi Xiao}, \bibinfo{person}{Lingyi Huang}, \bibinfo{person}{Yu Gong}, \bibinfo{person}{Chendi Li}, \bibinfo{person}{Jinghua Yan}, \bibinfo{person}{Yu Bai}, \bibinfo{person}{Ponnuswamy Sadayappan}, \bibinfo{person}{Xia Hu}, {et~al\mbox{.}}} \bibinfo{year}{2025}\natexlab{}.
\newblock \showarticletitle{TopV: Compatible Token Pruning with Inference Time Optimization for Fast and Low-Memory Multimodal Vision Language Model}.
\newblock \bibinfo{journal}{\emph{arXiv preprint arXiv:2503.18278}} (\bibinfo{year}{2025}).
\newblock


\bibitem[Yang et~al\mbox{.}(2024a)]%
        {visionzip}
\bibfield{author}{\bibinfo{person}{Senqiao Yang}, \bibinfo{person}{Yukang Chen}, \bibinfo{person}{Zhuotao Tian}, \bibinfo{person}{Chengyao Wang}, \bibinfo{person}{Jingyao Li}, \bibinfo{person}{Bei Yu}, {and} \bibinfo{person}{Jiaya Jia}.} \bibinfo{year}{2024}\natexlab{a}.
\newblock \showarticletitle{Visionzip: Longer is better but not necessary in vision language models}.
\newblock \bibinfo{journal}{\emph{arXiv preprint arXiv:2412.04467}} (\bibinfo{year}{2024}).
\newblock


\bibitem[Yao et~al\mbox{.}(2024)]%
        {deco}
\bibfield{author}{\bibinfo{person}{Linli Yao}, \bibinfo{person}{Lei Li}, \bibinfo{person}{Shuhuai Ren}, \bibinfo{person}{Lean Wang}, \bibinfo{person}{Yuanxin Liu}, \bibinfo{person}{Xu Sun}, {and} \bibinfo{person}{Lu Hou}.} \bibinfo{year}{2024}\natexlab{}.
\newblock \showarticletitle{Deco: Decoupling token compression from semantic abstraction in multimodal large language models}.
\newblock \bibinfo{journal}{\emph{arXiv preprint arXiv:2405.20985}} (\bibinfo{year}{2024}).
\newblock


\bibitem[Yuan et~al\mbox{.}(2025)]%
        {tarsier2}
\bibfield{author}{\bibinfo{person}{Liping Yuan}, \bibinfo{person}{Jiawei Wang}, \bibinfo{person}{Haomiao Sun}, \bibinfo{person}{Yuchen Zhang}, {and} \bibinfo{person}{Yuan Lin}.} \bibinfo{year}{2025}\natexlab{}.
\newblock \showarticletitle{Tarsier2: Advancing Large Vision-Language Models from Detailed Video Description to Comprehensive Video Understanding}.
\newblock \bibinfo{journal}{\emph{arXiv preprint arXiv:2501.07888}} (\bibinfo{year}{2025}).
\newblock


\bibitem[Yue et~al\mbox{.}(2023)]%
        {movie101}
\bibfield{author}{\bibinfo{person}{Zihao Yue}, \bibinfo{person}{Qi Zhang}, \bibinfo{person}{Anwen Hu}, \bibinfo{person}{Liang Zhang}, \bibinfo{person}{Ziheng Wang}, {and} \bibinfo{person}{Qin Jin}.} \bibinfo{year}{2023}\natexlab{}.
\newblock \bibinfo{title}{Movie101: A New Movie Understanding Benchmark}.
\newblock
\showeprint[arxiv]{2305.12140}~[cs.CV]
\urldef\tempurl%
\url{https://arxiv.org/abs/2305.12140}
\showURL{%
\tempurl}


\bibitem[Zala et~al\mbox{.}(2023)]%
        {hirest}
\bibfield{author}{\bibinfo{person}{Abhay Zala}, \bibinfo{person}{Jaemin Cho}, \bibinfo{person}{Satwik Kottur}, \bibinfo{person}{Xilun Chen}, \bibinfo{person}{Barlas Oğuz}, \bibinfo{person}{Yashar Mehdad}, {and} \bibinfo{person}{Mohit Bansal}.} \bibinfo{year}{2023}\natexlab{}.
\newblock \showarticletitle{Hierarchical Video-Moment Retrieval and Step-Captioning}. In \bibinfo{booktitle}{\emph{CVPR}}.
\newblock


\bibitem[Zhang et~al\mbox{.}(2025)]%
        {zhang2025videollama3}
\bibfield{author}{\bibinfo{person}{Boqiang Zhang}, \bibinfo{person}{Kehan Li}, \bibinfo{person}{Zesen Cheng}, \bibinfo{person}{Zhiqiang Hu}, \bibinfo{person}{Yuqian Yuan}, \bibinfo{person}{Guanzheng Chen}, \bibinfo{person}{Sicong Leng}, \bibinfo{person}{Yuming Jiang}, \bibinfo{person}{Hang Zhang}, \bibinfo{person}{Xin Li}, {et~al\mbox{.}}} \bibinfo{year}{2025}\natexlab{}.
\newblock \showarticletitle{VideoLLaMA 3: Frontier Multimodal Foundation Models for Image and Video Understanding}.
\newblock \bibinfo{journal}{\emph{arXiv preprint arXiv:2501.13106}} (\bibinfo{year}{2025}).
\newblock


\bibitem[Zhang et~al\mbox{.}(2023a)]%
        {videollama}
\bibfield{author}{\bibinfo{person}{Hang Zhang}, \bibinfo{person}{Xin Li}, {and} \bibinfo{person}{Lidong Bing}.} \bibinfo{year}{2023}\natexlab{a}.
\newblock \showarticletitle{Video-{LL}a{MA}: An Instruction-tuned Audio-Visual Language Model for Video Understanding}. In \bibinfo{booktitle}{\emph{Proceedings of the 2023 Conference on Empirical Methods in Natural Language Processing: System Demonstrations}}, \bibfield{editor}{\bibinfo{person}{Yansong Feng} {and} \bibinfo{person}{Els Lefever}} (Eds.). \bibinfo{pages}{543--553}.
\newblock
\href{https://doi.org/10.18653/v1/2023.emnlp-demo.49}{doi:\nolinkurl{10.18653/v1/2023.emnlp-demo.49}}


\bibitem[Zhang et~al\mbox{.}(2024d)]%
        {flashvstream}
\bibfield{author}{\bibinfo{person}{Haoji Zhang}, \bibinfo{person}{Yiqin Wang}, \bibinfo{person}{Yansong Tang}, \bibinfo{person}{Yong Liu}, \bibinfo{person}{Jiashi Feng}, \bibinfo{person}{Jifeng Dai}, {and} \bibinfo{person}{Xiaojie Jin}.} \bibinfo{year}{2024}\natexlab{d}.
\newblock \showarticletitle{Flash-vstream: Memory-based real-time understanding for long video streams}.
\newblock \bibinfo{journal}{\emph{arXiv preprint arXiv:2406.08085}} (\bibinfo{year}{2024}).
\newblock


\bibitem[Zhang et~al\mbox{.}(2024a)]%
        {zhang2024vinoground}
\bibfield{author}{\bibinfo{person}{Jianrui Zhang}, \bibinfo{person}{Mu Cai}, {and} \bibinfo{person}{Yong~Jae Lee}.} \bibinfo{year}{2024}\natexlab{a}.
\newblock \showarticletitle{Vinoground: Scrutinizing LMMs over Dense Temporal Reasoning with Short Videos}.
\newblock \bibinfo{journal}{\emph{arXiv preprint arXiv:2410.02763}} (\bibinfo{year}{2024}).
\newblock


\bibitem[Zhang et~al\mbox{.}(2024b)]%
        {internlm-xcomposer2.5-omnilive}
\bibfield{author}{\bibinfo{person}{Pan Zhang}, \bibinfo{person}{Xiaoyi Dong}, \bibinfo{person}{Yuhang Cao}, \bibinfo{person}{Yuhang Zang}, \bibinfo{person}{Rui Qian}, \bibinfo{person}{Xilin Wei}, \bibinfo{person}{Lin Chen}, \bibinfo{person}{Yifei Li}, \bibinfo{person}{Junbo Niu}, \bibinfo{person}{Shuangrui Ding}, {et~al\mbox{.}}} \bibinfo{year}{2024}\natexlab{b}.
\newblock \showarticletitle{Internlm-xcomposer2. 5-omnilive: A comprehensive multimodal system for long-term streaming video and audio interactions}.
\newblock \bibinfo{journal}{\emph{arXiv preprint arXiv:2412.09596}} (\bibinfo{year}{2024}).
\newblock


\bibitem[Zhang et~al\mbox{.}(2024c)]%
        {internlmxcomposer2d5-omni}
\bibfield{author}{\bibinfo{person}{Pan Zhang}, \bibinfo{person}{Xiaoyi Dong}, \bibinfo{person}{Yuhang Cao}, \bibinfo{person}{Yuhang Zang}, \bibinfo{person}{Rui Qian}, \bibinfo{person}{Xilin Wei}, \bibinfo{person}{Lin Chen}, \bibinfo{person}{Yifei Li}, \bibinfo{person}{Junbo Niu}, \bibinfo{person}{Shuangrui Ding}, \bibinfo{person}{Qipeng Guo}, \bibinfo{person}{Haodong Duan}, \bibinfo{person}{Xin Chen}, \bibinfo{person}{Han Lv}, \bibinfo{person}{Zheng Nie}, \bibinfo{person}{Min Zhang}, \bibinfo{person}{Bin Wang}, \bibinfo{person}{Wenwei Zhang}, \bibinfo{person}{Xinyue Zhang}, \bibinfo{person}{Jiaye Ge}, \bibinfo{person}{Wei Li}, \bibinfo{person}{Jingwen Li}, \bibinfo{person}{Zhongying Tu}, \bibinfo{person}{Conghui He}, \bibinfo{person}{Xingcheng Zhang}, \bibinfo{person}{Kai Chen}, \bibinfo{person}{Yu Qiao}, \bibinfo{person}{Dahua Lin}, {and} \bibinfo{person}{Jiaqi Wang}.} \bibinfo{year}{2024}\natexlab{c}.
\newblock \showarticletitle{InternLM-XComposer2.5-OmniLive: A Comprehensive Multimodal System for Long-term Streaming Video and Audio Interactions}.
\newblock \bibinfo{journal}{\emph{arXiv preprint arXiv:2412.09596}} (\bibinfo{year}{2024}).
\newblock


\bibitem[Zhang et~al\mbox{.}(2024f)]%
        {longva}
\bibfield{author}{\bibinfo{person}{Peiyuan Zhang}, \bibinfo{person}{Kaichen Zhang}, \bibinfo{person}{Bo Li}, \bibinfo{person}{Guangtao Zeng}, \bibinfo{person}{Jingkang Yang}, \bibinfo{person}{Yuanhan Zhang}, \bibinfo{person}{Ziyue Wang}, \bibinfo{person}{Haoran Tan}, \bibinfo{person}{Chunyuan Li}, {and} \bibinfo{person}{Ziwei Liu}.} \bibinfo{year}{2024}\natexlab{f}.
\newblock \showarticletitle{Long Context Transfer from Language to Vision}.
\newblock \bibinfo{journal}{\emph{ArXiv preprint}}  \bibinfo{volume}{abs/2406.16852} (\bibinfo{year}{2024}).
\newblock


\bibitem[Zhang et~al\mbox{.}(2024e)]%
        {llava-video-sft}
\bibfield{author}{\bibinfo{person}{Yuanhan Zhang}, \bibinfo{person}{Jinming Wu}, \bibinfo{person}{Wei Li}, \bibinfo{person}{Bo Li}, \bibinfo{person}{Zejun Ma}, \bibinfo{person}{Ziwei Liu}, {and} \bibinfo{person}{Chunyuan Li}.} \bibinfo{year}{2024}\natexlab{e}.
\newblock \bibinfo{title}{Video Instruction Tuning With Synthetic Data}.
\newblock
\showeprint[arxiv]{2410.02713}~[cs.CV]
\urldef\tempurl%
\url{https://arxiv.org/abs/2410.02713}
\showURL{%
\tempurl}


\bibitem[Zhang et~al\mbox{.}(2023b)]%
        {h2o}
\bibfield{author}{\bibinfo{person}{Zhenyu Zhang}, \bibinfo{person}{Ying Sheng}, \bibinfo{person}{Tianyi Zhou}, \bibinfo{person}{Tianlong Chen}, \bibinfo{person}{Lianmin Zheng}, \bibinfo{person}{Ruisi Cai}, \bibinfo{person}{Zhao Song}, \bibinfo{person}{Yuandong Tian}, \bibinfo{person}{Christopher R{\'e}}, \bibinfo{person}{Clark Barrett}, {et~al\mbox{.}}} \bibinfo{year}{2023}\natexlab{b}.
\newblock \showarticletitle{H2o: Heavy-hitter oracle for efficient generative inference of large language models}.
\newblock \bibinfo{journal}{\emph{Advances in Neural Information Processing Systems}}  \bibinfo{volume}{36} (\bibinfo{year}{2023}), \bibinfo{pages}{34661--34710}.
\newblock


\bibitem[Zhou et~al\mbox{.}(2024)]%
        {MLVU}
\bibfield{author}{\bibinfo{person}{Junjie Zhou}, \bibinfo{person}{Yan Shu}, \bibinfo{person}{Bo Zhao}, \bibinfo{person}{Boya Wu}, \bibinfo{person}{Shitao Xiao}, \bibinfo{person}{Xi Yang}, \bibinfo{person}{Yongping Xiong}, \bibinfo{person}{Bo Zhang}, \bibinfo{person}{Tiejun Huang}, {and} \bibinfo{person}{Zheng Liu}.} \bibinfo{year}{2024}\natexlab{}.
\newblock \showarticletitle{MLVU: A Comprehensive Benchmark for Multi-Task Long Video Understanding}.
\newblock \bibinfo{journal}{\emph{ArXiv preprint}}  \bibinfo{volume}{abs/2406.04264} (\bibinfo{year}{2024}).
\newblock


\bibitem[Zhou et~al\mbox{.}(2018)]%
        {youcook2}
\bibfield{author}{\bibinfo{person}{Luowei Zhou}, \bibinfo{person}{Chenliang Xu}, {and} \bibinfo{person}{Jason~J Corso}.} \bibinfo{year}{2018}\natexlab{}.
\newblock \showarticletitle{Towards Automatic Learning of Procedures From Web Instructional Videos}. In \bibinfo{booktitle}{\emph{AAAI Conference on Artificial Intelligence}}. \bibinfo{pages}{7590--7598}.
\newblock
\urldef\tempurl%
\url{https://www.aaai.org/ocs/index.php/AAAI/AAAI18/paper/view/17344}
\showURL{%
\tempurl}


\end{thebibliography}

\appendix
\section{Appendix}

In the appendix, we provide more results and analysis and summarize them as follows:
\begin{itemize}
    \item In Section A.1, we report extensive experimental results.
    \item In Section A.2, we present the training hyperparameter tables.
    \item In Section A.3, we present details of \dataset.
    \item In Section A.4, we introduce diverse visualization cases for the Differential Token Drop design.
\end{itemize}

\subsection{More Experimental Results}

\noindent\textbf{StreamingBench Full-set Results.} Table~\ref{tab:supp_streamingbench_full} presents the comprehensive results across all three categories of StreamingBench. \model achieves state-of-the-art performance among open-source models with an overall score of 58.11, outperforming the recent online model Dispider-7B~\cite{dispider} by 4.99 points (58.11 vs. 53.12). While proprietary model Gemini 1.5 pro leads with 67.07, \model surpasses Claude-3.5-Sonnet (57.68) and all open-source offline \videollms.
For Omni-Source and Contextual Understanding, \model scores 37.80 and 35.30 respectively, consistently surpassing Dispider-7B.

Most importantly, with 82.6\% token reduction, \model maintains 97.3\% of its original performance (56.56 vs. 58.11), confirming that most visual tokens in streaming videos are redundant while our approach preserves essential information for comprehensive understanding.

\noindent\textbf{Selection of Hyperparameters $\tau$.}
The values of $\tau_{pixel}$ and $\tau_{feat}$ control the overall \textit{drop ratio} of video tokens by dynamically measuring the visual difference between temporally consecutive frames. This approach identifies redundancy based solely on the video's inherent temporal characteristics, without requiring language guidance. Remarkably, we observe that the relationship between $\tau$ values and drop ratios remains consistent across diverse datasets, as shown in Table~\ref{tab:tau_drop_ratio}. For example, $\tau_{feat}=0.25$ consistently yields approximately 85\% token reduction across StreamingBench, OVOBench, LongVideoBench, and MLVU, while $\tau_{feat}=0.5$ results in approximately 45\%-55\% reduction. More $\tau$-dropratio correspondence can also be found in Figure 1 (top right) in the main paper.

This consistency demonstrates that visual redundancy is an intrinsic property of videos regardless of content type or task domain. Based on our comprehensive experiments, we conclude that over 80\% of visual information in long-form videos is naturally redundant. Therefore, we recommend using $\tau_{feat}=0.25$ as the optimal setting for video token dropping, as it maintains high performance while significantly reducing computational requirements.

\noindent\textbf{Breakdown Analysis of Fine-grained subtasks on MLVU.} Table~\ref{tab:mlvu_subtasks} shows the performance breakdown on fine-grained subtasks of MLVU. Remarkably, even with 89.5\% token reduction, our model not only maintains but improves performance across multiple challenging fine-grained visual understanding tasks compared to the full token baseline. The needle-in-haystack task (NQA) improves from 80.0 to 82.0, plot reading (PQA) from 66.4 to 68.8, temporal ordering (AO) from 40.9 to 49.4, and counting (AC) from 34.0 to 42.2, resulting in an overall improvement from 62.0 to 64.1.

These results demonstrate that \module effectively preserves essential visual details while discarding redundant information. Despite dropping nearly 90\% of tokens, the model maintains the capability to recognize small objects (needle task), interpret plots, track temporal relationships, and count objects accurately. This is enabled by our spatial-temporal aware design, which preserves the original video's spatiotemporal structure even after significant token reduction. The improved performance on the counting task (AC) is particularly noteworthy, showing an 8.2 point gain, which confirms that our approach selectively retains detailed visual information critical for complex analysis tasks.

\begin{table}[t]
    \centering
    \caption{Consistency of drop ratios (\%) across different datasets with the same $\tau_{feat}$ values, demonstrating that our method captures intrinsic video redundancy independent of dataset characteristics.}
    \label{tab:tau_drop_ratio}
    \begin{adjustbox}{max width=\columnwidth}
    \begin{tabular}{c|cccc}
    \toprule
    & \multicolumn{4}{c}{\textbf{Drop Ratio (\%)}} \\
    $\tau_{feat}$ & StreamingBench & OVOBench & LongVideoBench & MLVU \\
    \midrule
    0.5 & 44.1\% & 44.6\% & 56.5\% & 46.3\% \\
    0.25 & 82.6\% & 84.8\% & 85.1\% & 84.7\% \\
    \bottomrule
    \end{tabular}
    \end{adjustbox}
\end{table}

\begin{table}[t]
    \small
    \centering
    \begin{adjustbox}{max width=\columnwidth}
    \begin{tabular}{lr}
    \toprule
    \textbf{Hyper-parameter} & \textbf{Value} \\
    \midrule
    \multicolumn{2}{l}{\textcolor{gray}{\textit{Visual Encoder}}} \\
    \midrule
    Frame Sampling Rate &FPS=1.0 \\
    Input Resolution & 448*448 \\
    Visual Tokens per Image & 128 \\
    Max Image per Sequence & 64 \\
    Patch Size & 14x14 \\
    \midrule

    \multicolumn{2}{l}{\textcolor{gray}{\textit{Large Language Model}}} \\
    \midrule
    Number of Layers & 80 \\
    Hidden Size & 8192  \\
    Vocabularyy Size & 152064 \\
    Number of Attention Heads & 64 \\
    Number of KV Heads & 64 \\
    Number of KV Heads & 8 \\
    \midrule

    \midrule

    \multicolumn{2}{l}{\textcolor{gray}{\textit{Model Training}}} \\
    \midrule
    Max Context Length & 11264 \\
    Batch Size & 128 \\
    Learning Rate & 1e-5 \\
    Warmup Ratio & 0.05 \\
    Training epoch1 & 1 \\ %
    LR Scheduler Type & Cosine \\
    \bottomrule
    \end{tabular}
    \end{adjustbox}
    \caption{Training hyper-parameters for \text{\model}.}
    \label{tab:supp_hyperpara}
\end{table}

\begin{table}[t]
    \centering
    \caption{Breakdown analysis on fine-grained subtasks of MLVU. Our approach with feature-level token dropping (89.5\% reduction) improves performance across all subtasks. ↓ indicates lower than human performance, while ↑ indicates comparable or better than human performance.}
    \label{tab:mlvu_subtasks}
    \begin{adjustbox}{max width=\columnwidth}
    \begin{tabular}{l|ccccc|c}
    \toprule
     & \textbf{Drop Ratio (\%)} & \textbf{NeedleQA} & \textbf{PlotQA)} & \textbf{Action Order} & \textbf{Action Counting} & \textbf{M-Avg} \\
    \midrule
    No drop & 100\% & 80.0 & 66.4 & 40.9 & 34.0 & 62.0 \\
    $\tau_feat=0.5$ & 46.4\% & 80.8 & 67.7 & 44.0 & 32.0 & 61.9 \\
    $\tau_feat=0.25$ & 89.5\% & 82.0 & 68.8 & 49.4 & 42.2 & 64.1 \\ \midrule
    \rowcolor{gray!20} $\Delta$ & \textbf{-} & \textbf{+2.0} & \textbf{+2.4} & \textbf{+8.5} & \textbf{+8.2} & \textbf{+2.1} \\
    \bottomrule
    \end{tabular}
    \end{adjustbox}
\end{table}

\subsection{Training Hyperparameter}

We provide detailed hyper-parameters in Table~\ref{tab:supp_hyperpara}.

\subsection{Dataset Statistics}
\label{sec:data_stat}

\begin{figure}[t]
    \centering
    \includegraphics[width=0.9\columnwidth]{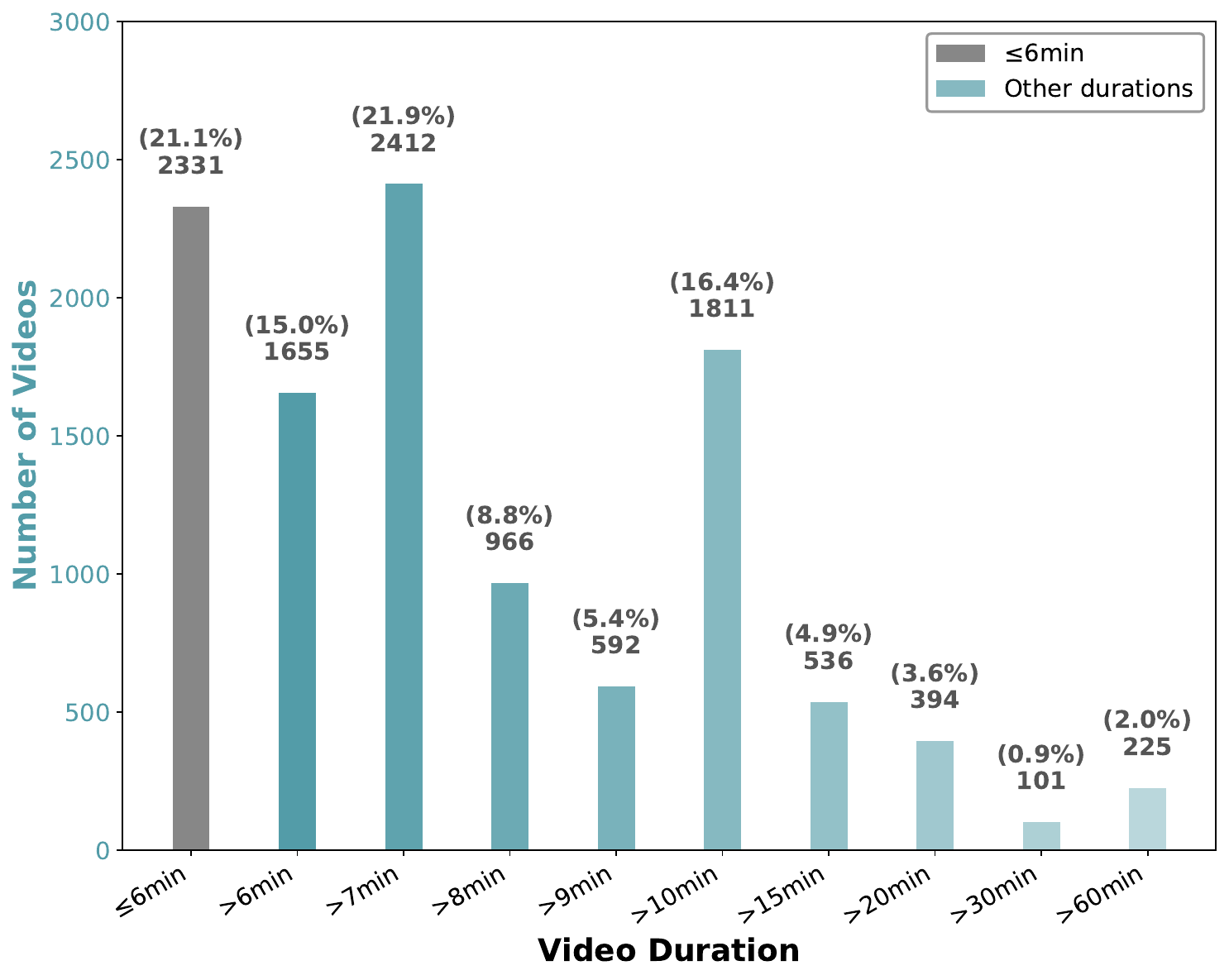}
    \caption{Distribution of video durations across the 11,043 videos in our dataset. The minimum video length in our dataset is 5 minutes.}
    \label{fig:video_durations_distribution}
\end{figure}

Our dataset comprises 11,043 videos sampled from 12 publicly available source datasets, with an average duration of 11.1 minutes per video. The composition of the video sources and the distribution of video durations are shown in Table~\ref{tab:video-sources} and Figure~\ref{fig:video_durations_distribution}, respectively. Based on these videos, we generate a total of 139K question-answer (QA) pairs, following the procedure outlined in main Section 3.3. These QA pairs are categorized into four high-level types—\textit{Temporal-enhanced}, \textit{Backward Tracing}, \textit{Real-Time Visual Perception}, and \textit{Forward Active Responding}—encompassing eleven fine-grained subcategories. Representative prompt examples for each subtask are provided in Table~\ref{tab:task-taxonomy}, while the prompt templates for Step 2: Scene-oriented Detailed Caption Generation and Step 3: Streaming QA Pair Construction are shown in Table~\ref{tab:frame_caption_prompt} and Table~\ref{tab:question_generation_prompt}, respectively.

\begin{table*}[t]
\centering
\setlength{\tabcolsep}{10pt}
\renewcommand{\arraystretch}{1.2}
\begin{tabular}{l r l r l r}
\toprule
\textbf{Dataset} & \textbf{\#Videos} & \textbf{Dataset} & \textbf{\#Videos} & \textbf{Dataset} & \textbf{\#Videos} \\
\midrule
COIN~\cite{coin}           & 151   & QV-Highlights~\cite{qvhighlights} & 1778  & ActivityNet~\cite{activitynet} & 12    \\
HD-VILA~\cite{hd-vila}      & 695   & YouCook2~\cite{youcook2}          & 710   & TVSum~\cite{tvsum}              & 10    \\
ViTT~\cite{vitt}           & 2000  & QuerYD~\cite{queryd}              & 566   & YouMakeup~\cite{youmakeup}      & 1801  \\
VideoIC~\cite{videoic}     & 2649  & Movie101~\cite{movie101}          & 202   & HiREST~\cite{hirest}            & 469   \\
\midrule
\multicolumn{5}{r}{\textbf{Total}} & \textbf{11,043} \\
\bottomrule
\end{tabular}
\caption{Source datasets and number of unique videos sampled from each. Dataset names link to their respective original publications.}
\label{tab:video-sources}
\end{table*}

\begin{table*}[t]
\centering
\begin{adjustbox}{width=0.98\textwidth}
\renewcommand{\arraystretch}{1.8}
\setlength{\tabcolsep}{8pt}
\begin{tabular}{llp{10cm}}
\toprule
\textbf{Main Category} & \textbf{Subcategory} & \textbf{Prompt Example} \\
\midrule
\multirow{5}{*}{Temporal-enhanced} 
  & Event order & What is the correct order of events in this video? \\
  & Key Attribute change & What are the key changes in the main character's attire from the beginning to the end of the video? \\
  & Camera transitions & How does the camera angle change between the beginning and middle frames? \\
  & Event relationship & Which event directly leads to [Cui Hu] aiming his gun at an adversary? \\
  & Key Frame Focus & Which frame captures the progression from indoor reflection to outdoor contemplation for Grandma Li Ailian? \\
\midrule
\multirow{3}{*}{Backward Tracing} 
  & Episodic Memory & Which event occurred before the individual pointed towards the forest? \\
  & Action Sequence Identification & Which action occurred first in the makeup tutorial? \\
  & Hallucination Detection & Did the presenter ever apply a green eyeshadow during the tutorial? \\
\midrule
\multirow{1}{*}{Real-Time Visual Perception} 
  & Future Prediction & What is likely to happen next after the presenter shows the highlighter stick? \\
\midrule
\multirow{2}{*}{Forward Active Responding} 
  & Sequential Steps Recognition & What are the steps involved in applying bronzer as shown in the tutorial? \\
  & Clues Reveal Responding & What accessory was visible when the person was blending the contour product? \\
\bottomrule
\end{tabular}
\end{adjustbox}
\caption{Task taxonomy and corresponding prompt examples.}
\label{tab:task-taxonomy}
\end{table*}

\begin{table*}[t]
    \centering
    \caption{Performance comparison on StreamingBench full set including three categories: Real-Time Visual Understanding, Omni-Source Understanding and Contextual Understanding.}
    \label{tab:supp_streamingbench_full}
    \small
    \begin{adjustbox}{max width=1.0\linewidth}
    \begin{tabular}{l@{}c@{\hspace{3pt}}c | cccccccccc c | cccc c | cccc c | c}
    \toprule
    \multirow{2}{*}{\textbf{Model}} & \multirow{2}{*}{\textbf{Params}} & \multirow{2}{*}{\textbf{Frames}} & \multicolumn{11}{c|}{\textbf{Real-Time Visual Understanding}} & \multicolumn{5}{c|}{\textbf{Omni-Source Understanding}} & \multicolumn{5}{c|}{\textbf{Contextual Understanding}} & \multirow{2}{*}{\textbf{Overall}} \\
    \addlinespace[2pt]
    \cline{4-24}
    \addlinespace[2pt]
    & & & \textbf{OP} & \textbf{CR} & \textbf{CS} & \textbf{ATP} & \textbf{EU} & \textbf{TR} & \textbf{PR} & \textbf{SU} & \textbf{ACP} & \textbf{CT} & \textbf{All} & \textbf{ER} & \textbf{SCU} & \textbf{SD} & \textbf{MA} & \textbf{All} & \textbf{ACU} & \textbf{MCU} & \textbf{SQA} & \textbf{PO} & \textbf{All} & \\
    \midrule
    \midrule
    Human$^{\dagger}$ & - & - & 89.47 & 92.00 & 93.60 & 91.47 & 95.65 & 92.52 & 88.00 & 88.75 & 89.74 & 91.30 & 91.46 & 88.00 & 88.24 & 93.60 & 90.27 & 90.26 & 88.80 & 90.40 & 95.00 & 100 & 93.55 & 91.66 \\
    \midrule
    \multicolumn{25}{c}{\textbf{Proprietary MLLMs}} \\
    \midrule
    Gemini 1.5 pro & - & 1 fps & 79.02 & 80.47 & 83.54 & 79.67 & 80.00 & 84.74 & 77.78 & 64.23 & 71.95 & 48.70 & 75.69 & 46.80 & 39.60 & 74.90 & 80.00 & 60.22 & 51.41 & 40.73 & 54.80 & 45.10 & 48.73 & 67.07 \\
    GPT-4o & - & 64 & 77.11 & 80.47 & 83.91 & 76.47 & 70.19 & 83.80 & 66.67 & 62.19 & 69.12 & 49.22 & 73.28 & 41.20 & 37.20 & 43.60 & 56.00 & 44.50 & 41.20 & 38.40 & 32.80 & 56.86 & 38.70 & 60.15 \\
    Claude 3.5 Sonnet & - & 20 & 80.49 & 77.34 & 82.02 & 81.73 & 72.33 & 75.39 & 61.11 & 61.79 & 69.32 & 43.09 & 72.44 & 31.60 & 34.00 & 32.80 & 48.80 & 36.80 & 38.40 & 34.80 & 34.40 & 64.71 & 37.70 & 57.68 \\
    \midrule
    \multicolumn{25}{c}{\textbf{Open-Source Video MLLMs}} \\
    \midrule
    LLaVA-OneVision & 7B & 32 & 80.38 & 74.22 & 76.03 & 80.72 & 72.67 & 71.65 & 67.59 & 65.45 & 65.72 & 45.08 & 71.12 & 40.80 & 37.20 & 33.60 & 44.80 & 38.40 & 35.60 & 36.00 & 27.27 & 29.55 & 32.74 & 56.36 \\
    Qwen2-VL & 7B & 0.2-1 fps & 75.20 & 82.81 & 73.19 & 77.45 & 68.32 & 71.03 & 72.22 & 61.19 & 61.47 & 46.11 & 69.04 & 41.20 & 22.00 & 32.80 & 43.60 & 34.90 & 31.20 & 26.00 & 39.60 & 22.73 & 31.66 & 54.14 \\
    MiniCPM-V 2.6 & 8B & 32 & 71.93 & 71.09 & 77.92 & 75.82 & 64.60 & 65.73 & 70.37 & 56.10 & 62.32 & 53.37 & 67.44 & 40.80 & 24.00 & 34.00 & 41.20 & 35.00 & 34.00 & 31.60 & 41.92 & 22.22 & 34.97 & 53.85 \\
    LLaVA-NeXT-Video & 32B & 64 & 78.20 & 70.31 & 73.82 & 76.80 & 63.35 & 69.78 & 57.41 & 56.10 & 64.31 & 38.86 & 66.96 & 37.69 & 24.80 & 34.40 & 42.80 & 34.90 & 29.20 & 30.40 & 35.35 & 18.18 & 30.79 & 52.77 \\
    InternVL-V2 & 8B & 16 & 68.12 & 60.94 & 69.40 & 77.12 & 67.70 & 62.93 & 59.26 & 53.25 & 54.96 & 56.48 & 63.72 & 37.60 & 26.40 & 37.20 & 42.00 & 35.80 & 32.00 & 31.20 & 32.32 & 40.91 & 32.42 & 51.40 \\
    Kangaroo & 7B & 64 & 71.12 & 84.38 & 70.66 & 73.20 & 67.08 & 61.68 & 56.48 & 55.69 & 62.04 & 38.86 & 64.60 & 37.60 & 31.20 & 28.80 & 39.20 & 34.20 & 32.80 & 26.40 & 33.84 & 16.00 & 30.06 & 51.10 \\
    LongVA & 7B & 128 & 70.03 & 63.28 & 61.20 & 70.92 & 62.73 & 59.50 & 61.11 & 53.66 & 54.67 & 34.72 & 59.96 & 39.60 & 32.40 & 28.00 & 41.60 & 35.40 & 32.80 & 29.60 & 30.30 & 15.91 & 29.95 & 48.66 \\
    VILA-1.5 & 8B & 14 & 53.68 & 49.22 & 70.98 & 56.86 & 53.42 & 53.89 & 54.63 & 48.78 & 50.14 & 17.62 & 52.32 & 41.60 & 26.40 & 28.40 & 36.00 & 33.10 & 26.80 & 34.00 & 23.23 & 17.65 & 27.35 & 43.20 \\
    Video-CCAM & 14B & 96 & 56.40 & 57.81 & 65.30 & 62.75 & 64.60 & 51.40 & 42.59 & 47.97 & 49.58 & 31.61 & 53.96 & 33.60 & 22.00 & 28.40 & 34.80 & 29.70 & 27.60 & 24.40 & 16.67 & 22.73 & 22.88 & 42.53 \\
    Video-LLaMA2 & 7B & 32 & 55.86 & 55.47 & 57.41 & 58.17 & 52.80 & 43.61 & 39.81 & 42.68 & 45.61 & 35.23 & 49.52 & 30.40 & 32.40 & 30.40 & 36.00 & 32.40 & 24.80 & 26.80 & 18.67 & 0.00 & 21.93 & 40.40 \\
    \midrule
    \multicolumn{25}{c}{\textbf{Streaming MLLMs}} \\
    \midrule
    Flash-VStream & 7B & - & 25.89 & 43.57 & 24.91 & 23.87 & 27.33 & 13.08 & 18.52 & 25.20 & 23.87 & 48.70 & 23.23 & 25.91 & 24.90 & 25.60 & 28.40 & 26.00 & 24.80 & 25.20 & 26.80 & 1.96 & 24.12 & 24.04 \\
    VideoLLM-online & 8B & 2 fps & 39.07 & 40.06 & 34.49 & 31.05 & 45.96 & 32.40 & 31.48 & 34.16 & 42.49 & 27.89 & 35.99 & 31.20 & 26.51 & 24.10 & 32.00 & 28.45 & 24.19 & 29.20 & 30.80 & 3.92 & 26.55 & 32.48 \\
    Dispider & 7B & 1 fps & 74.92 & 75.53 & 74.10 & 73.08 & 74.44 & 59.92 & 76.14 & 62.91 & 62.16 & 45.80 & 67.63 & 35.46 & 25.26 & 38.57 & 43.34 & 35.66 & 39.62 & 27.65 & 34.80 & 25.34 & 33.61 & 53.12 \\
    \rowcolor{gray!20} \textbf{\model-7B} & 7B & 1 fps ($\downarrow$44.2\%) & 80.76 & 79.69 & 80.76 & 83.33 & 74.84 & 78.82 & 78.70 & 64.23 & 68.75 & 57.98 & 75.28 & 41.60 & 29.60 & 34.80 & 52.00 & 39.50 & 41.20 & 30.40 & 42.80 & 18.80 & 33.30 & \textbf{58.00} \\
    \rowcolor{gray!40} \textbf{\model-7B} & 7B & 1 fps ($\downarrow$82.6\%) & 79.13 & 81.25 & 78.86 & 80.77 & 70.44 & 77.26 & 77.78 & 67.07 & 66.19 & 53.72 & 73.64 & 40.00 & 32.80 & 36.80 & 52.00 & 40.40 & 38.40 & 26.80 & 40.40 & 14.40 & 30.00 & \underline{56.56} \\
    \midrule
    \textbf{\model-7B} & 7B & 1fps (100\%) & 80.22 & 82.03 & 79.50 & 83.33 & 76.10 & 78.50 & 78.70 & 64.63 & 69.60 & 57.98 & 75.36 & 38.40 & 26.80 & 35.60 & 50.40 & 37.80 & 39.20 & 31.60 & 41.60 & 28.80 & 35.30 & \textbf{58.11} \\
    \bottomrule
    \end{tabular}
    \end{adjustbox}
\end{table*}




\subsection{Diverse Visualization Cases}

\begin{figure*}[t]
    \centering
    \begin{minipage}[t]{0.48\textwidth}
        \centering
        \includegraphics[width=\linewidth]{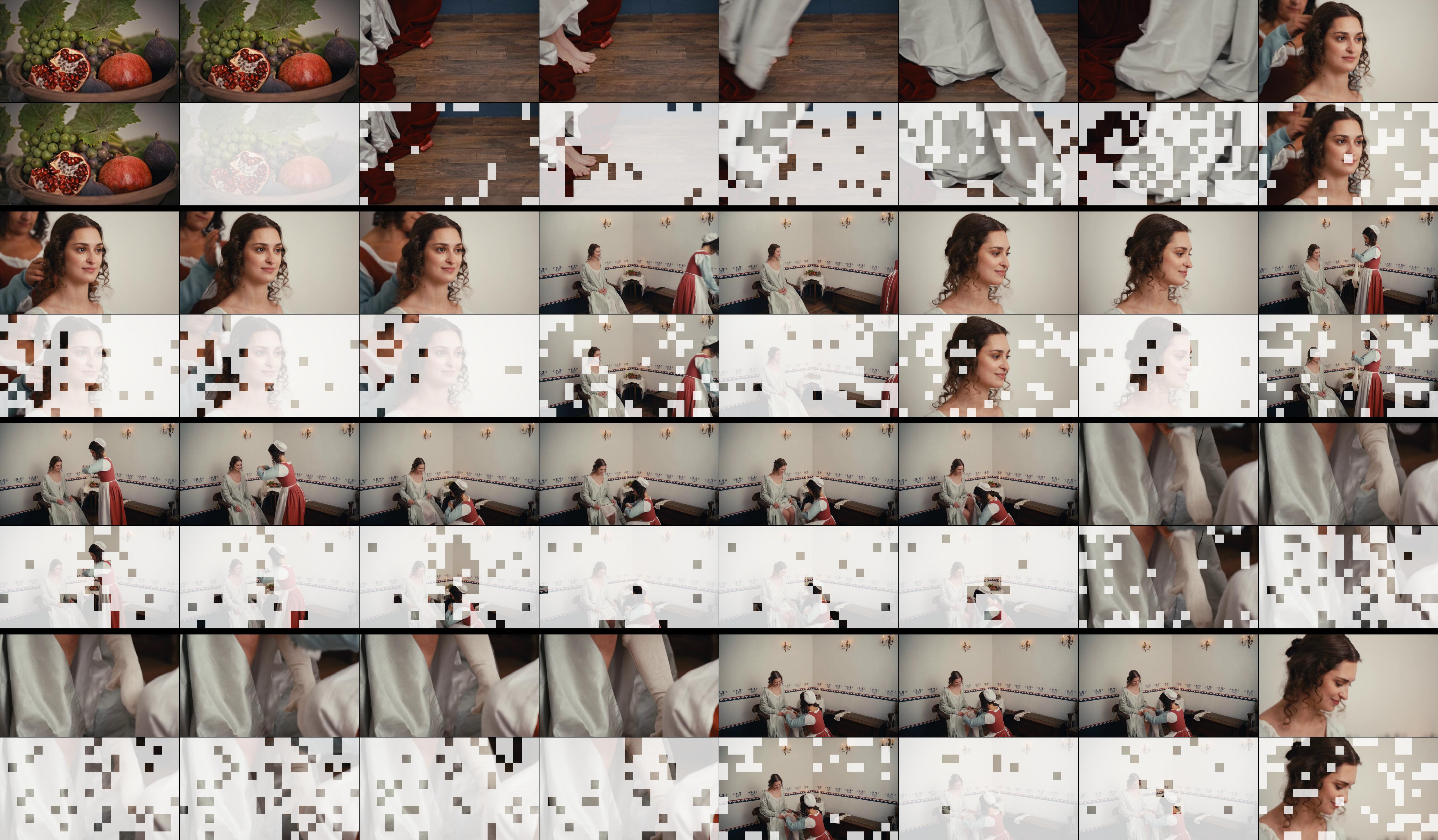}
        \subcaption{Feature-level: $\tau_{feat}=0.4, \text{drop ratio}=58.3\%$}
        \label{fig:supp_case752_ft}
    \end{minipage}%
    \hfill
    \begin{minipage}[t]{0.48\textwidth}
        \centering
        \includegraphics[width=\linewidth]{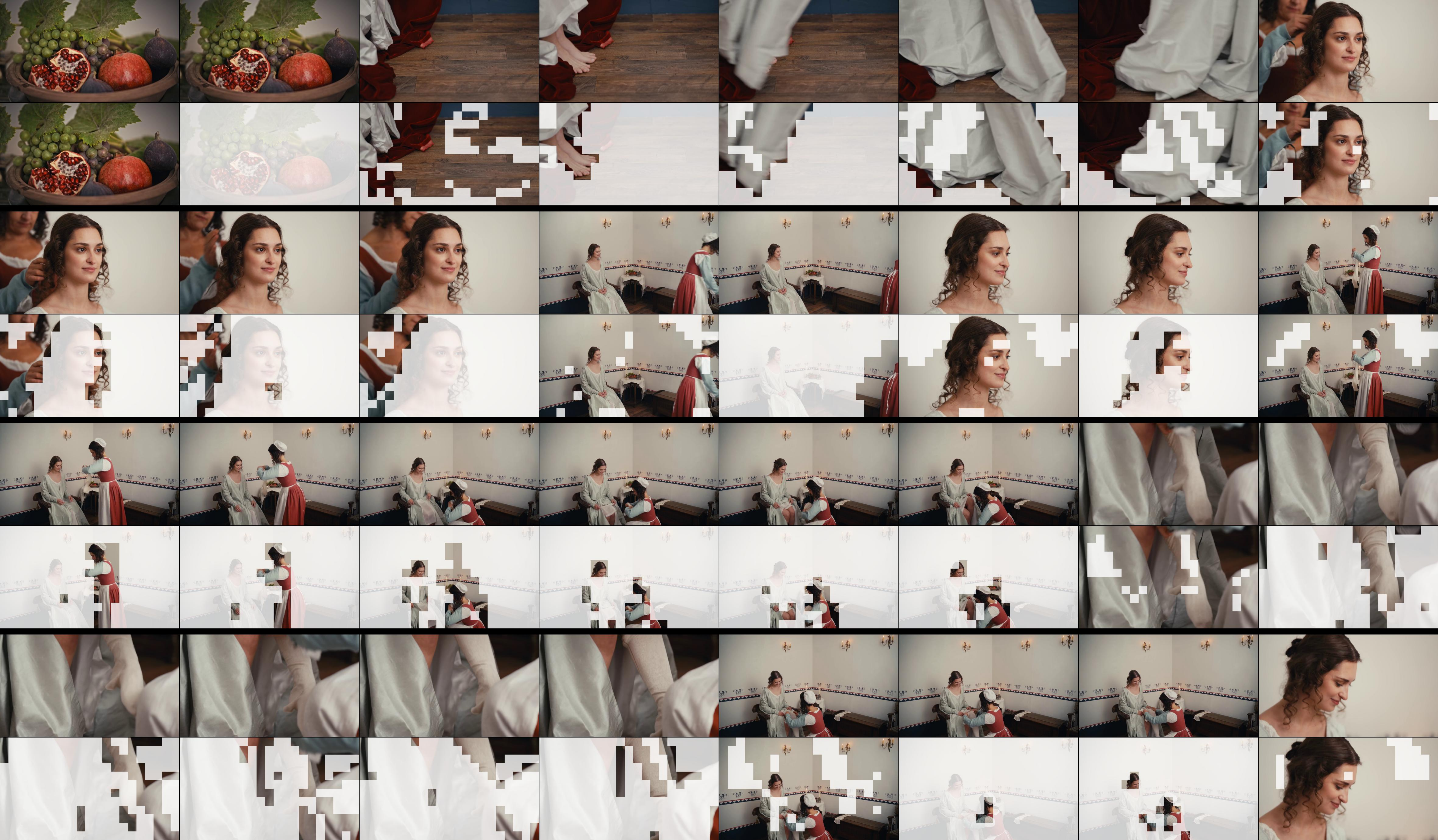}
        \subcaption{Pixel-level: $\tau_{pixel}=0.1$}
        \label{fig:supp_case752_px}
    \end{minipage}
    \caption{Visualization of (a) Feature-level token dropping and (b) Pixel-level token dropping for the video case 752 from StreamingBench.}
    \label{fig:supp_case752}
\end{figure*}

\begin{figure*}[t]
    \centering
    \begin{minipage}[t]{0.48\textwidth}
        \centering
        \includegraphics[width=\linewidth]{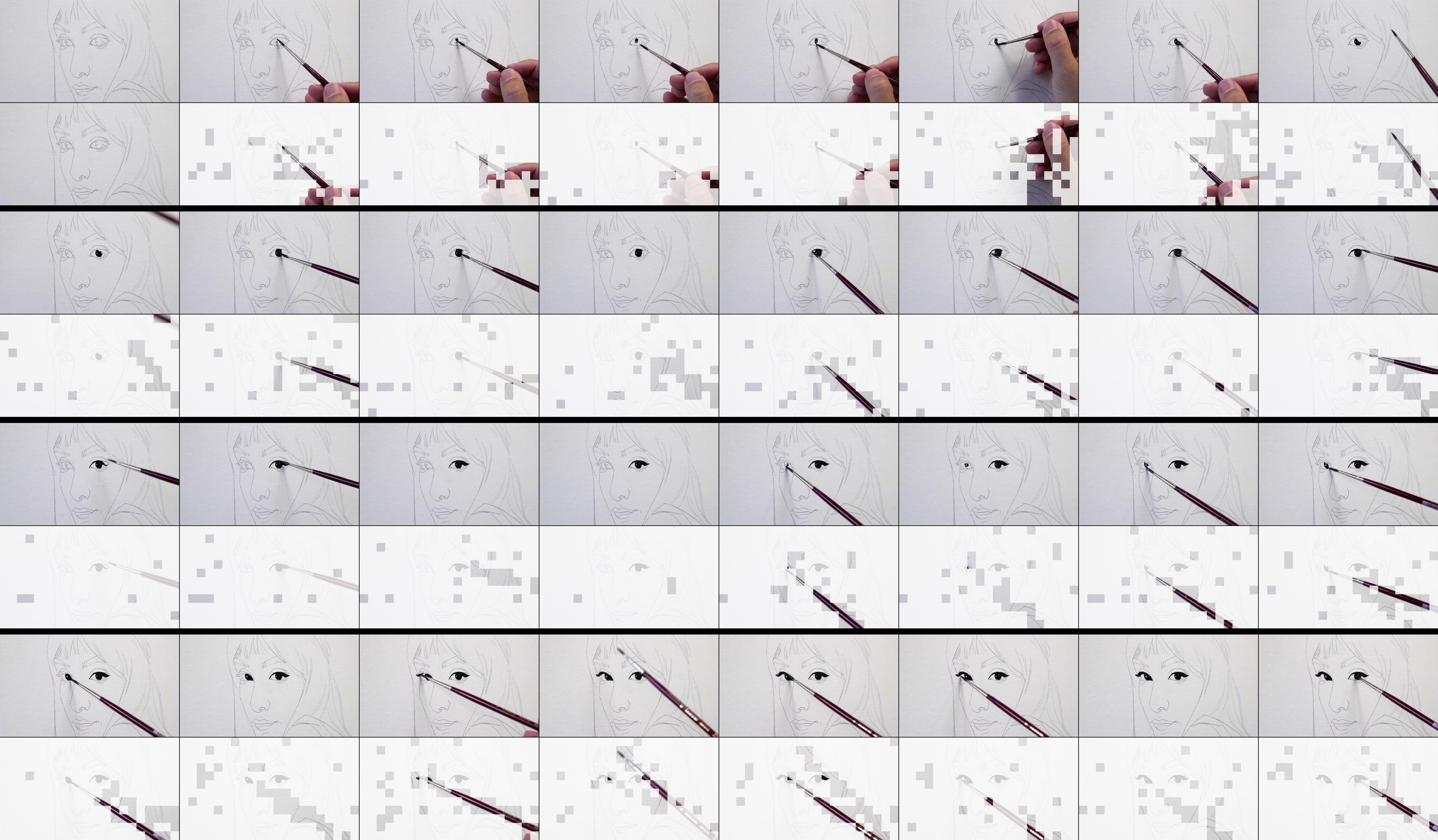}
        \subcaption{Feature-level: $\tau_{feat}=0.4, \text{drop ratio}=89.5\%$}
        \label{fig:supp_case671_ft}
    \end{minipage}%
    \hfill
    \begin{minipage}[t]{0.48\textwidth}
        \centering
        \includegraphics[width=\linewidth]{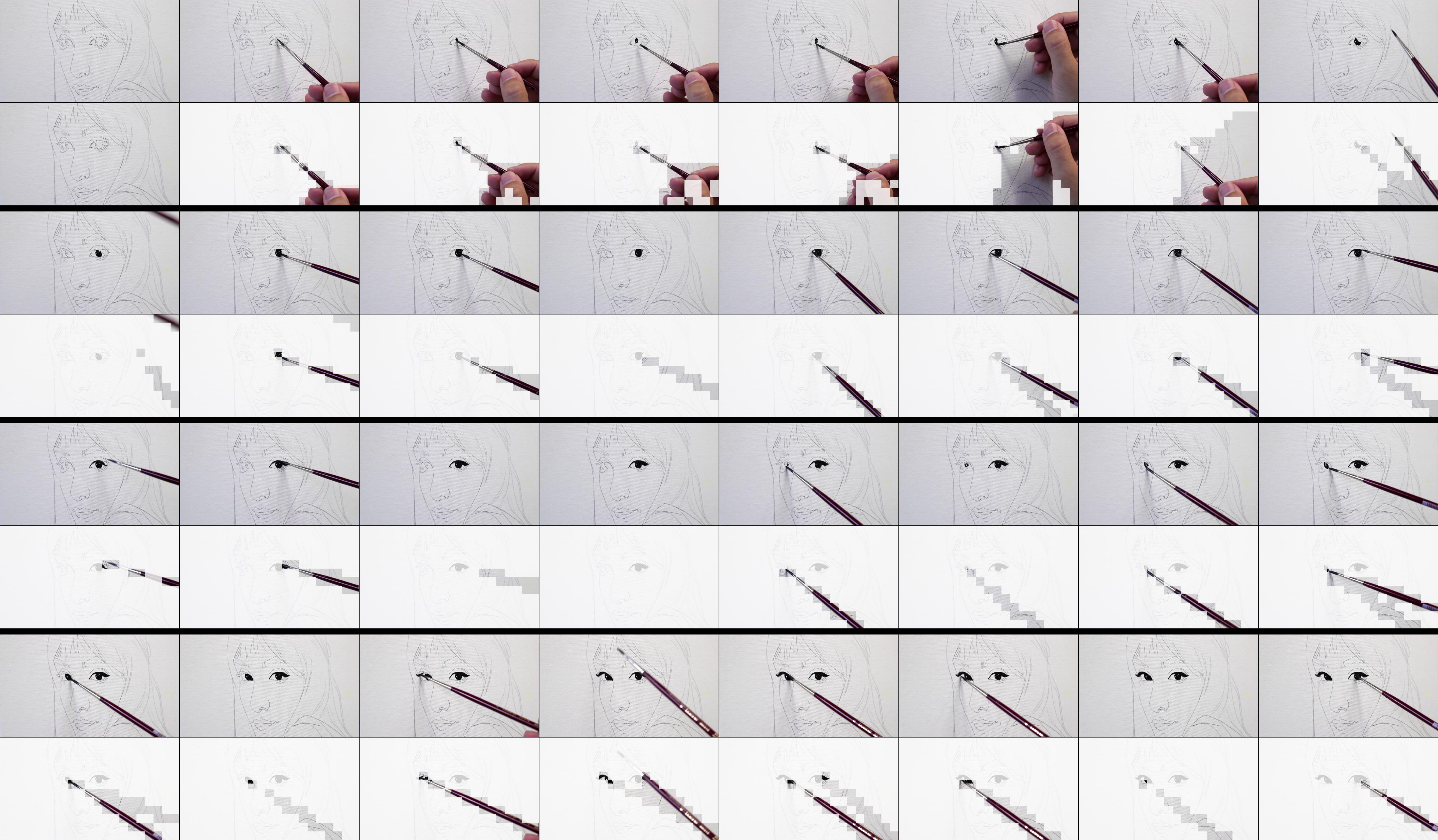}
        \subcaption{Pixel-level: $\tau_{pixel}=0.1$}
        \label{fig:supp_case671_px}
    \end{minipage}
    \caption{Visualization of (a) Feature-level token dropping and  (b) Pixel-level token dropping for the video case 671 from StreamingBench.}
    \label{fig:supp_case671}
\end{figure*}

\begin{figure*}[t]
    \centering
    \begin{minipage}[t]{0.48\textwidth}
        \centering
        \includegraphics[width=\linewidth]{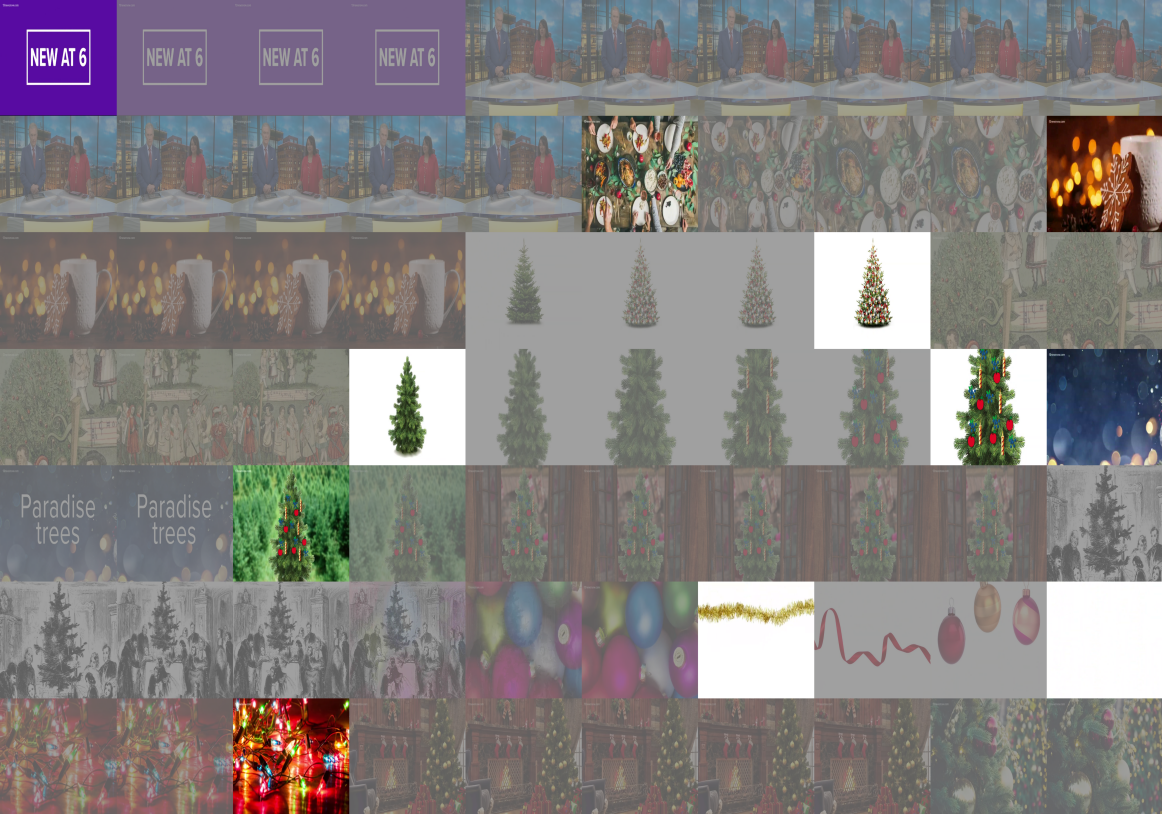}
        \subcaption{Scene Transition Point w/ Trigger Time}
    \end{minipage}%
    \hfill
    \begin{minipage}[t]{0.5\textwidth}
        \centering
        \includegraphics[width=\linewidth]{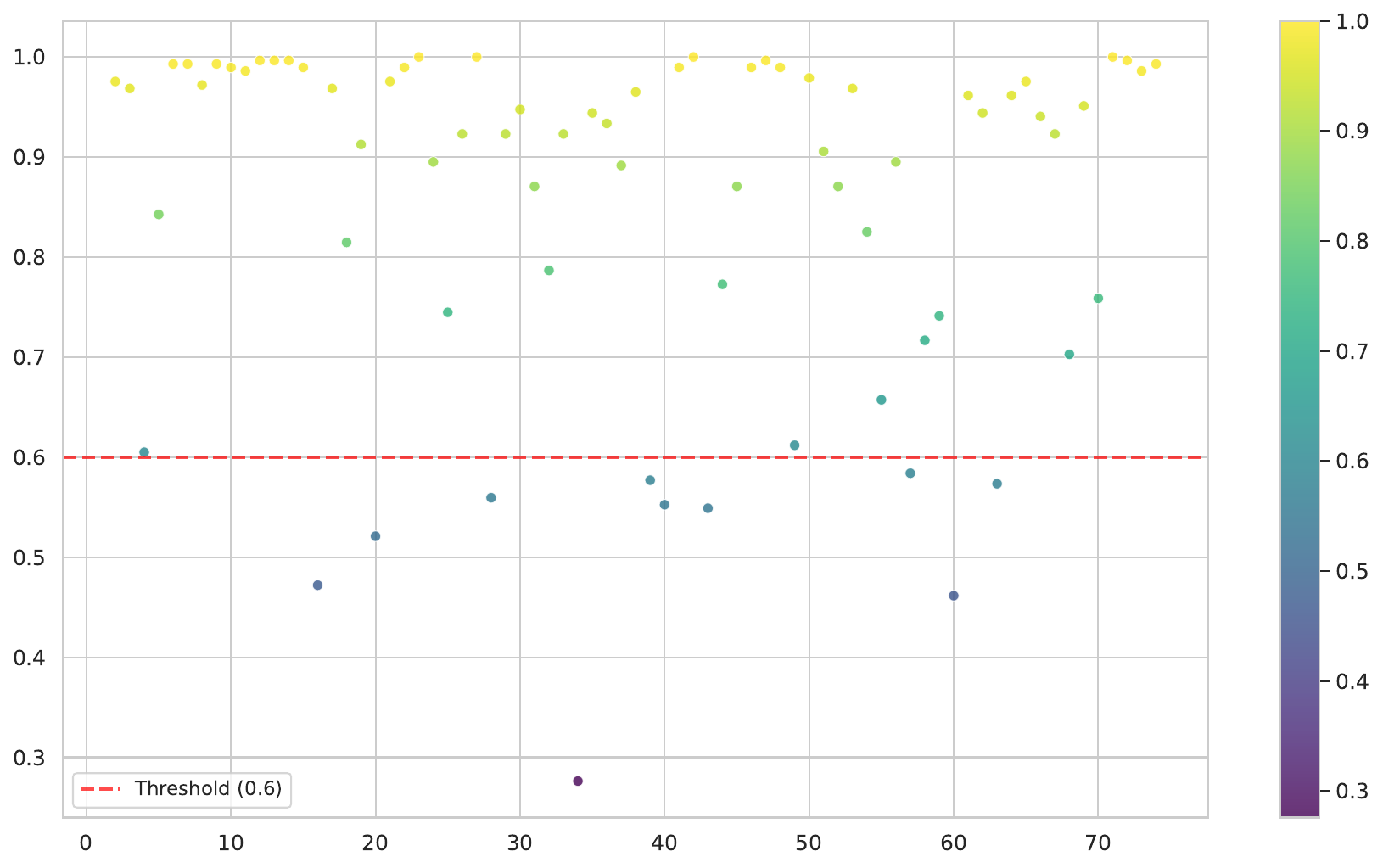}
        \subcaption{Drop Ratio - Timeline Curve}
    \end{minipage}
    \caption{Visualization of monitored \textit{Trigger Time} via drop ratio curve. The colored highlighted frames correspond to trigger times that reveal video scene transitions. Our model utilizes a temporal patch size of 2.}
    \label{fig:supp_triggertime_case1}
\end{figure*}

\begin{figure*}[t]
    \centering
    \begin{minipage}[t]{0.38\textwidth}
        \centering
        \includegraphics[width=\linewidth]{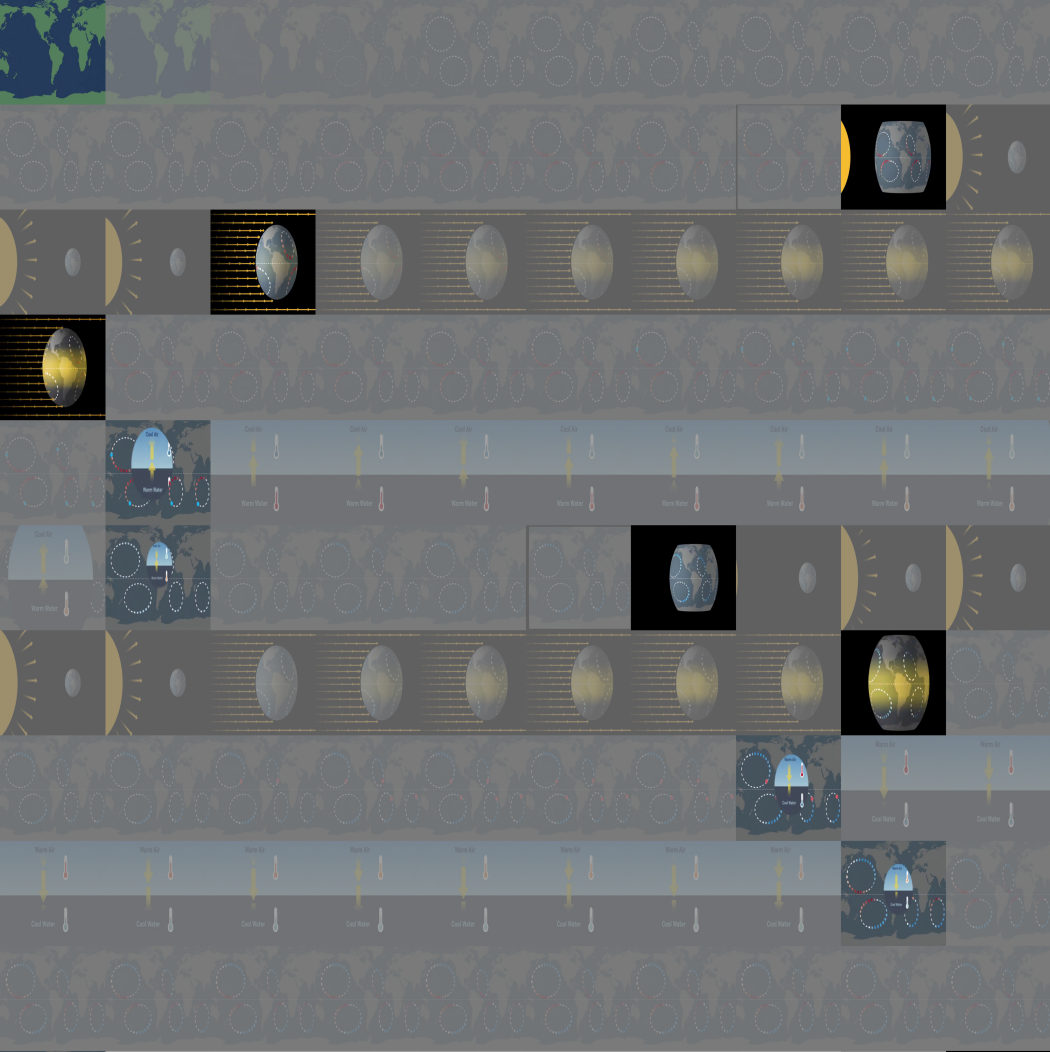}
        \subcaption{Scene Transition Point w/ Trigger Time}
    \end{minipage}%
    \hfill
    \begin{minipage}[t]{0.6\textwidth}
        \centering
        \includegraphics[width=\linewidth]{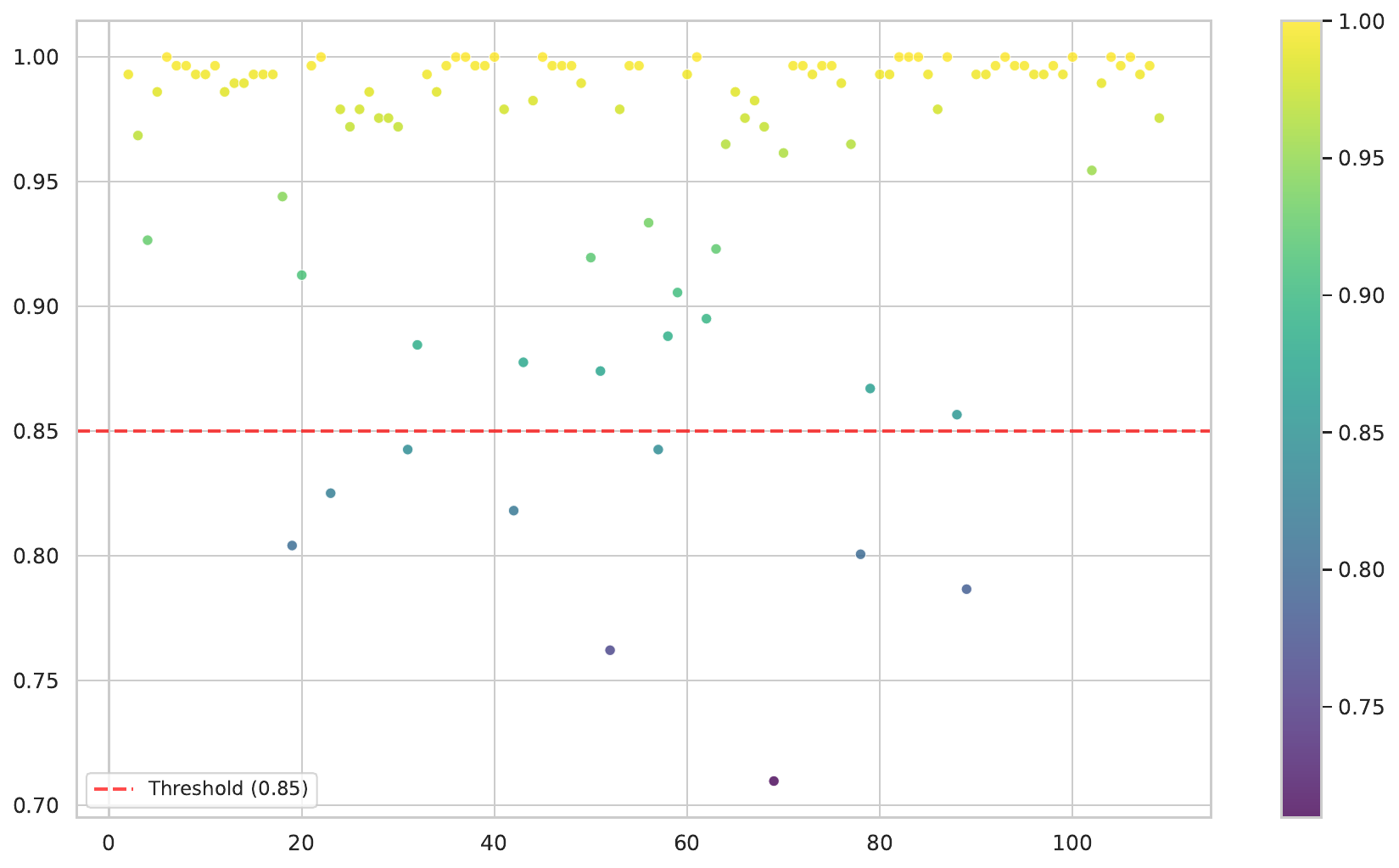}
        \subcaption{Drop Ratio - Timeline Curve}
    \end{minipage}
    \caption{Visualization of monitored \textit{Trigger Time} via drop ratio curve. The colored highlighted frames correspond to trigger times that reveal video scene transitions. Our model utilizes a temporal patch size of 2.}
    \label{fig:supp_triggertime_case2}
\end{figure*}

\begin{figure*}[h]
    \centering
    \includegraphics[width=\linewidth]{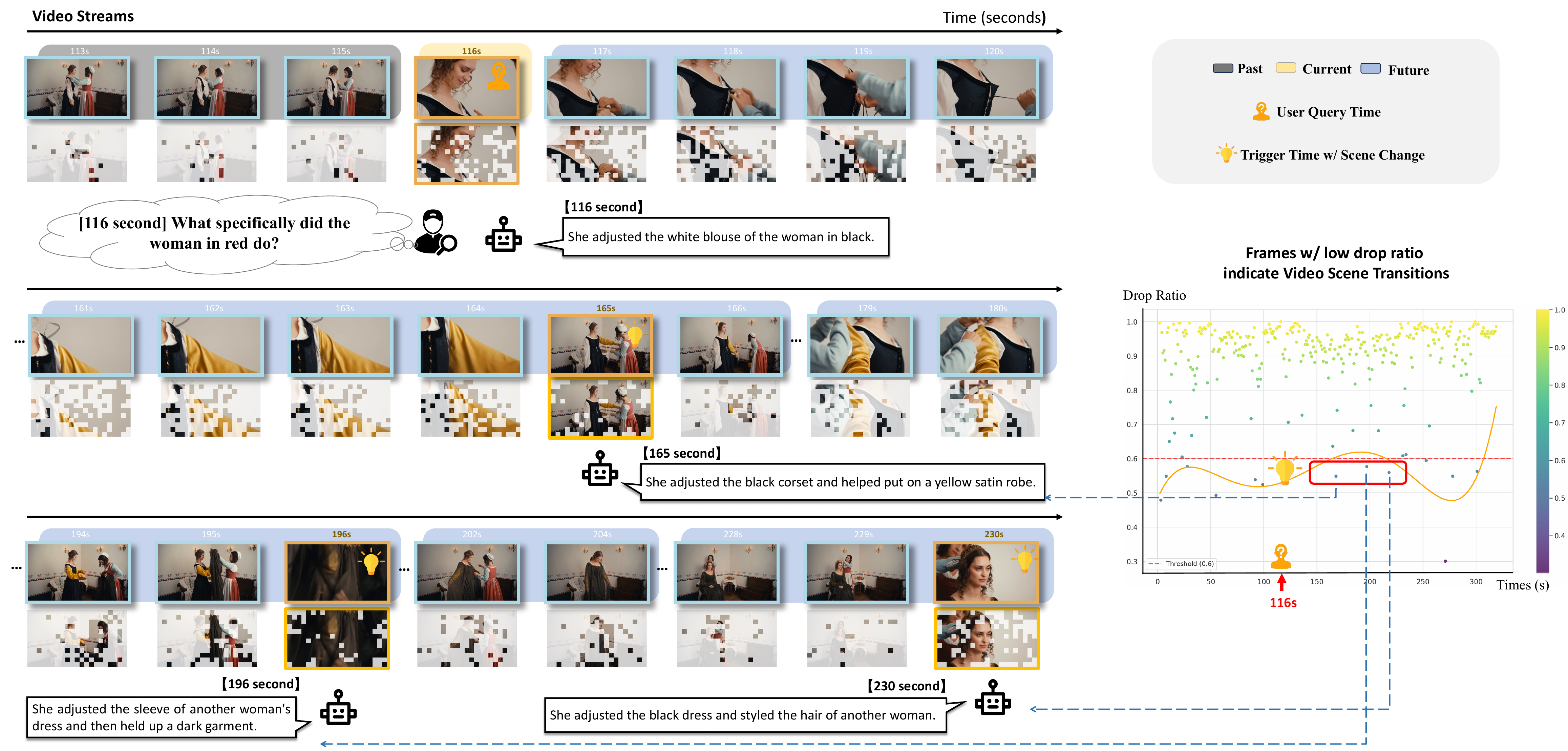}
    \caption{Case study of \model on StreamingBench with drop ratio curve. When a user proposes a question ``What specifically did the woman in red do?'' that can also be answered by the future moments, \model will proactively generate responses at the future trigger time (i.e., the video scene transition timestamps), which are indicated by the frames with low token drop ratios.}
    \label{fig:supp_streamingbench_case}
\end{figure*}

\begin{figure*}[t]
    \centering
    \begin{minipage}[t]{0.33\textwidth}
        \centering
        \includegraphics[width=\linewidth]{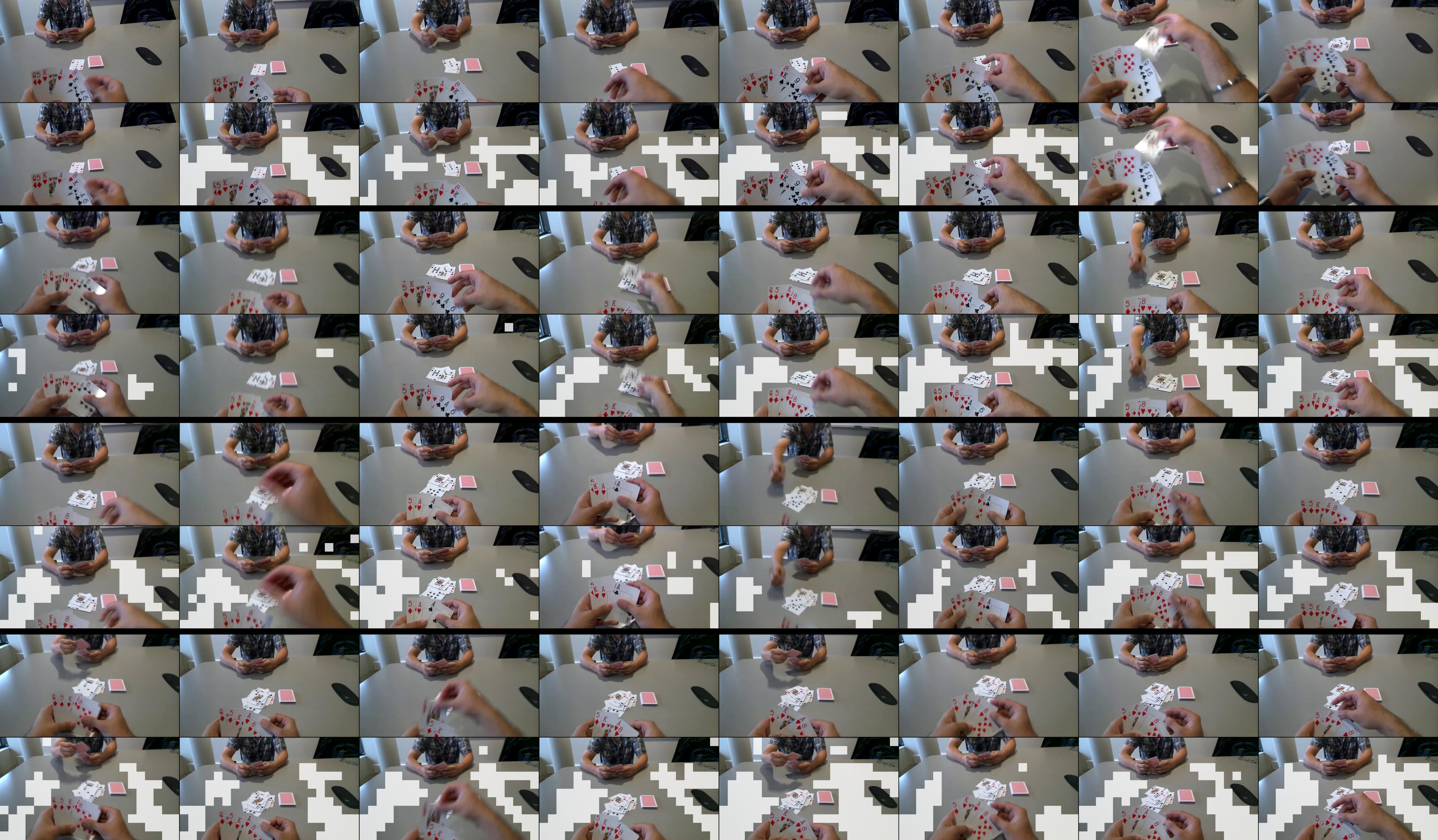}
        \subcaption{Pixel-level: $\tau_{pixel}=0.01$}
    \end{minipage}%
    \hfill
    \begin{minipage}[t]{0.33\textwidth}
        \centering
        \includegraphics[width=\linewidth]{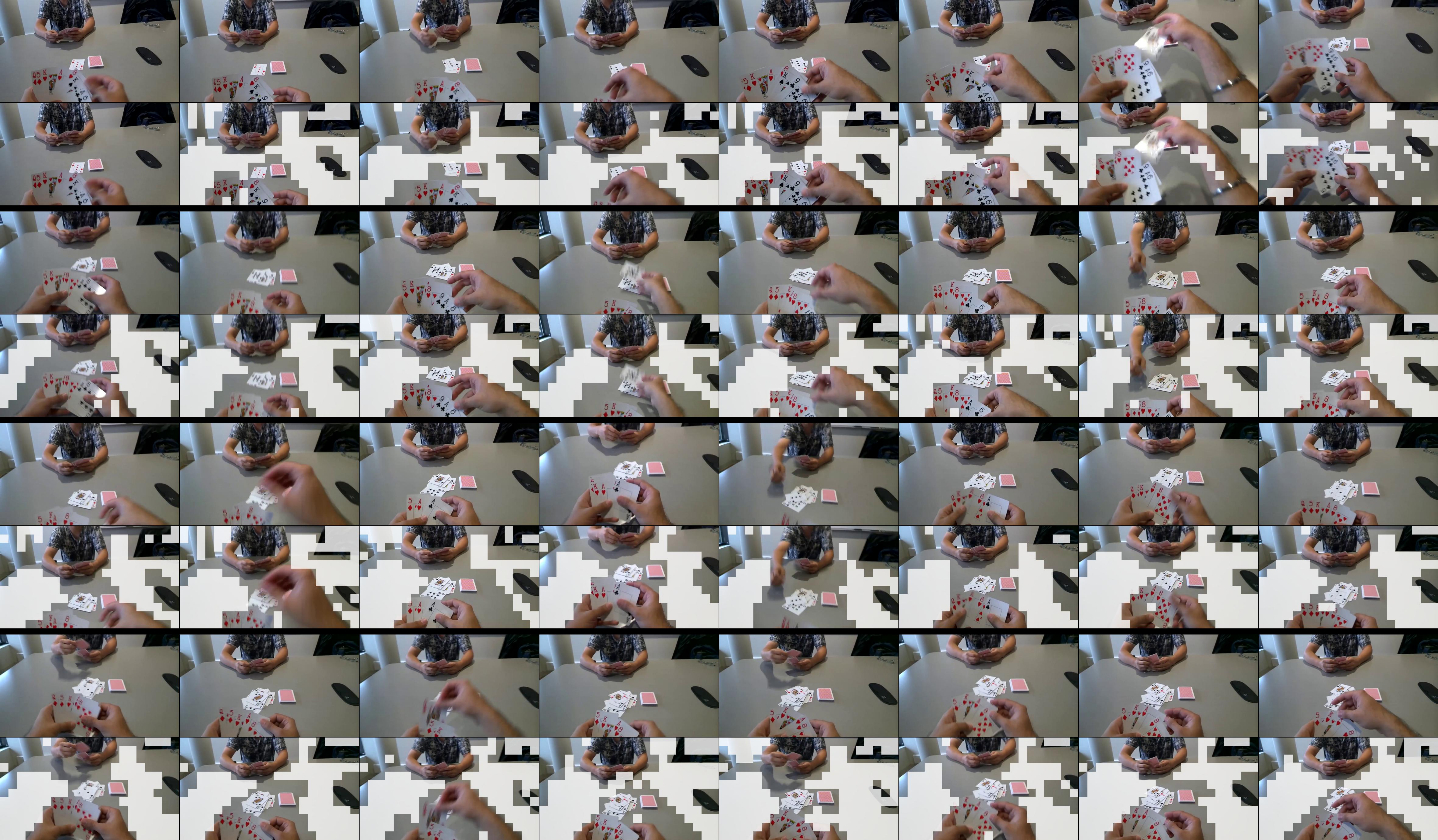}
        \subcaption{Pixel-level: $\tau_{pixel}=0.05$}
        
    \end{minipage}
    \hfill
    \begin{minipage}[t]{0.33\textwidth}
        \centering
        \includegraphics[width=\linewidth]{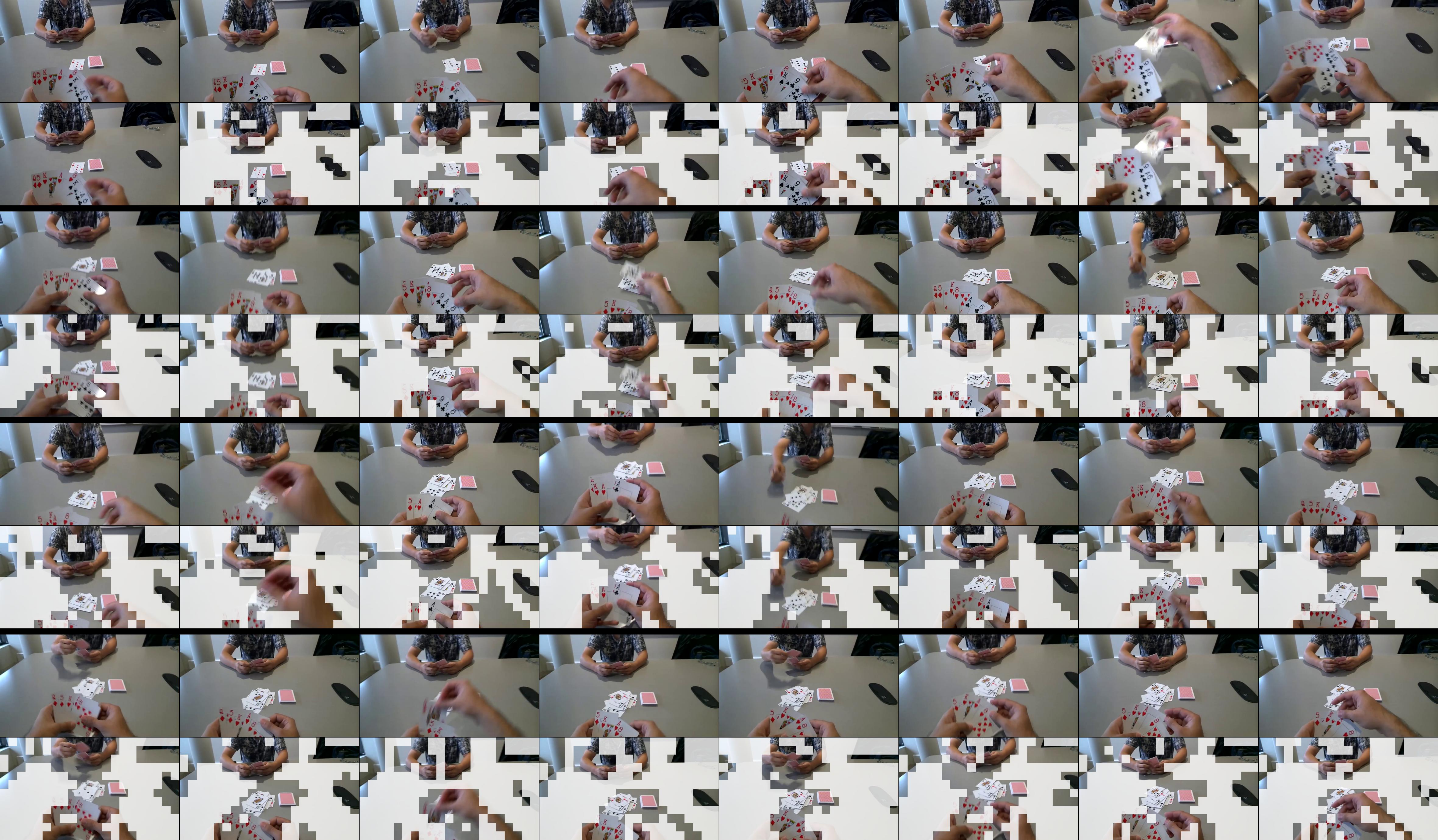}
        \subcaption{Pixel-level: $\tau_{pixel}=0.1$}
    \end{minipage}

    \vspace{0.5cm}
    
    \begin{minipage}[t]{0.33\textwidth}
        \centering
        \includegraphics[width=\linewidth]{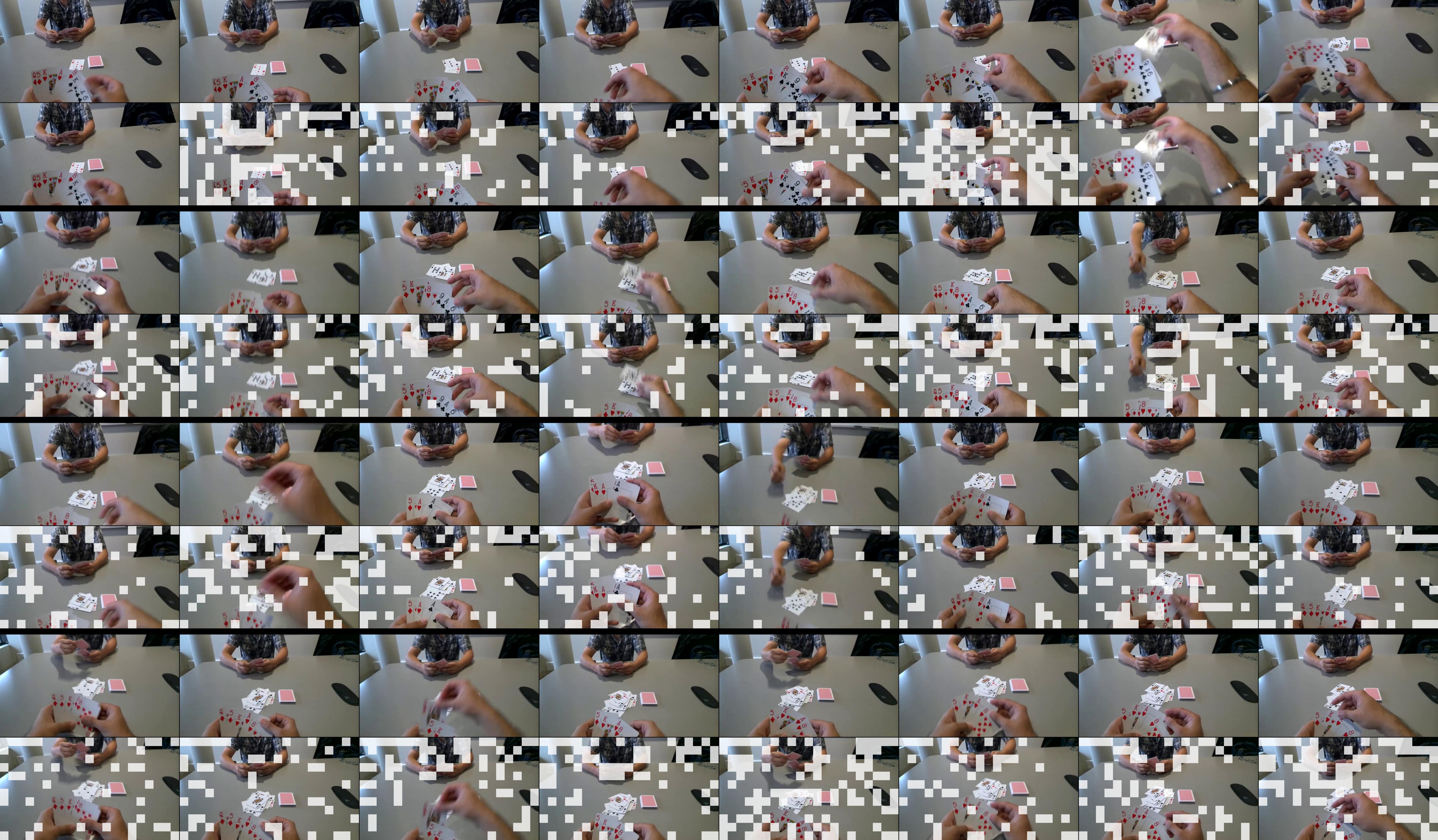}
        \subcaption{Feature-level: $\tau_{feat}=0.7$}
        \label{fig:supp_case42_c}
    \end{minipage}%
    \hfill
    \begin{minipage}[t]{0.33\textwidth}
        \centering
        \includegraphics[width=\linewidth]{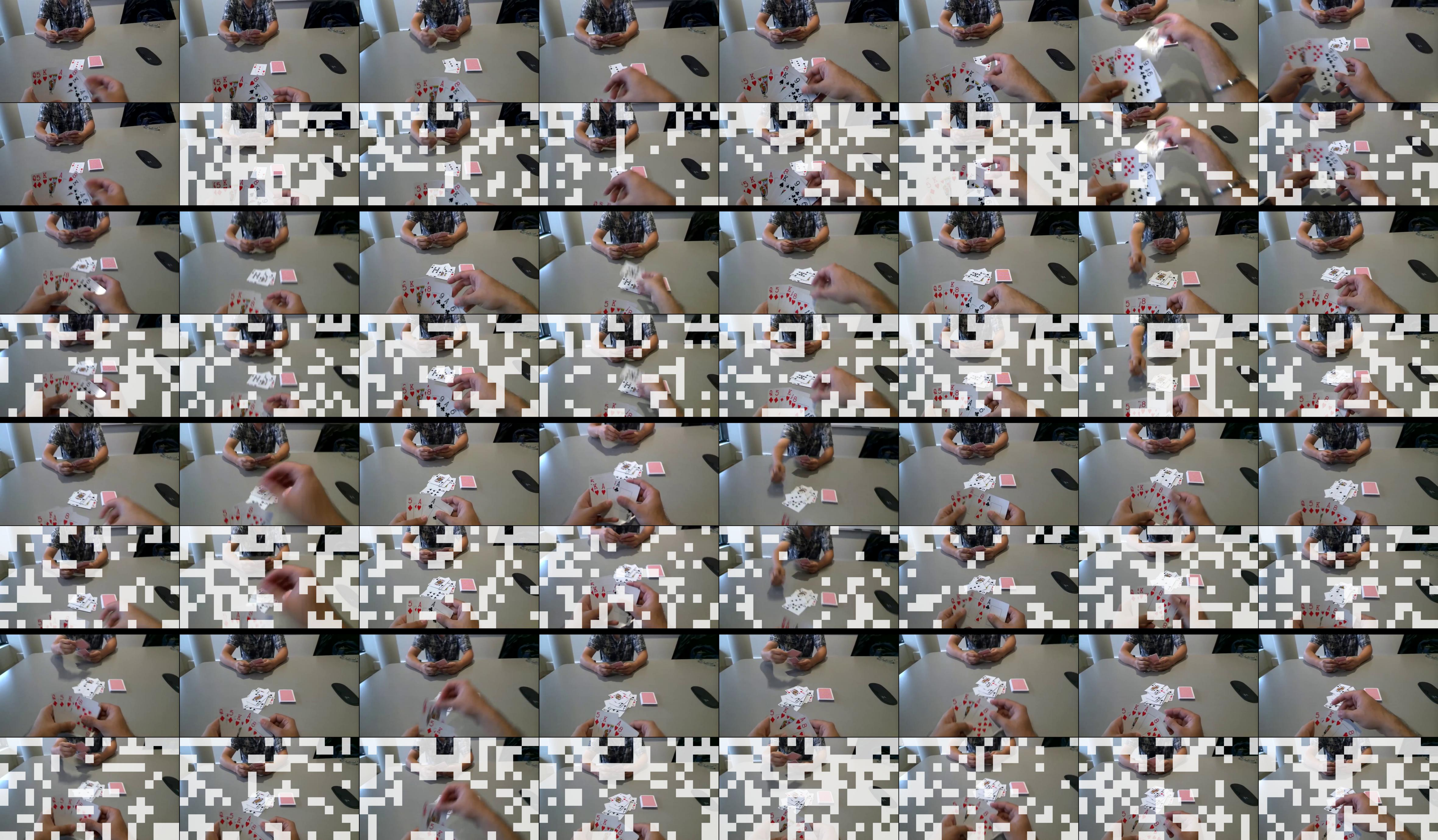}
        \subcaption{Feature-level: $\tau_{feat}=0.6$}
        \label{fig:supp_case42_b}
    \end{minipage}
    \hfill
    \begin{minipage}[t]{0.33\textwidth}
        \centering
        \includegraphics[width=\linewidth]{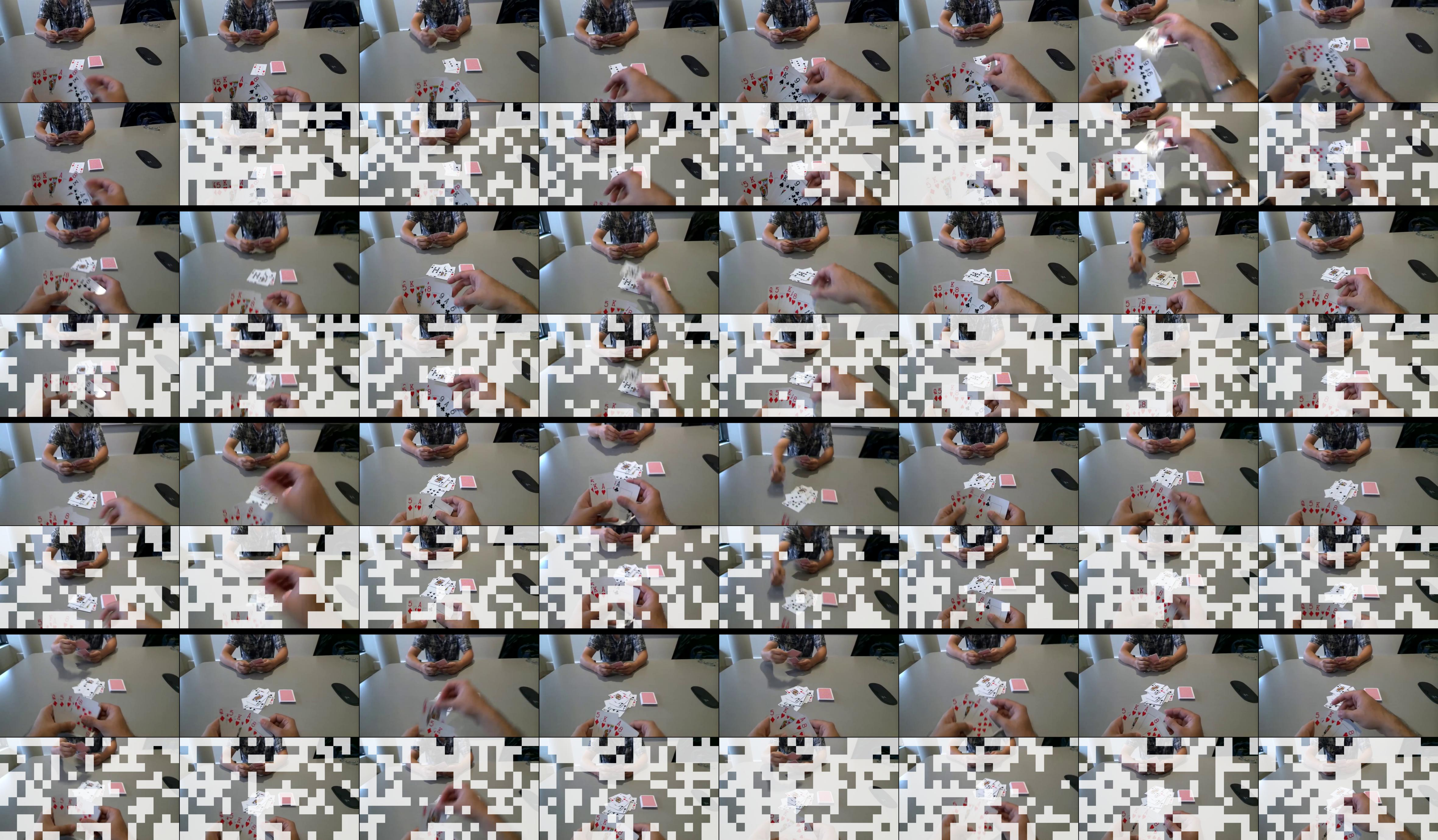}
        \subcaption{Feature-level: $\tau_{feat}=0.5$}
        \label{fig:supp_case42_a}
    \end{minipage}
    \caption{Comprehensive visualization of feature-level and pixel-level token dropping with varying threshold values ($\tau_{feat}$) for the same video case.}
    \label{fig:supp_case42}
\end{figure*}

\begin{figure*}[t]
    \centering
    \begin{minipage}[t]{0.48\textwidth}
        \centering
        \includegraphics[width=\linewidth]{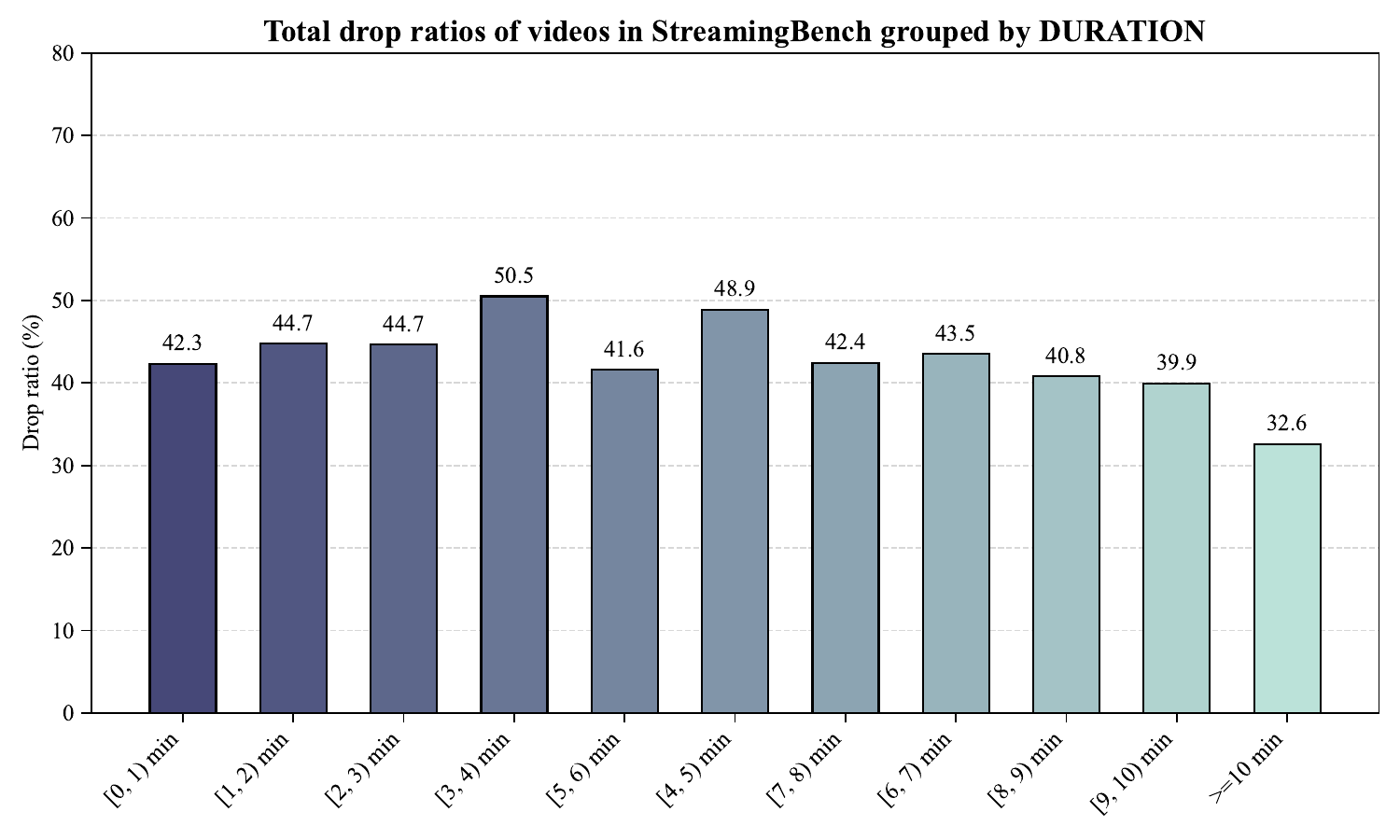}
        \subcaption{StreamingBench}
        \label{fig:supp_case752_ft}
    \end{minipage}%
    \hfill
    \begin{minipage}[t]{0.48\textwidth}
        \centering
        \includegraphics[width=\linewidth]{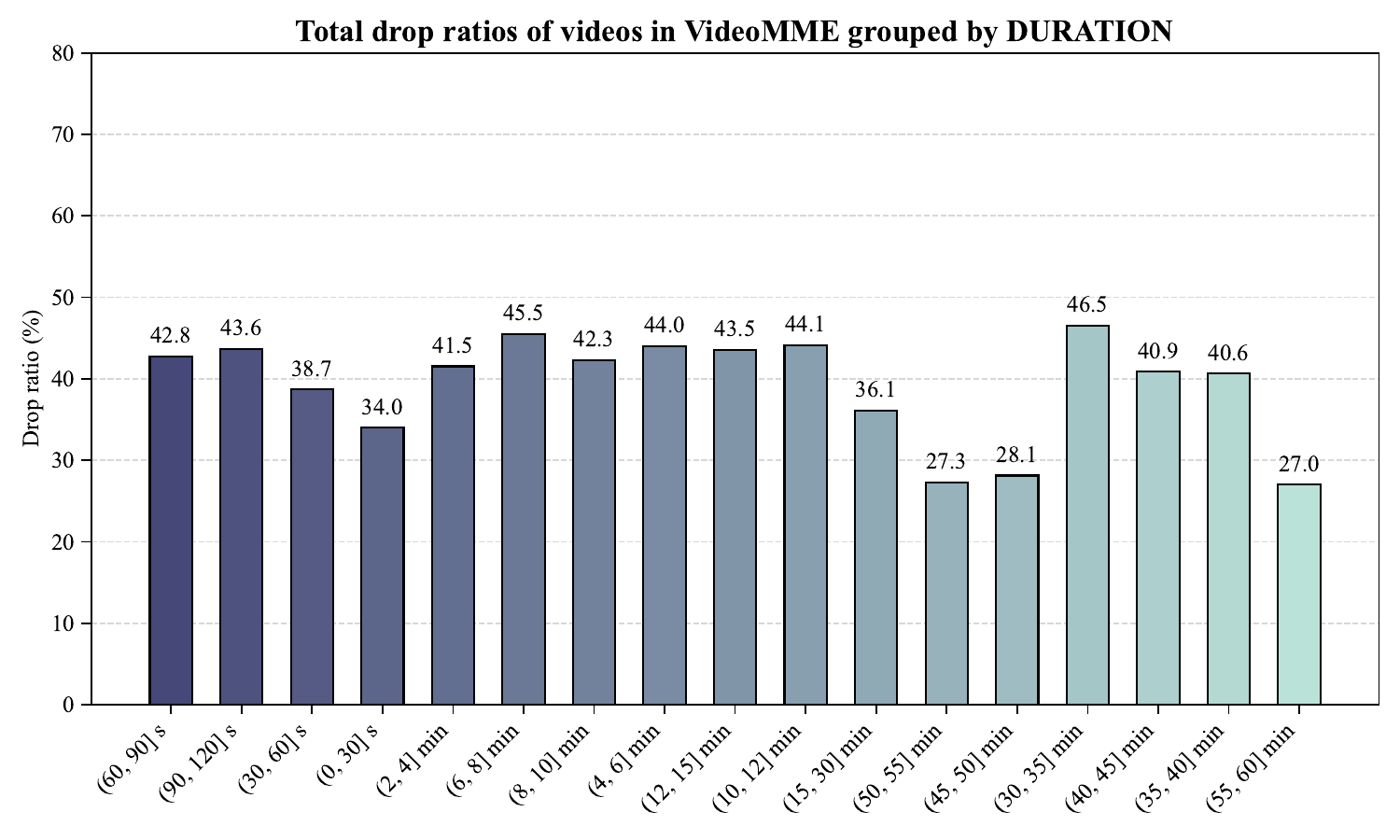}
        \subcaption{VideoMME}
    \end{minipage}
    \caption{Average token drop ratio across different video durations using the same threshold parameter $\tau$.}
    \label{fig:supp_duration}
\end{figure*}

\begin{figure*}[t]
    \centering
    \begin{minipage}[t]{0.48\textwidth}
        \centering
        \includegraphics[width=\linewidth]{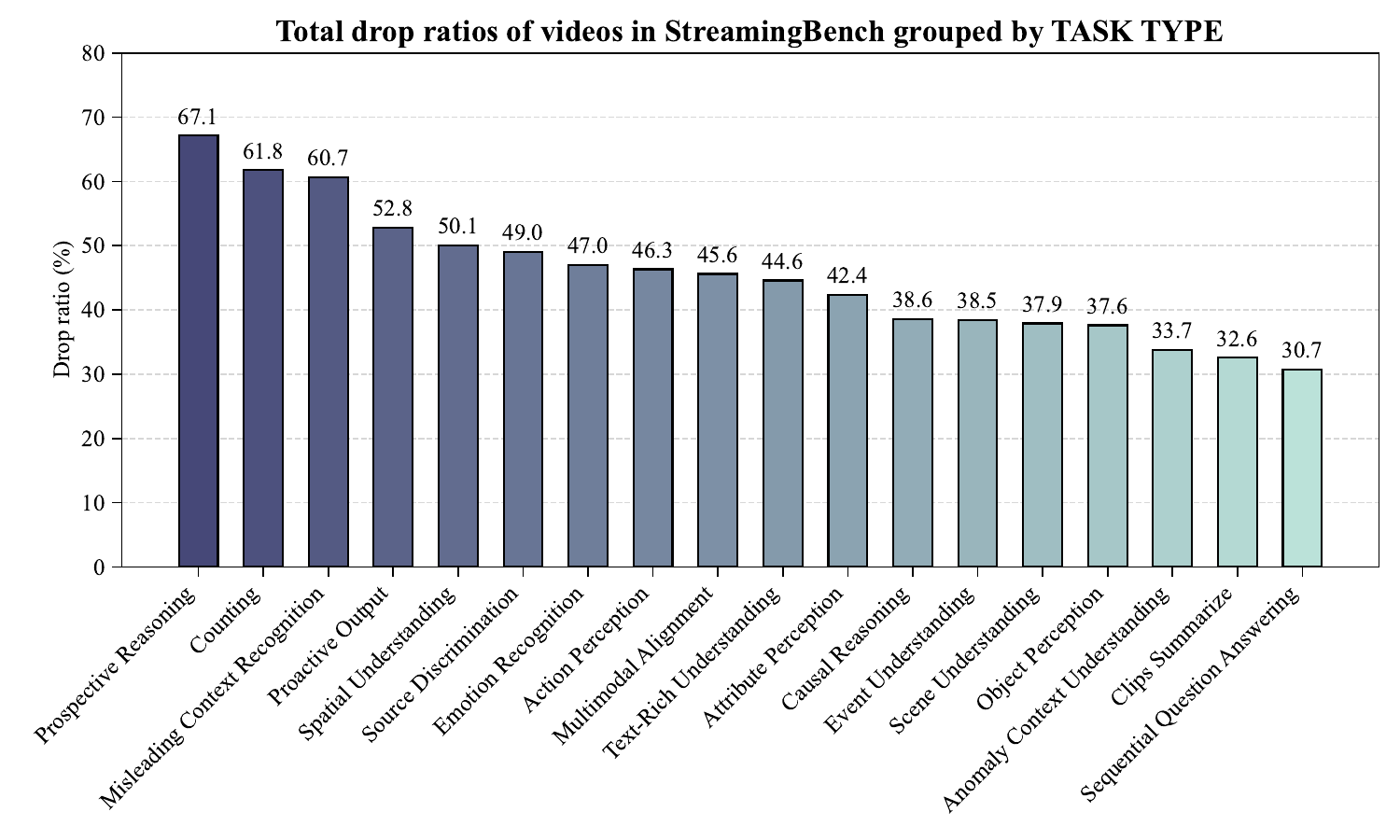}
        \subcaption{StreamingBench}
        \label{fig:supp_case752_ft}
    \end{minipage}%
    \hfill
    \begin{minipage}[t]{0.48\textwidth}
        \centering
        \includegraphics[width=\linewidth]{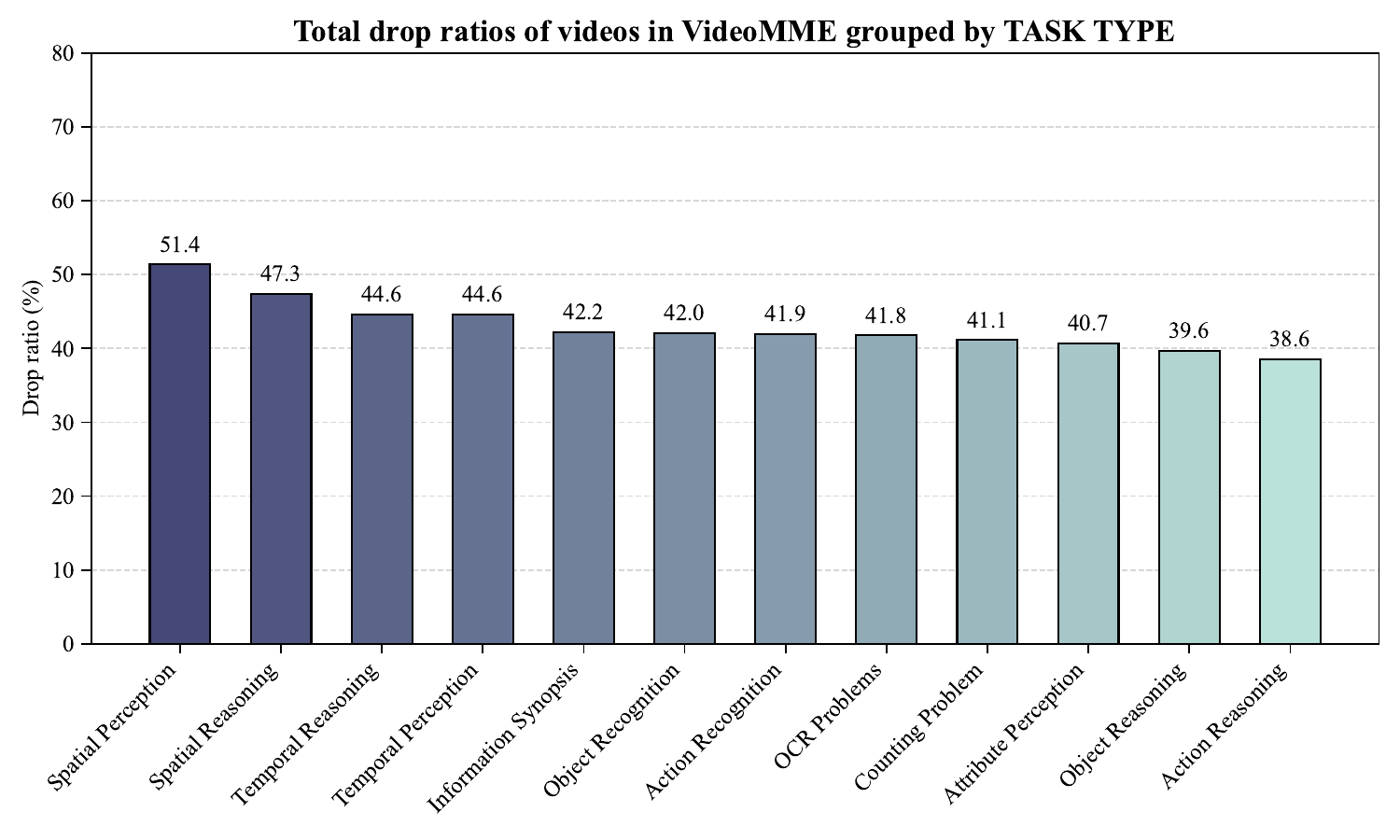}
        \subcaption{VideoMME}
    \end{minipage}
    \caption{Average token drop ratio across different task types using same hyperparameter $\tau$ settings.}
    \label{fig:supp_task_type}
\end{figure*}

We visualize qualitative results in the following figures:
\begin{itemize}

    \item \textbf{Figure~\ref{fig:supp_case752}} and \textbf{Figure~\ref{fig:supp_case671}}: Visualization of Feature-level token dropping and Pixel-level token dropping. These demonstrate that \module performs \textit{video-aware dynamic pruning} based on specific video content. With identical $\tau$ hyperparameters, different frames within the same video exhibit varying drop ratios due to adaptive temporal redundancy. Different videos also show significant redundancy differences: informative videos with diverse scenes have less redundancy, while monotonous videos with static scenes display high redundancy.

    \item \textbf{Figure~\ref{fig:supp_case42}} demonstrates how different hyperparameter $\tau$ values lead to different levels of redundancy control.
    
    \item \textbf{Figure~\ref{fig:supp_triggertime_case1}} and \textbf{Figure~\ref{fig:supp_triggertime_case2}} illustrate the \textit{Trigger Time} naturally monitored in the drop ratio-timeline curve, where valleys of low drop ratio reveal video scene transitions. \model leverages this scene change detection without requiring additional perception modules as in Dispider~\cite{dispider}, thereby reducing response delay.
    
    \item \textbf{Figure~\ref{fig:supp_streamingbench_case}} shows a case study of \model on StreamingBench with specific drop ratio curve. When a user proposes a question ``What specifically did the woman in red do?'' that can also be answered by the future moments, \model will proactively generate responses at the future trigger time (i.e., the video scene transition timestamps), which are indicated by the frames (165 seconds, 196 seconds and 230 seconds) with low token drop ratios.

    \item \textbf{Figure~\ref{fig:supp_duration}} shows the average drop ratio across different video durations. Since temporal redundancy is inherent to a video, it has little correlation with video length. The analysis reveals that spatial-related tasks may have more visual redundancy, while temporal-reasoning tasks typically have less visual redundancy as they require more informative visual content for reasoning and response.
    
    \item \textbf{Figure~\ref{fig:supp_task_type}} presents the average drop ratio across different video subtask types.

\end{itemize}

\begin{table*}[t]
    \centering
    \small
    
    \begin{tcolorbox}

\textbf{GPT-4 Prompt for Detailed Video Frame Caption Generation:}\\
You are an advanced AI visual assistant tasked with describing frames extracted from a video clip. When provided with a frame, describe it in detailed and accurate terms, focusing on both static and dynamic elements visible within, as well as the shot type, object appearances and actions, environment and background variations, and camera movements. While your primary focus should be the current frame, you may reference the provided caption of the preceding frame to describe any relevant relationships between the two frames, particularly regarding camera movements, which often require contrasts between frames. Ensure your description reflects only the contents that can be determined in the current frame without analysis or speculation. Do not include any elements of this prompt in your response. Your caption should be in a narrative format, avoiding list-like itemizations. Begin with "This frame".

\textbf{Here's an output example:}

This frame shows an arrangement of six small white bowls placed on a dark granite countertop. Each bowl contains different ingredients key to the cooking recipe. Starting from the top left and moving clockwise, the first bowl has a white powder which seems like cornstarch. A hand is seen above this bowl, with a finger pointing towards it, indicating either an instruction or a choice selection.
The top middle bowl is empty, noticeably different from the rest. The top right bowl is filled with what appears to be brown sugar, granulated and slightly heaped.
In the bottom row, the bowl on the left is also empty, mirroring the top middle bowl. The bottom middle bowl contains a light yellow-brown liquid, which could be oil or a prepared marinade. The bottom right bowl is filled with a dark liquid, possibly soy sauce.
The lighting in this frame is uniform, ensuring all the details and texture of the ingredients are clearly visible. No background elements distract from the central subject. This static shot is focused on presenting the different ingredients required, providing a clear visual reference for the cooking process.

\textbf{The current frame is uploaded as a .jpg file, and here's some relevant information:}
\begin{verbatim}
<video_meta_information>
    <video_name>
    <video_id>
    <video_subtitle>
    <video_category>
    <video_duration>
</video_meta_information>
\end{verbatim}

\textbf{The caption of the preceding frame is provided below:}
\begin{verbatim}
<caption_of_preceding_frame>
\end{verbatim}

\end{tcolorbox}
\caption{The prompt template used for scene-oriented key frame caption generation.}
\label{tab:frame_caption_prompt}
\end{table*}

\begin{table*}[t]
    \centering
    \small
    
    \begin{tcolorbox}

\shortstack[l]{\textbf{GPT-4 Prompt for Streaming VideoQA Generation:} \\
Given the following video frame descriptions, generate 5 different question-answer pairs \\
that focus on temporal understanding and visual perception. These questions should cover \\
the following cognitive abilities:}

\vspace{0.5em}
1. Episodic Memory: Questions about where objects were located before events or recalling details from earlier frames\\
2. Action Sequence Identification: Questions about what actions occurred before others or the order of events\\
3. Hallucination Detection: Questions requiring verification of visual elements that actually appeared in the video\\
4. Future Prediction: Questions predicting what might happen next based on the video context\\
5. Sequential Steps Recognition: Questions about step-by-step processes shown in the video\\
6. Clues Reveal Responding: Questions requiring attention to specific details or clues spread across frames

\vspace{0.5em}
\shortstack[l]{\textbf{Video Description:} \\
\{description\}}

\vspace{0.5em}
\shortstack[l]{\textbf{Question Requirements:} \\
1. Create multiple-choice questions (A, B, C, D format) with one correct answer\\
2. Include at least one question from each of the six cognitive abilities listed above\\
3. Ensure questions require temporal understanding across frames\\
4. Include questions that require connecting information from different parts of the video\\
5. Create at least 2-3 questions that specifically require comparing frames with long distances between them}

\vspace{0.5em}
\shortstack[l]{\textbf{CRITICAL REQUIREMENTS:} \\
1. NEVER reference specific frame numbers/indices in questions or answers (e.g., DO NOT say "In Frame 201" or "From Frame 10 to Frame 20")\\
2. Instead, describe the content and events directly (e.g., "When the person was in the kitchen" or "After picking up the cup")\\
3. You MAY reference timestamps if needed (e.g., "At 1:20 in the video")\\
4. Questions should focus on the content and temporal relationships, not on arbitrary frame numbering}

\vspace{0.5em}
\shortstack[l]{\textbf{Target Format:} \\
Please format your response as a JSON object using the following structure:}

\begin{verbatim}
[
    {
      "question": <question>,
      "options": <options>,
      "answer": <answer>,
      "answer_text": <answer_text>,
      "category": <category>,
      "rationale": <rationale>,
      "rationale_frames": <rationale_frames>
    }
]
\end{verbatim}

Ensure that your response includes 5 multiple-choice question-answer pairs in this JSON format, with each pair addressing different cognitive abilities from the six categories listed. **The specific frame numbers (index) or timestamps should NOT appear in the questions or options themselves.**

    \end{tcolorbox}

    \caption{The prompt template used for generating streaming question-answer pairs.}
    \label{tab:question_generation_prompt}
\end{table*}


\end{document}